\documentclass[12pt,journal,draftcls,letterpaper,onecolumn]{IEEEtran}

\usepackage{times}
\usepackage{graphicx}
\usepackage{float}
\usepackage{amsmath}
\usepackage{amssymb}
\usepackage{algorithm}
\usepackage{algorithmic}
\usepackage{booktabs}
\usepackage{hyperref}

\newtheorem{Lem}{Lemma}

\def\eg{\emph{e.g. }}
\def\ie{\emph{i.e. }}
\def\etc{\emph{etc. }}

\begin{document}

\pubid{0000--0000/00\$00.00~\copyright~201X
IEEE}

\title{Moving Object Detection by Detecting Contiguous Outliers in the Low-Rank Representation}

\author{Xiaowei Zhou, Can Yang and Weichuan Yu
\thanks{The authors of this manuscript are with the Department of Electronic and Computer Engineering, The Hong Kong University of Science and Technology, Hong Kong SAR, China}
}



\IEEEcompsoctitleabstractindextext{%
\begin{abstract}
Object detection is a fundamental step for automated video analysis in many vision applications. Object detection in a video is usually performed by object detectors or background subtraction techniques. Often, an object detector requires manually labeled examples to train a binary classifier, while background subtraction needs a training sequence that contains no objects to build a background model. To automate the analysis, object detection without a separate training phase becomes a critical task. People have tried to tackle this task by using motion information. But existing motion-based methods are usually limited when coping with complex scenarios such as nonrigid motion and dynamic background. In this paper, we show that above challenges can be addressed in a unified framework named DEtecting Contiguous Outliers in the LOw-rank Representation (DECOLOR). This formulation integrates object detection and background learning into a single process of optimization, which can be solved by an alternating algorithm efficiently. We explain the relations between DECOLOR and other sparsity-based methods. Experiments on both simulated data and real sequences demonstrate that DECOLOR outperforms the state-of-the-art approaches and it can work effectively on a wide range of complex scenarios.
\end{abstract}
\begin{keywords}
Moving object detection, low-rank modeling, Markov Random Fields, motion segmentation.
\end{keywords}}

\maketitle

\section{Introduction}

Automated video analysis is important for many vision applications such as surveillance, traffic monitoring, augmented reality, vehicle navigation, \etc \cite{yilmaz2006object,moeslund2006survey}. As pointed out in \cite{yilmaz2006object}, there are three key steps for automated video analysis: object detection, object tracking and behavior recognition. As the first step, object detection aims to locate and segment interesting objects in a video. Then, such objects can be tracked from frame to frame, and the tracks can be analyzed to recognize object behavior. Thus, object detection plays a critical role in practical applications.

Object detection is usually achieved by object detectors or background subtraction \cite{yilmaz2006object}. An object detector is often a classifier that scans the image by a sliding window and labels each subimage defined by the window as either object or background. Generally, the classifier is built by offline learning on separate datasets \cite{papageorgiou1998general,viola2005detecting} or by online learning initialized with a manually labeled frame at the start of a video \cite{grabner2006line,babenko2011robust}. Alternatively, background subtraction \cite{piccardi2004background} compares images with a background model and detects the changes as objects. It usually assumes that no object appears in images when building the background model \cite{toyama1999wallflower,moeslund2006survey}. Such requirements of training examples for object or background modeling actually limit the applicability of above mentioned methods in automated video analysis.

\begin{figure*}
\centering
\begin{minipage}[b]{0.49\linewidth}
    \begin{minipage}[b]{0.32\linewidth}
        \centerline{\includegraphics[width=\linewidth,height=0.82\linewidth]{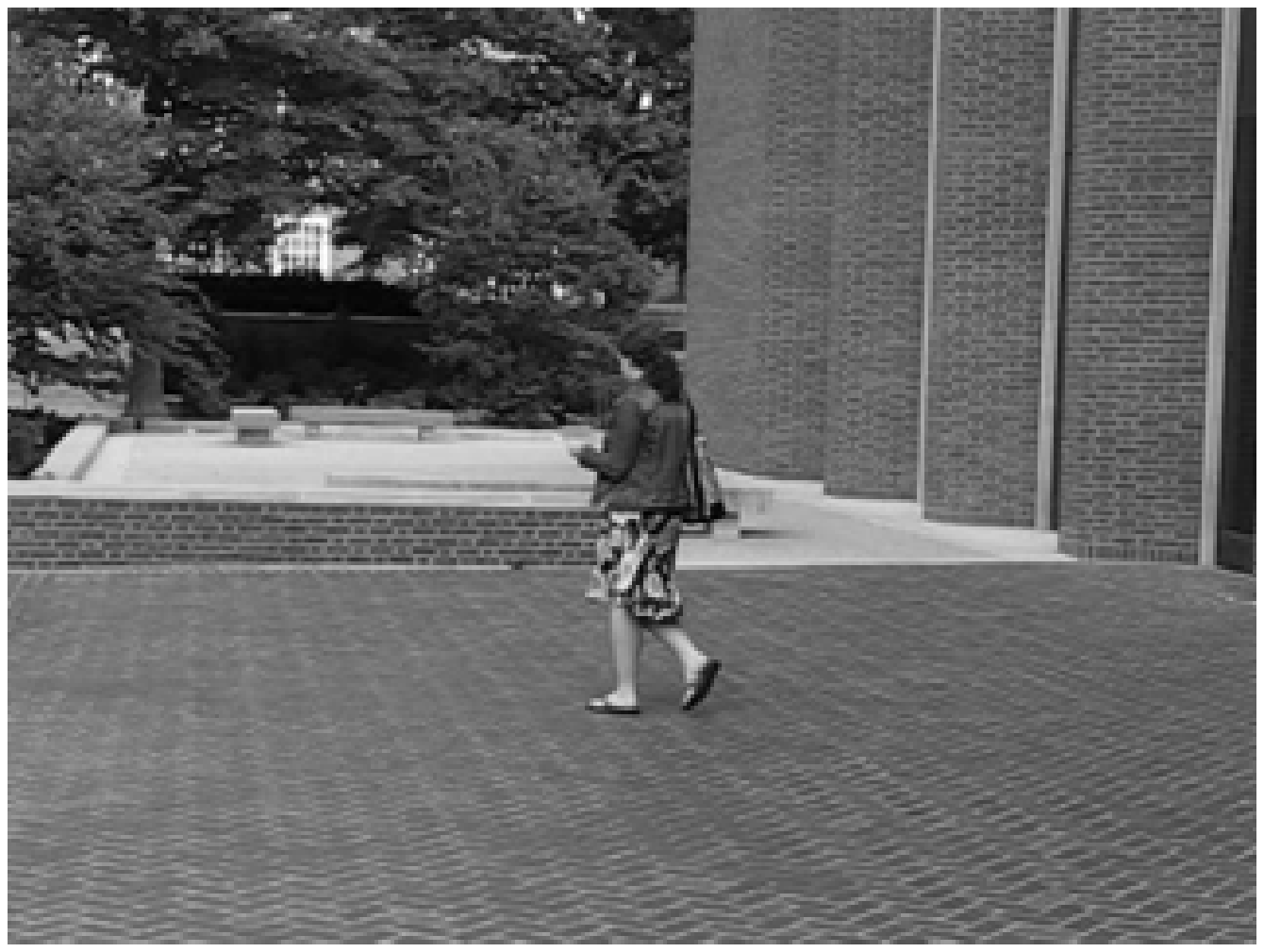}}
    \end{minipage}
    \begin{minipage}[b]{0.32\linewidth}
        \centerline{\includegraphics[width=\linewidth,height=0.82\linewidth]{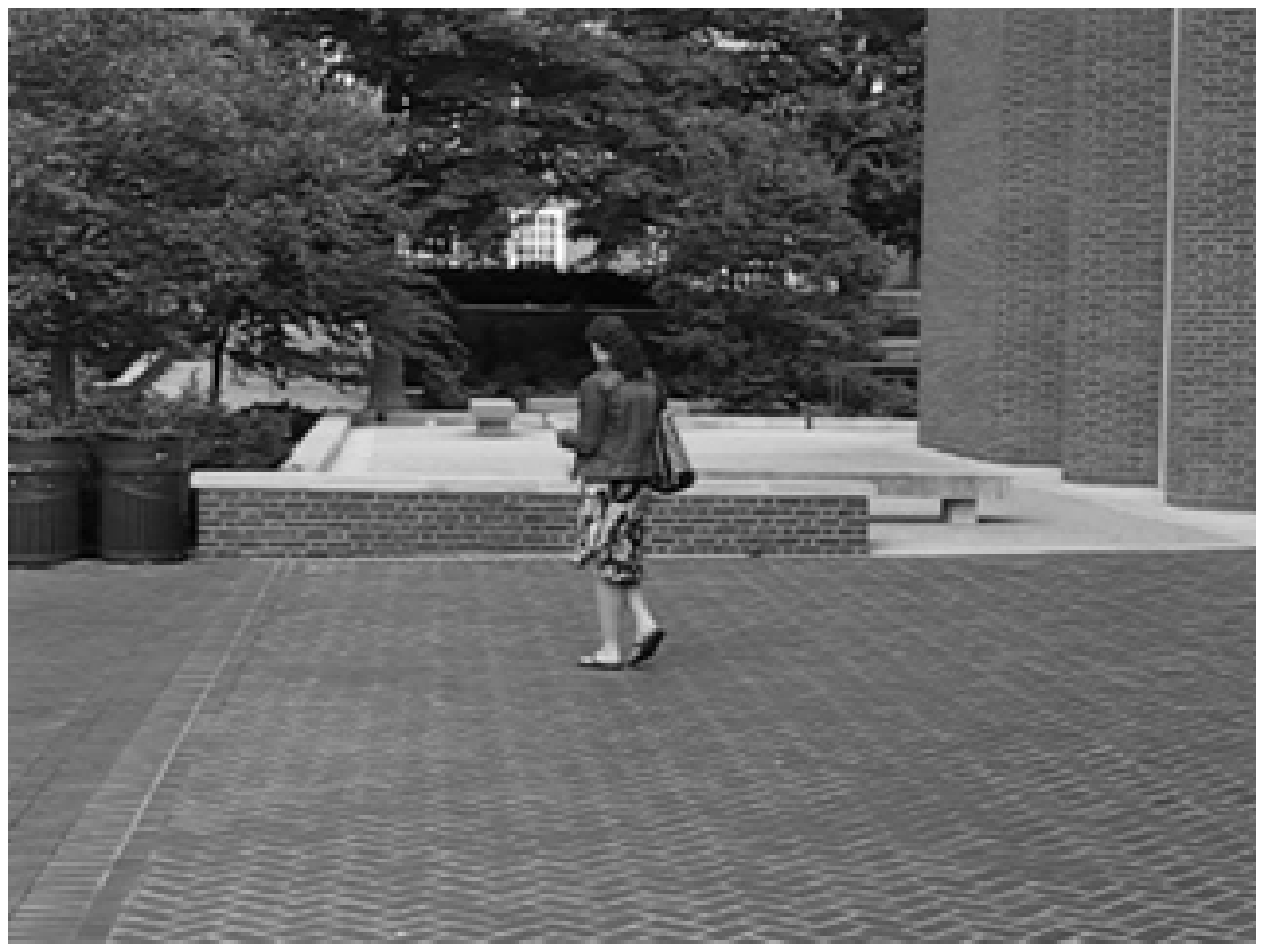}}
    \end{minipage}
    \begin{minipage}[b]{0.32\linewidth}
        \centerline{\includegraphics[width=\linewidth,height=0.82\linewidth]{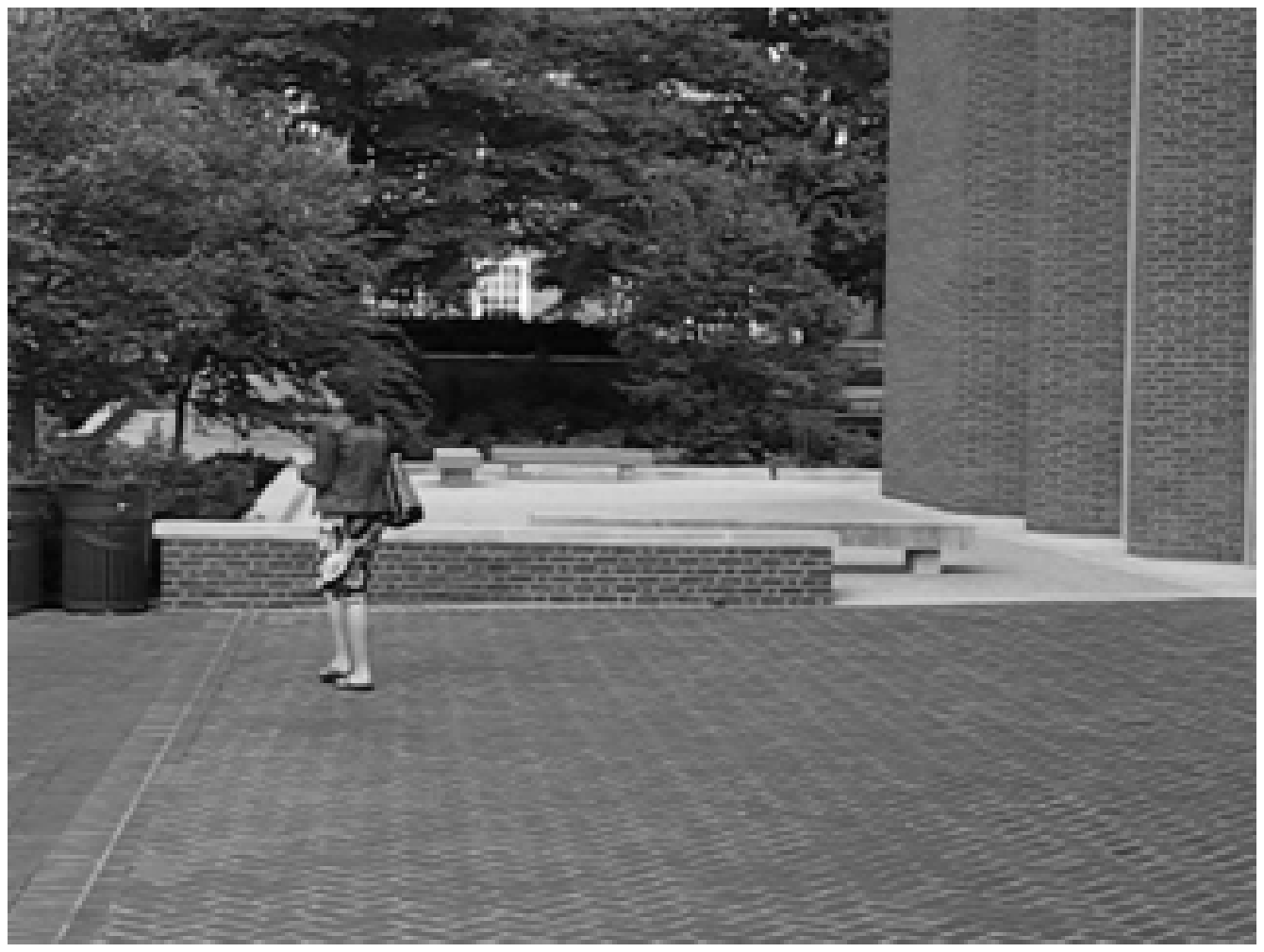}}
    \end{minipage}
    \centerline{(a)}
\end{minipage}
\hfill
\begin{minipage}[b]{0.49\linewidth}
    \begin{minipage}[b]{0.32\linewidth}
        \centerline{\includegraphics[width=\linewidth,height=0.82\linewidth]{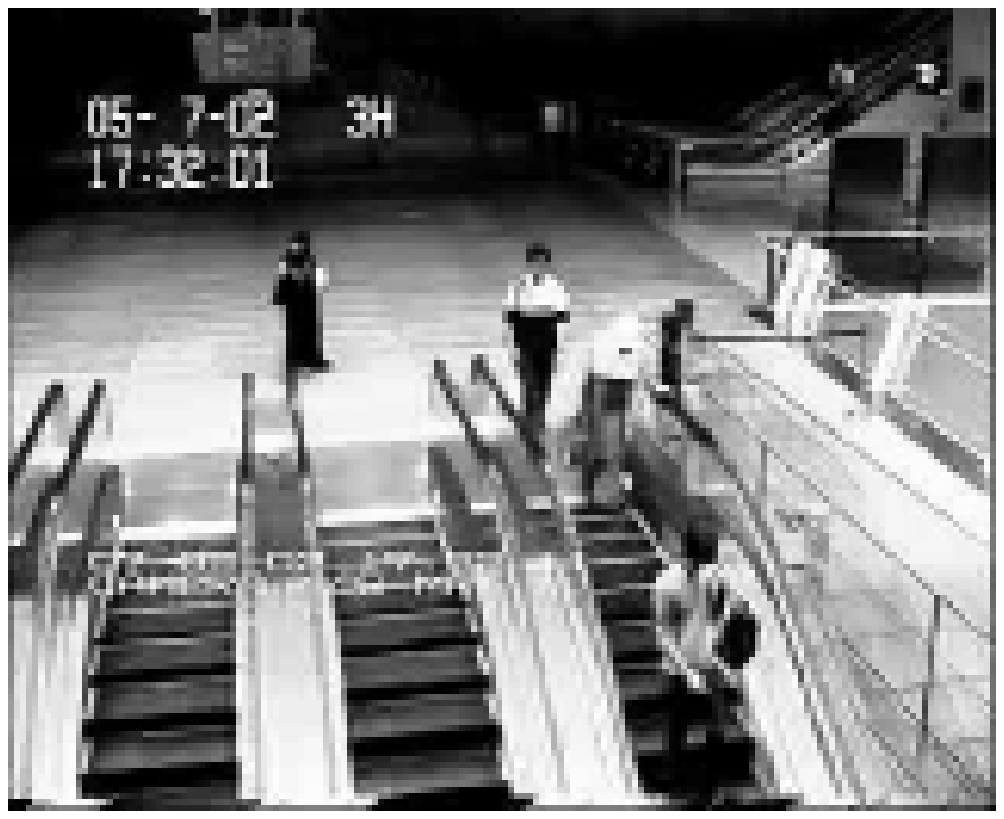}}
    \end{minipage}
    \begin{minipage}[b]{0.32\linewidth}
        \centerline{\includegraphics[width=\linewidth,height=0.82\linewidth]{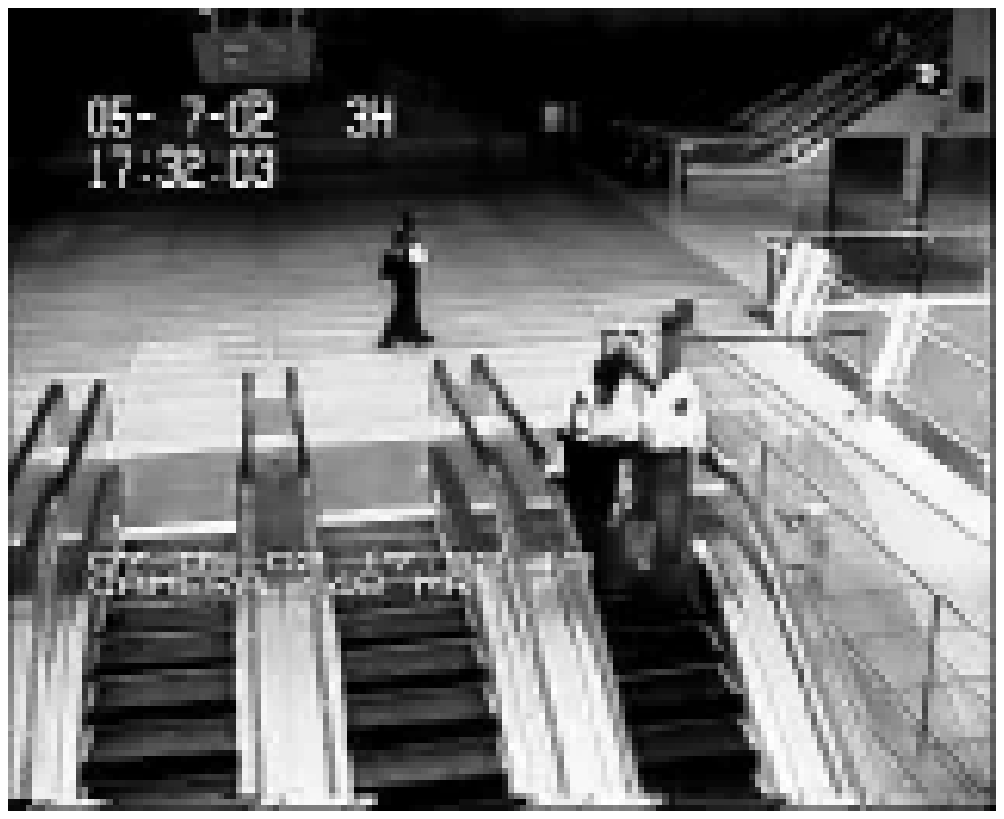}}
    \end{minipage}
    \begin{minipage}[b]{0.32\linewidth}
        \centerline{\includegraphics[width=\linewidth,height=0.82\linewidth]{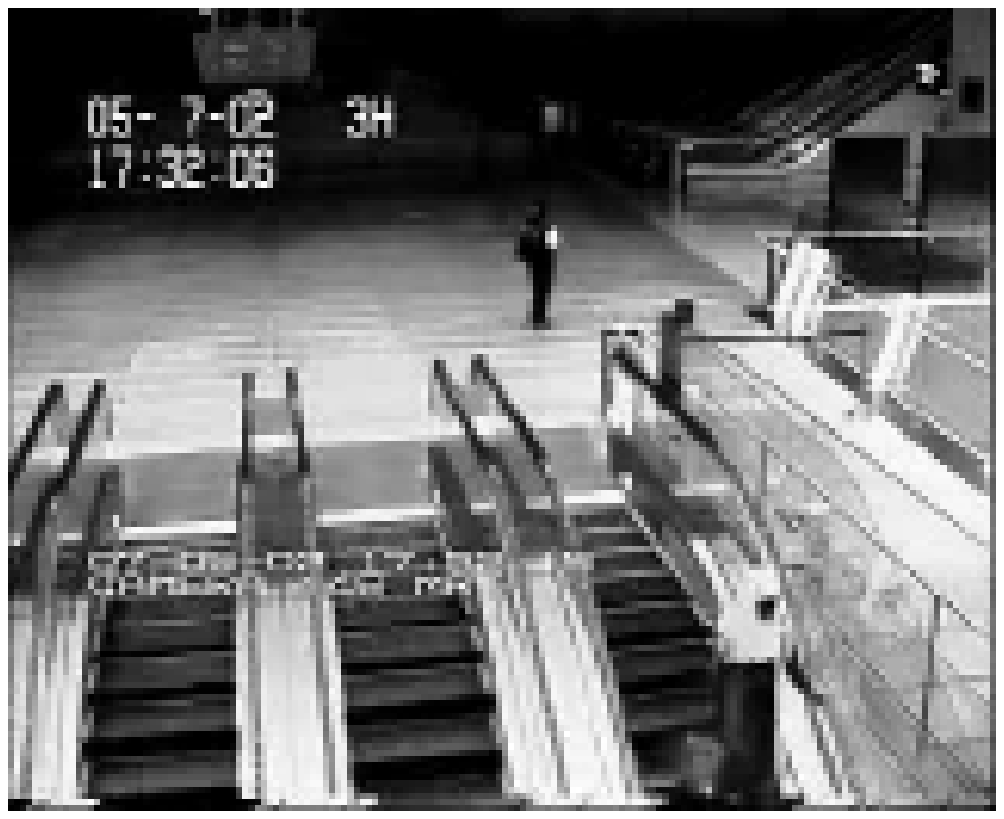}}
    \end{minipage}
    \centerline{(b)}
\end{minipage}
\caption{\small Two examples to illustrate the problem. (a) A sequence of 40 frames, where a walking lady is recorded by a hand-held camera. From left to right are the 1st, 20th and 40th frames. (b) A sequence of 48 frames clipped from a surveillance video at the airport. From left to right are the 1st, 24th and 48th frames. Notice that the escalator is moving. The objective is to segment the moving people automatically without extra inputs. }\label{Fig_example}
\end{figure*}

Another category of object detection methods that can avoid training phases are motion-based methods \cite{yilmaz2006object,moeslund2006survey}, which only use motion information to separate objects from the background. The problem can be rephrased as follows. \emph{Given a sequence of images in which foreground objects are present and moving differently from the background, can we separate the objects from the background automatically?} Fig. \ref{Fig_example}(a) shows such an example, where a walking lady is always present and recorded by a hand-held camera. 
The goal is to take the image sequence as input and directly output a mask sequence of the walking lady.

The most natural way for motion-based object detection is to classify pixels according to motion patterns, which is usually named motion segmentation \cite{vidal2004unified,Cremers05}. These approaches achieve both segmentation and optical flow computation accurately and they can work in the presence of large camera motion. However, they assume rigid motion \cite{vidal2004unified} or smooth motion \cite{Cremers05} in respective regions, which is not generally true in practice. In practice, the foreground motion can be very complicated with nonrigid shape changes. Also, the background may be complex, including illumination changes and varying textures such as waving trees and sea waves. Fig. \ref{Fig_example}(b) shows such a challenging example. The video includes an operating escalator, but it should be regarded as background for human tracking purpose. An alternative motion-based approach is background estimation \cite{gutchess2001background,nair2004unsupervised}. Different from background subtraction, it estimates a background model directly from the testing sequence. Generally, it tries to seek temporal intervals inside which the pixel intensity is unchanged and uses image data from such intervals for background estimation. However, this approach also relies on the assumption of static background. Hence, it is difficult to handle the scenarios with complex background or moving cameras.

In this paper, we propose a novel algorithm for moving object detection, which falls into the category of motion-based methods. It solves the challenges mentioned above in a unified framework named DEtecting Contiguous Outliers in the LOw-rank Representation (DECOLOR). We assume that the underlying background images are linearly correlated. Thus, the matrix composed of vectorized video frames can be approximated by a low-rank matrix, and the moving objects can be detected as outliers in this low-rank representation. Formulating the problem as outlier detection allows us to get rid of many assumptions on the behavior of foreground. The low-rank representation of background makes it flexible to accommodate the global variations in the background. Moreover, DECOLOR performs object detection and background estimation simultaneously without training sequences. The main contributions can be summarized as follows:
\begin{itemize}
  \item[1.] We propose a new formulation of outlier detection in the low-rank representation, in which the outlier support and the low-rank matrix are estimated simultaneously.
      We establish the link between our model and other relevant models in the framework of Robust Principle Component Analysis (RPCA) \cite{Candes09}. Different from other formulations of RPCA, we model the outlier support explicitly. DECOLOR can be interpreted as $\ell_0$-penalty regularized RPCA, which is a more faithful model for the problem of moving object segmentation. Following the novel formulation, an effective and efficient algorithm is developed to solve the problem. We demonstrate that, although the energy is non-convex, DECOLOR achieves better accuracy in terms of both object detection and background estimation compared against the state-of-the-art algorithm of RPCA \cite{Candes09}.
  \item[2.] In other models of RPCA, no prior knowledge on the spatial distribution of outliers has been considered. In real videos, the foreground objects usually are small clusters. Thus, contiguous regions should be preferred to be detected. Since the outlier support is modeled explicitly in our formulation, we can naturally incorporate such contiguity prior using Markov Random Fields (MRFs) \cite{li2009markov}.
  \item[3.] We use a parametric motion model to compensate for camera motion. The compensation of camera motion is integrated into our unified framework and computed in a batch manner for all frames during segmentation and background estimation.
\end{itemize}

The MATLAB implementation of DECOLOR, experimental data and more results are publicly available at:
http://bioinformatics.ust.hk/decolor/decolor.html.

\section{Related Work}\label{section_ref}

Previous methods for object detection are vast, including object detectors (supervised learning), image segmentation, background subtraction, \etc \cite{yilmaz2006object}. Our method aims to segment objects based on motion information and it comprises a component of background modeling. Thus, motion segmentation and background subtraction are the most related topics to this paper.

\subsection{Motion Segmentation}

In motion segmentation, the moving objects are continuously present in the scene, and the background may also move due to camera motion. The target is to separate different motions.

A common approach for motion segmentation is to partition the dense optical-flow field \cite{black1996robust}. This is usually achieved by decomposing the image into different motion layers \cite{Amiaz06,Brox06,Cremers05}. The assumption is that the optical-flow field should be smooth in each motion layer, and sharp motion changes only occur at layer boundaries. Dense optical flow and motion boundaries are computed in an alternating manner named \emph{motion competition} \cite{Cremers05}, which is usually implemented in a level set framework. The similar scheme is later applied to dynamic texture segmentation \cite{chan2009layered,cremers2003dynamic,fazekas2009dynamic}. While high accuracy can be achieved in these methods, accurate motion analysis itself is a challenging task due to the difficulties raised by aperture problem, occlusion, video noises, \etc \cite{beauchemin1995computation}. Moreover, most of the motion segmentation methods require object contours to be initialized and the number of foreground objects to be specified \cite{Cremers05}.

An alternative approach for motion segmentation tries to segment the objects by analyzing point trajectories \cite{vidal2004unified,tron2007benchmark,sheikh2009camera,brox2010object}. Some sparse feature points are firstly detected and tracked throughout the video and then separated into several clusters via subspace clustering \cite{Vidal10subspace} or spectral clustering \cite{brox2010object}. The formulation is mathematically elegant and it can handle large camera motion. However, these methods require point trajectories as input and only output a segmentation of sparse points. The performance relies on the quality of point tracking and postprocessing is needed to obtain the dense segmentation \cite{ochs2011object}. Also, they are limited when dealing with noisy data and nonrigid motion \cite{Vidal10subspace}.

\subsection{Background Subtraction}

In background subtraction, the general assumption is that a background model can be obtained from a training sequence that does not contain foreground objects. Moreover, it usually assumes that the video is captured by a static camera \cite{piccardi2004background}. Thus, foreground objects can be detected by checking the difference between the testing frame and the background model built previously.

A considerable number of works have been done on background modeling, \ie building a proper representation of the background scene. Typical methods include single Gaussian distribution \cite{Wren2002Pfinder}, Mixture of Gaussian \cite{stauffer1999adaptive}, kernel density estimation \cite{Elgammal2000nonparametric,mittal2004motion}, block correlation \cite{matsuyama2000background}, codebook model \cite{kim2005real}, Hidden Markov model \cite{friedman1997image,rittscher2000probabilistic} and linear autoregressive models \cite{toyama1999wallflower,monnet2003background,zhong2003segmenting}.

Learning with sparsity has drawn a lot of attentions in recent machine learning and computer vision research \cite{wright2010sparse}, and several methods based on the sparse representation for background modeling have been developed. One pioneering work is the \emph{eigen backgrounds} model \cite{Oliver2000bayesian}, where the principle component analysis (PCA) is performed on a training sequence. When a new frame is arrived, it is projected onto the subspace spanned by the principle components, and the residues indicate the presence of new objects. An alternative approach that can operate sequentially is the sparse signal recovery \cite{cevher2008sparse,huang2009learning,mairal2010network}. Background subtraction is formulated as a regression problem with the assumption that a new-coming frame should be sparsely represented by a linear combination of preceding frames except for foreground parts. These models capture the correlation between video frames. Thus, they can naturally handle the global variations in the background such as illumination change and dynamic textures.

Background subtraction methods mentioned above rarely consider the scenario where the objects appear at the start and continuously present in the scene (\ie the training sequence is not available). Few literatures consider the problem of background initialization \cite{gutchess2001background,wang2006novel}. Most of them seek a stable interval, inside which the intensity is relatively smooth for each pixel independently. Pixels during such intervals are regarded as background, and the background scene is estimated from these intervals. The validity of this approach relies on the assumption of static background. Thus, it is limited when processing dynamic background or videos captured by a moving camera.

\section{Contiguous Outlier Detection in the Low-Rank Representation}\label{section_cod}

In this section, we focus on the problem of detecting contiguous outliers in the low-rank representation. We first consider the case without camera motion. We will discuss the scenarios with moving cameras in Section \ref{section_moving}.

\subsection{Notations}

In this paper, we use following notations. $I_j \in \mathbb{R}^m$ denotes the $j$-th frame of a video sequence, which is written as a column vector consisting of $m$ pixels. The $i$-th pixel in the $j$-th frame is denoted as $ij$. $D = [I_1,\cdots,I_n] \in \mathbb{R}^{m \times n}$ is a matrix representing all $n$ frames of a sequence. $B \in \mathbb{R}^{m \times n}$ is a matrix with the same size of $D$, which denotes the underlying background images. $S \in \{0,1\}^{m \times n}$ is a binary matrix denoting the foreground support:
\begin{align}
S_{ij} =  \left\{ {
\begin{array}{ll}
   {0, \hspace{1mm}} & \mbox{if $ij$ is background} \\
   {1, \hspace{1mm}} & \mbox{if $ij$ is foreground}
\end{array}} \right.
\end{align}
We use $\mathcal{P}_{S}(X)$ to represent the orthogonal projection of a matrix $X$ onto the linear space of matrices supported by $S$:
\begin{align}\label{Eq_projection}
\mathcal{P}_{S}(X)(i,j) =  \left\{ {
\begin{array}{ll}
   {0, \hspace{1mm}} &{{\rm if} \hspace{1mm} S_{ij} = 0 } \\
   {X_{ij},\hspace{1mm}} &{{\rm if} \hspace{1mm} S_{ij} = 1 }
\end{array}} \right.
\end{align}
and $\mathcal{P}_{S^{\perp}}(X)$ be its complementary projection, \ie
$\mathcal{P}_{S}(X) + \mathcal{P}_{S^{\perp}}(X) = X$.

Four norms of a matrix are used throughout this paper. $\|X\|_0$ denotes the $\ell_0$-norm, which counts the number of nonzero entries. $\|X\|_1=\sum_{ij}{|X_{ij}|}$ denotes the $\ell_1$-norm. $\|X\|_F=\sqrt{\sum_{ij}{X_{ij}^2}}$ is the Frobenius norm. $\|X\|_*$ means the nuclear norm, \ie sum of singular values.

\subsection{Formulation}\label{section_formulation}

Given a sequence $D$, our objective is to estimate the foreground support $S$ as well as the underlying background images $B$. To make the problem well-posed, we have following models to describe the foreground, the background and the formation of observed signal:

\textbf{Background model:} The background intensity should be unchanged over the sequence except for variations arising from illumination change or periodical motion of dynamic textures\footnote{Background motion caused by moving cameras will be considered in Section \ref{section_moving}}. Thus, background images are linearly correlated with each other, forming a low-rank matrix $B$. Besides the low-rank property, we don't make any additional assumption on the background scene. Thus, we only impose the following constraint on $B$:
\begin{align}\label{Eq_modelB}
{\rm rank}(B) \leq K,
\end{align}
where $K$ is a constant to be predefined. Intrinsically, $K$ constrains the complexity of the background model. We will discuss more on this parameter in Section \ref{section_simu}.

\textbf{Foreground model:} The foreground is defined as any object that moves differently from the background. Foreground motion gives intensity changes that can not be fitted into the low-rank model of background. Thus, they can be detected as outliers in the low-rank representation. Generally, we have a prior that foreground objects should be contiguous pieces with relatively small size. The binary states of entries in foreground support $S$ can be naturally modeled by a Markov Random Field \cite{geman1984stochastic,li2009markov}. Consider a graph $\mathcal{G}=(\mathcal{V},\mathcal{E})$, where $\mathcal{V}$ is the set of vertices denoting all $m \times n$ pixels in the sequence and $E$ is the set of edges connecting spatially or temporally neighboring pixels. Then, the energy of $S$ is given by the Ising model \cite{li2009markov}:
\begin{align}\label{Eq_modelS}
\sum_{ij \in \mathcal{V}}{ u_{ij}(S_{ij})} + \sum_{(ij,kl) \in \mathcal{E}}{ \lambda_{ij,kl} |S_{ij}-S_{kl}| },
\end{align}
where $u_{ij}$ denotes the unary potential of $S_{ij}$ being $0$ or $1$, and the parameter $\lambda_{ij,kl}>0$ controls the strength of dependency between $S_{ij}$ and $S_{kl}$. To prefer $S_{ij}=0$ that indicates sparse foreground, we define the unary potential $u_{ij}$ as:
\begin{align}
u_{ij}(S_{ij}) = \left\{ {
\begin{array}{ll}
   {0,\hspace{1mm}} &{{\rm if} \hspace{1mm} S_{ij}=0 } \\
   {\lambda_{ij},\hspace{1mm}} &{{\rm if} \hspace{1mm} S_{ij}=1}
\end{array}}, \right.
\end{align}
where the parameter $\lambda_{ij}>0$ penalizes $S_{ij}=1$. For simplicity, we set $\lambda_{ij}$ and $\lambda_{ij,kl}$ as constants over all locations. That is, $\lambda_{ij}=\beta$ and $\lambda_{ij,kl}=\gamma$, where $\beta>0$ and $\gamma>0$ are positive constants. This means that we have no additional prior about the locations of objects.

\textbf{Signal model:} The signal model describes the formation of $D$, given $B$ and $S$. In the background region where $S_{ij}=0$, we assume that $D_{ij}=B_{ij}+\epsilon_{ij}$, where $\epsilon_{ij}$ denotes i.i.d. Gaussian noise. That is, $D_{ij} \sim \mathcal{N}(B_{ij},\sigma^2)$ with $\sigma^2$ being the variance of Gaussian noise. Thus, $B_{ij}$ should be the best fitting to $D_{ij}$ in the least-squares sense, when $S_{ij}=0$. In the foreground regions where $S_{ij}=1$, the background scene is occluded by the foreground. Thus, $D_{ij}$ equals the foreground intensity. Since we don't make any assumption about the foreground appearance, $D_{ij}$ is not constrained when $S_{ij}=1$.

Combining above three models, we propose to minimize the following energy to estimate $B$ and $S$:
\begin{align}\label{Eq_energy_S_0}
    \min_{B,S_{ij} \in \{0,1\}} \hspace{2mm} & \frac{1}{2} \sum_{ij:S_{ij}=0}{(D_{ij}-B_{ij})^2}
    + \beta \sum_{ij}{S_{ij}} + \gamma \hspace{-2mm} \sum_{(ij,kl) \in \mathcal{E}}{|S_{ij}-S_{kl}|}, \nonumber \\
    \rm{s.t.} \hspace{7mm} & {\rm rank}(B) \leq K.
\end{align}
This formulation says that the background images should form a low-rank matrix and fit the observed sequence in the least-squares sense except for foreground regions that are sparse and contiguous.

To make the energy minimization tractable, we relax the rank operator on $B$ with the nuclear norm. The nuclear norm has proven to be an effective convex surrogate of the rank operator \cite{recht2010guaranteed}. Moreover, it can help to avoid overfitting, which will be illustrated by experiments in Section \ref{section_simu_par}.

Writing (\ref{Eq_energy_S_0}) in its dual form and introducing matrix operators, we obtain the final form of the energy function:
\begin{align}\label{Eq_energy_S_1}
    \min_{B,S_{ij} \in \{0,1\}}{ \hspace{2mm} \frac{1}{2}\|\mathcal{P}_{S^{\perp}}(D-B) \|_F^2 + \alpha \hspace{0.5mm} \|B\|_*}
    + \beta \hspace{0.5mm} \|S\|_1 + \gamma \hspace{0.5mm} \| A \hspace{0.5mm} {\rm vec}(S) \|_1.
\end{align}
Here, $A$ is the node-edge incidence matrix of $\mathcal{G}$, and $\alpha>0$ is a parameter associated with $K$, which controls the complexity of the background model. Proper choice of $\alpha$, $\beta$ and $\gamma$ will be discussed in details in Section \ref{section_par}.

\subsection{Algorithm}\label{section_alg}

The objective function defined in (\ref{Eq_energy_S_1}) is non-convex and it includes both continuous and discrete variables. Joint optimization over $B$ and $S$ is extremely difficult. Hence, we adopt an alternating algorithm that separates the energy minimization over $B$ and $S$ into two steps. $B$-step is a convex optimization problem and $S$-step is a combinatorial optimization problem. It turns out that the optimal solutions of $B$-step and $S$-step can be computed efficiently.

\subsubsection{Estimation of the low-rank matrix $B$}

Given an estimate of the support $\hat S$, the minimization in (\ref{Eq_energy_S_1}) over $B$ turns out to be the matrix completion problem \cite{Mazumder2010spectral}:
\begin{align}\label{Eq_completion}
\min_{B}{\hspace{2mm} \frac{1}{2} \|\mathcal{P}_{\hat S^{\perp}}(D-B)\|_F^2 + \alpha \|B\|_* }.
\end{align}
This is to learn a low-rank matrix from partial observations. The optimal $B$ in (\ref{Eq_completion}) can be computed efficiently by the SOFT-IMPUTE algorithm \cite{Mazumder2010spectral}, which makes use of the following Lemma \cite{cai2010singular}:

\begin{Lem}\label{softsvd}
Given a matrix $Z$, the solution to the optimization problem
\begin{equation}\label{NuclearNormRularization}
    \min_{X} \frac{1}{2}||Z-X||^2_F +\alpha ||X||_*
\end{equation}
is given by $\hat{X}= \Theta_{\alpha}(Z)$, where $\Theta_{\alpha}$ means the singular value thresholding:
\begin{equation}\label{softshrinksolution}
\Theta_{\alpha}(Z)=U\Sigma_{\alpha}V^T.
\end{equation}
Here, $\Sigma_{\alpha}=\mbox{diag}[(d_1-\alpha)_+, \dots, (d_r-\alpha)_+]$, $U\Sigma V^T$ is the SVD of $Z$, $\Sigma=\mbox{diag}[d_1,\dots,d_r]$ and $t_+ = {\rm max}(t,0)$.
\end{Lem}

Rewriting (\ref{Eq_completion}), we have
\begin{equation}\label{Eq_completion1}
\begin{aligned}
&\min_{B}{\hspace{2mm} \frac{1}{2} \|\mathcal{P}_{\hat
S^{\perp}}(D-B)\|_F^2 + \alpha \|B\|_* }\\
=&\min_{B}{\hspace{2mm} \frac{1}{2} \|[\mathcal{P}_{\hat
S^{\perp}}(D)+\mathcal{P}_{\hat S}(B)]-B\|_F^2 + \alpha \|B\|_* }.
\end{aligned}
\end{equation}
Using Lemma 1, the optimal solution to (\ref{Eq_completion}) can be obtained
by iteratively using:
\begin{align}\label{Eq_SoftImpute}
\hat B \leftarrow \Theta_{\alpha}( \mathcal{P}_{\hat S^{\perp}}(D) + \mathcal{P}_{\hat S}({\hat B})).
\end{align}
with arbitrarily initialized $\hat{B}$. Please refer to \cite{Mazumder2010spectral} for the details of SOFT-IMPUTE and the proof of its convergence.

\subsubsection{Estimation of the outlier support $S$}\label{section_graphcuts}

Next, we investigate how to minimize the energy in (\ref{Eq_energy_S_1}) over $S$ given the low-rank matrix $\hat B$. Noticing that $S_{ij} \in \{0,1\}$, the energy can be rewritten as follows:
\begin{align}\label{Eq_S}
&\frac{1}{2}\|\mathcal{P}_{S^{\perp}}(D-{\hat B}) \|_F^2
+ \beta \hspace{0.5mm} \|S\|_1 + \gamma \hspace{0.5mm} \| A \hspace{0.5mm} {\rm vec}(S) \|_1 \nonumber \\
=&\frac{1}{2} \sum_{ij}{ (D_{ij}-{\hat B}_{ij})^2 (1-S_{ij})} + \beta \sum_{ij} S_{ij} + \gamma \hspace{0.5mm} \| A \hspace{0.5mm} {\rm vec}(S) \|_1 \nonumber \\
=& \sum_{ij}{ (\beta -
\frac{1}{2}(D_{ij}-{\hat B}_{ij})^2)S_{ij}} + \gamma \| A
\hspace{0.5mm} {\rm vec}(S) \|_1 + \mathcal{C},
\end{align}
where $\mathcal{C}=\frac{1}{2}\sum_{ij}{(D_{ij}-{\hat B}_{ij})^2}$ is a constant when $\hat B$ is fixed. Above energy is in the standard form of the first-order MRFs with binary labels, which can be solved exactly using graph cuts \cite{Boykov01,kolmogorov2004energy}.

Ideally, both spatial and temporal smoothness can be imposed by connecting all pairs of nodes in $\mathcal{G}$ which correspond to all pairs of spatially or temporally neighboring pixels in the sequence. However, this will make $\mathcal{G}$ extremely large and difficult to solve. In implementation, we only connect spatial neighbors. Thus, $\mathcal{G}$ can be separated into subgraphs of single images, and the graph cuts can be operated for each image separately. This dramatically reduces the computational cost. Based on our observation, the spatial smoothness is sufficient to obtain satisfactory results.

\subsubsection{Parameter tuning}\label{section_par}

The parameter $\alpha$ in (\ref{Eq_energy_S_1}) controls the complexity of the background model. A larger $\alpha$ gives a $\hat B$ with smaller nuclear norm. In our algorithm, we first give a rough estimate to the rank of the background model, \ie $K$ in (\ref{Eq_energy_S_0}). Then, we start from a large $\alpha$. After each run of SOFT-IMPUTE, if ${\rm rank}(\hat B) \leq K$, we reduce $\alpha$ by a factor $\eta_1 < 1$ and repeat SOFT-IMPUTE until ${\rm rank}(\hat B) > K$. Using \emph{warm-start}, this sequential optimization is efficient \cite{Mazumder2010spectral}. In our implementation, we initialize $\alpha$ to be the second largest singular value of $D$, and $\eta_1=1/\sqrt{2}$.

The parameter $\beta$ in (\ref{Eq_energy_S_1}) controls the sparsity of the outlier support. From (\ref{Eq_S}) we can see that $\hat S_{ij}$ is more likely to be 1 if $\frac{1}{2}(D_{ij}-{\hat B}_{ij})^2 > \beta$. Thus the choice of $\beta$ should depend on the noise level in images. Typically we set $\beta = 4.5{\hat\sigma}^2$, where ${\hat\sigma}^2$ is estimated online by the variance of $D_{ij}-{\hat B}_{ij}$. Since the estimation of $\hat B$ and $\hat\sigma$ is biased at the beginning iterations, we propose to start our algorithm with a relatively large $\beta$, and then reduce $\beta$ by a factor $\eta_2=0.5$ after each iteration until $\beta$ reaches $4.5{\hat\sigma}^2$. In other words, we tolerate more error in model fitting at the beginning, since the model itself is not accurate enough. With the model estimation getting better and better, we decrease the threshold and declare more and more outliers.

In conclusion, we only have two parameters to choose, \ie $K$ and $\gamma$. In Section \ref{section_simu_par} we will show that DECOLOR performs stably if $K$ and $\gamma$ are in proper ranges. In all our experiments, we let $K=\sqrt{n}$, and $\gamma=\beta$ and $5\beta$ for simulation and real sequences, respectively.

\subsubsection{Convergence}

For fixed parameters, we always minimize a single lower-bounded energy in each step. The convergence property of SOFT-IMPUTE has been proved in \cite{Mazumder2010spectral}. Therefore, the algorithm must converge to a local minimum. For adaptive parameter tuning, our strategy guarantees that the coefficients ($\alpha,\beta,\gamma$) keep decreasing for each change. Thus, the energy in (\ref{Eq_energy_S_1}) decreases monotonically with the algorithm running. Furthermore, we can manually set lower bounds for both $\alpha$ and $\beta$ to stop the iteration. Empirically, DECOLOR converges in about 20 iterations for a convergence precision of $10^{-5}$.

\subsection{Relation to Other Methods}

\subsubsection{Robust Principle Component Analysis}\label{section_pcp}

RPCA has drawn a lot of attention in computer vision \cite{de2003framework,ke2005robust}. Recently, the seminal work \cite{Candes09} shows that, under some mild conditions, the low-rank model can be recovered from unknown corruption patterns via a convex program named Principal Component Pursuit (PCP). The examples in \cite{Candes09} demonstrate the superior performance of PCP compared with previous methods of RPCA and its promising potential for background subtraction.

As discussed in \cite{Candes09}, PCP can be regarded as a special case of the following decomposition model:
\begin{align}\label{Eq_decomposition}
D = B + E + \epsilon,
\end{align}
where $B$ is a low-rank matrix, $E$ represents the intensity shift caused by outliers and $\epsilon$ denotes the Gaussian noise. PCP only seeks for the low-rank and sparse decomposition $D=B+E$ without considering $\epsilon$. Recently, Stable Principle Component Pursuit (SPCP) has been proposed \cite{zhou2010stable}. It extends PCP \cite{Candes09} to handle both sparse gross errors and small entrywise noises. It tries to find the decomposition by minimizing the following energy:
\begin{align}\label{Eq_energy_E_0}
\min_{B,E}\hspace{2mm} \frac{1}{2}\| D-B-E \|_F^2 + \alpha \hspace{0.5mm} \mbox{rank}(B) + \beta \hspace{0.5mm} \| E \|_0.
\end{align}
To make the optimization tractable, (\ref{Eq_energy_E_0}) is relaxed by replacing ${\rm rank}(B)$ with $\|B\|_*$ and $\|E\|_0$ with $\|E\|_1$ in PCP or SPCP. Thus, the problem turns out to be convex and can be solved efficiently via convex optimization. However, the $\ell_1$ relaxation requires that the distribution of corruption should be sparse and random enough, which is not generally true in the problem of motion segmentation. Experiments in Section \ref{section_exp} show that PCP is not robust enough when the moving objects take up relatively large and contiguous space of the sequence.

Next, we shall explain the relation between our formulation in (\ref{Eq_energy_S_1}) and the formulation in (\ref{Eq_energy_E_0}). It is easy to see that, as long as $E_{ij} \neq 0$, we must have $E_{ij}=D_{ij}-B_{ij}$ to minimize (\ref{Eq_energy_E_0}). Thus, (\ref{Eq_energy_E_0}) has the same minimizer with the following energy:
\begin{align}\label{Eq_energy_E_00}
\min_{B,E}\hspace{2mm} \frac{1}{2} \sum_{ij:E_{ij}=0}{\hspace{-3mm}(D_{ij}-B_{ij})^2} + \alpha
\hspace{0.5mm}\mbox{rank}(B) + \beta \hspace{0.5mm} \| E \|_0.
\end{align}
The first term in (\ref{Eq_energy_E_00}) can be rewritten as $\frac{1}{2}\|\mathcal{P}_{S^{\perp}}(D-B) \|_F^2$. Noticing that $\|E\|_0=\|S\|_1$ and replacing ${\rm rank}(B)$ with $\|B\|_*$, (\ref{Eq_energy_E_00}) can be finally rewritten as (\ref{Eq_energy_S_1}) if the last smoothness term in (\ref{Eq_energy_S_1}) is ignored.

Thus, DECOLOR can be regarded as a special form of RPCA, where the $\ell_0$-penalty on $E$ is not relaxed and the problem in (\ref{Eq_energy_E_0}) is converted to the optimization over $S$ in (\ref{Eq_energy_S_0}). One recent work \cite{She10} has shown that the $\ell_0$-penalty works effectively for outlier detection in regression, while the $\ell_1$-penalty does not. As pointed out in \cite{She10}, the theoretical reason for the unsatisfactory performance of the $\ell_1$-penalty is that the irrepresentable condition \cite{zhao2006model} is often not satisfied in the outlier detection problem. In order to go beyond the $\ell_1$-penalty, non-convex penalties have been explored in recent literature \cite{She10,mazumder2011sparsenet}. Compared with the $\ell_1$-norm, non-convex penalties give an estimation with less bias but higher variance. Thus, these non-convex penalties are superior to the $\ell_1$-penalty when the signal-noise-ratio (SNR) is relatively high \cite{mazumder2011sparsenet}. For natural video analysis, it is the case.

In summary, both PCP \cite{Candes09} and DECOLOR aim to recover a low-rank model from corrupted data. PCP \cite{Candes09,zhou2010stable} uses the convex relaxation by replacing ${\rm rank}(B)$ with $\|B\|_*$ and $\|E\|_0$ with $\|E\|_1$. DECOLOR only relaxes the rank penalty and keeps the $\ell_0$-penalty on $E$ to preserve the robustness to outliers. Moreover, DECOLOR estimates the outlier support $S$ explicitly by formulating the problem as the energy minimization over $S$, and models the continuity prior on $S$ using MRFs to improve the accuracy of detecting contiguous outliers.

\subsubsection{Sparse signal recovery}\label{section_regression}

With the success of compressive sensing \cite{donoho2006compressed}, sparse signal recovery has become a popular framework to deal with various problems in machine learning and signal processing \cite{wright2010sparse,Zhou10,Peng10}. To make use of structural information about nonzero patterns of variables, the structured-sparsity is defined in recent works \cite{yuan2006model,zhao2009composite}, and several algorithms have been developed and applied successfully on background subtraction, such as Lattice Matching Pursuit (LaMP) \cite{cevher2008sparse}, Dynamic Group Sparsity (DGS) recovery \cite{huang2009learning} and Proximal Operator using Network Flow (ProxFlow) \cite{mairal2010network}.

In sparse signal recovery for background subtraction, a testing image $y \in \mathbb{R}^m$ is modeled as a sparse linear combination of $n$ previous frames $\Phi \in \mathbb{R}^{m \times n}$ plus a sparse error term $e \in \mathbb{R}^m$ and a Gaussian noise term $\epsilon\in\mathbb{R}^m$:
\begin{align}\label{Eq_regression}
y = \Phi w + e + \epsilon.
\end{align}
$w \in \mathbb{R}^n$ is the coefficient vector. The first term $\Phi w$ accounts for the background shared between $y$ and $\Phi$, while the sparse error $e$ corresponds to the foreground in $y$. Thus, background subtraction can be achieved by recovering $w$ and $e$. Taking the latest algorithm ProxFlow \cite{mairal2010network} as an example, the following optimization is proposed:
\begin{align}\label{Eq_ProxFlow}
\min_{w,e}{\frac{1}{2}\|y - \Phi w-e\|_2^2+\lambda_1\|w\|_1+\lambda_2\|e\|_{\ell_1/\ell_{\infty}}},
\end{align}
where $\|\cdot\|_{\ell_1/\ell_{\infty}}$ is a norm to induce the group-sparsity. Please refer to \cite{mairal2010network} for the detailed definition. In short, the $\ell_1/\ell_{\infty}$-norm is used as a structured regularizer to encode the prior that nonzero entries of $e$ should be in a group structure, where the groups are specified to be all overlapping $3\times3$-squares on the image plane \cite{mairal2010network}.

In (\ref{Eq_regression}), $\Phi$ can be interpreted as a basis matrix for linear regression to fit the testing image $y$. In the literatures mentioned above, $\Phi$ is fixed to be the training sequence \cite{mairal2010network} or previous frames on which background subtraction has been performed \cite{huang2009learning}. Then, the only task is to recover the sparse coefficients.

In our problem formulation, $\Phi$ is unknown. DECOLOR learns the bases and coefficients for a batch of test images simultaneously. To illustrate this, we can rewrite (\ref{Eq_decomposition}) as:
\begin{align}
D = \Phi W + E + \epsilon,
\end{align}
where the original low-rank $B$ is factorized as a product of a basis matrix $\Phi \in \mathbb{R}^{m\times r}$ and a coefficient matrix $W \in \mathbb{R}^{r \times n}$ with $r$ being the rank of $B$.

In summary, LaMP, DGS and ProxFlow aim to detect new objects in a new testing image given a training sequence not containing such objects. The problem is formulated as linear regression with fixed bases. DECOLOR aims to segment moving objects from a short sequence during which the objects continuously appear, which is a more challenging problem. To this end, DECOLOR estimates the foreground and background jointly by outlier detection during matrix learning. The difference between DECOLOR and sparse signal recovery will be further demonstrated using experiments on real sequences in Section \ref{section_exp_regression}.

\section{Extension to Moving Background}\label{section_moving}

Above derivation is based on the assumption that the videos are captured by static cameras. In this section, we introduce domain transformations into our model to compensate for the background motion caused by moving cameras. Here we use the 2D parametric transforms \cite{szeliski2010computer} to model the translation, rotation and planar deformation of the background.

Let $D_j \circ {\tau_j}$ denote the $j$-th frame after the transformation parameterized by vector $\tau_j \in {\mathbb R}^p$, where $p$ is the number of parameters of the motion model (\eg $p=6$ for the affine motion or $p=8$ for the projective motion). Then the proposed decomposition becomes $D\circ\tau = B + E + \epsilon$, where $D\circ\tau=[D_1\circ\tau_1,\cdots,D_n\circ\tau_n]$ and $\tau\in\mathbb{R}^{p \times n}$ is a vector comprising all $\tau_j$. A similar idea can be found in the recent work on batch image alignment \cite{Peng10}.

Next, we substitute $D$ in (\ref{Eq_energy_S_1}) with $D\circ\tau$ and estimate $\tau$ along with $B$, $S$ by iteratively minimizing:
\begin{align}\label{Eq_energy_final2}
    \min_{\tau,B,S}{ \hspace{2mm} \frac{1}{2}\|\mathcal{P}_{S^{\perp}}(D\circ\tau-B) \|_F^2
    + \alpha \hspace{0.5mm} \|B\|_*}
    + \beta \hspace{0.5mm} \|S\|_1 + \gamma \hspace{0.5mm} \| A \hspace{0.5mm} {\rm vec}(S) \|_1.
\end{align}

Now we investigate how to minimize the energy in (\ref{Eq_energy_final2}) over $\tau$, given $\hat B$ and $\hat S$:
\begin{align}\label{Eq_tau}
\hat\tau = \arg\min_{\tau} \hspace{2mm} \|\mathcal{P}_{\hat
S^{\perp}}(D\circ{\tau}-{\hat B}) \|_F^2.
\end{align}

Here we use the incremental refinement \cite{Peng10,szeliski2010computer} to solve this parametric motion estimation problem: at each iteration, we update $\hat\tau$ by a small increment $\Delta\tau$ and linearize $D\circ{\tau}$ as $D\circ{\hat\tau} + J_{\hat\tau}\Delta\tau$, where $J_{\hat\tau}$ denotes the Jacobian matrix $\frac{{\partial D}}{{\partial \tau}}|_{\tau={\hat \tau}}$. Thus, $\tau$ can be updated in the following way:
\begin{align}\label{Eq_tauUpdate}
{\hat \tau} \leftarrow {\hat\tau} +
\arg\min_{\Delta\tau} \|\mathcal{P}_{\hat
S^{\perp}}(D\circ{{\hat\tau}} - {\hat B} +
J_{\hat\tau}\Delta\tau) \|_F^2.
\end{align}

The minimization over $\Delta\tau$ in (\ref{Eq_tauUpdate}) is a weighted least-squares problem, which has a closed-form solution.

In practice, the update of $\tau_1,\cdots,\tau_n$ can be done separately since the transformation is applied on each image individually. Thus the update of $\tau$ is efficient. To accelerate the convergence of DECOLOR, we initialize $\tau$ by roughly aligning each frame $D_j$ to the middle frame $D_{\frac{n}{2}}$ before the main loops of DECOLOR. The pre-alignment is done by the robust multiresolution method proposed in \cite{odobez1995robust}.

All steps of DECOLOR with adaptive parameter tuning are summarized in Algorithm \ref{Alg_overall}.

\begin{algorithm}[t]
    \caption{Moving Object Segmentation by DECOLOR}\label{Alg_overall}
    \begin{algorithmic}[1]
    \algsetup{linenodelimiter=.}
        \STATE {\bf Input:}$D = [I_1,\cdots,I_n] \in \mathbb{R}^{m \times n}$
        \STATE {\bf Initialize:} ${\hat\tau},{\hat B}\leftarrow D\circ{\hat\tau},{\hat S \leftarrow {\mathbf 0}},\alpha,\beta.$
        \REPEAT
              \STATE ${\hat \tau} \leftarrow {\hat \tau} +
                        \arg\min\limits_{\Delta\tau} \|\mathcal{P}_{\hat
                        S^{\perp}}(D\circ{{\hat \tau}} - {\hat B} + J_{\hat\tau}\Delta\tau) \|_2^2$;
            \vspace{1.5mm}
            \REPEAT
                    \STATE ${\hat B} \leftarrow \Theta_{\alpha}(\mathcal{P}_{\hat
                    S^{\perp}}(D\circ{\hat\tau}) + \mathcal{P}_{\hat S}({\hat B}))$;
                    \UNTIL{convergence}
            \IF{${\rm rank}({\hat B})\leq K$}
                \STATE $\alpha \leftarrow \eta_1\alpha$;
                \STATE {\bf go to} Step 5;
            \ENDIF
            \vspace{1.5mm}
            \STATE estimate $\hat\sigma$;
            \STATE $\beta \leftarrow \max{( \eta_2\beta,4.5{\hat\sigma}^2 )}$;
            \STATE
            {\footnotesize
            ${\hat S} \leftarrow \arg\min\limits_{S}{
            \sum\limits_{ij}{(\beta - \frac{1}{2}([D\circ{\hat\tau}]_{ij}-{\hat
            B}_{ij})^2)S_{ij}} + \gamma \| A \hspace{0.5mm} {\rm vec}(S) \|_1}$}
        \UNTIL{convergence}
        \STATE {\bf Output:} $\hat B$,$\hat S$,$\hat\tau$
    \end{algorithmic}
\end{algorithm}

\section{Experiments}\label{section_exp}

\subsection{Simulation}\label{section_simu}

In this section, we perform numerical experiments on synthesized data. We consider the situations with no background motion and mainly investigate whether DECOLOR can successfully separate the contiguous outliers from the low-rank model.

\begin{figure}[t]
\centering
\begin{minipage}[b]{.12\linewidth}
    \centerline{\includegraphics[width=1.0\linewidth,height=1\linewidth]{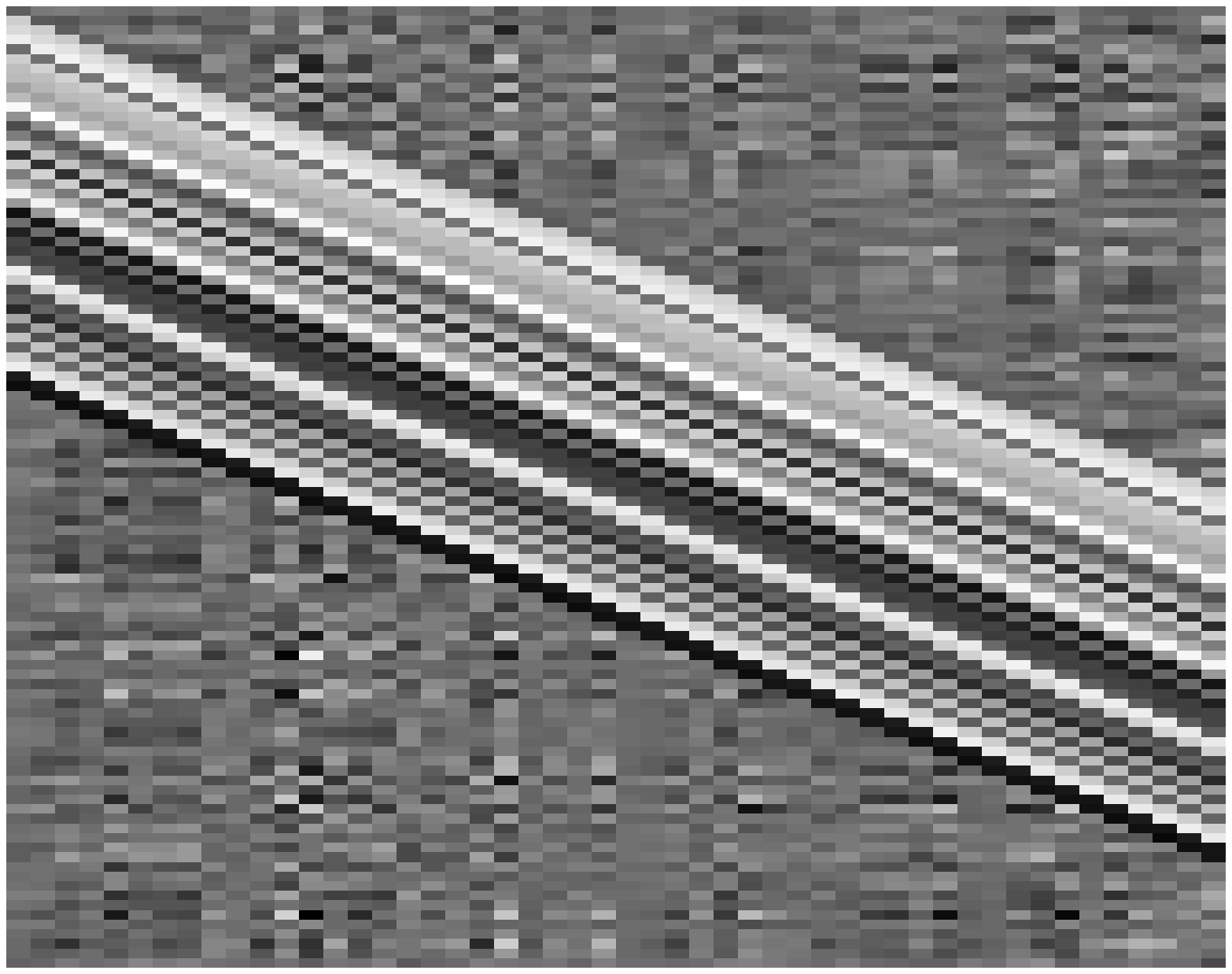}}
    \centerline{\small (a)Data}\medskip
\end{minipage}
\begin{minipage}[b]{.12\linewidth}
    \centerline{\includegraphics[width=1.0\linewidth,height=1\linewidth]{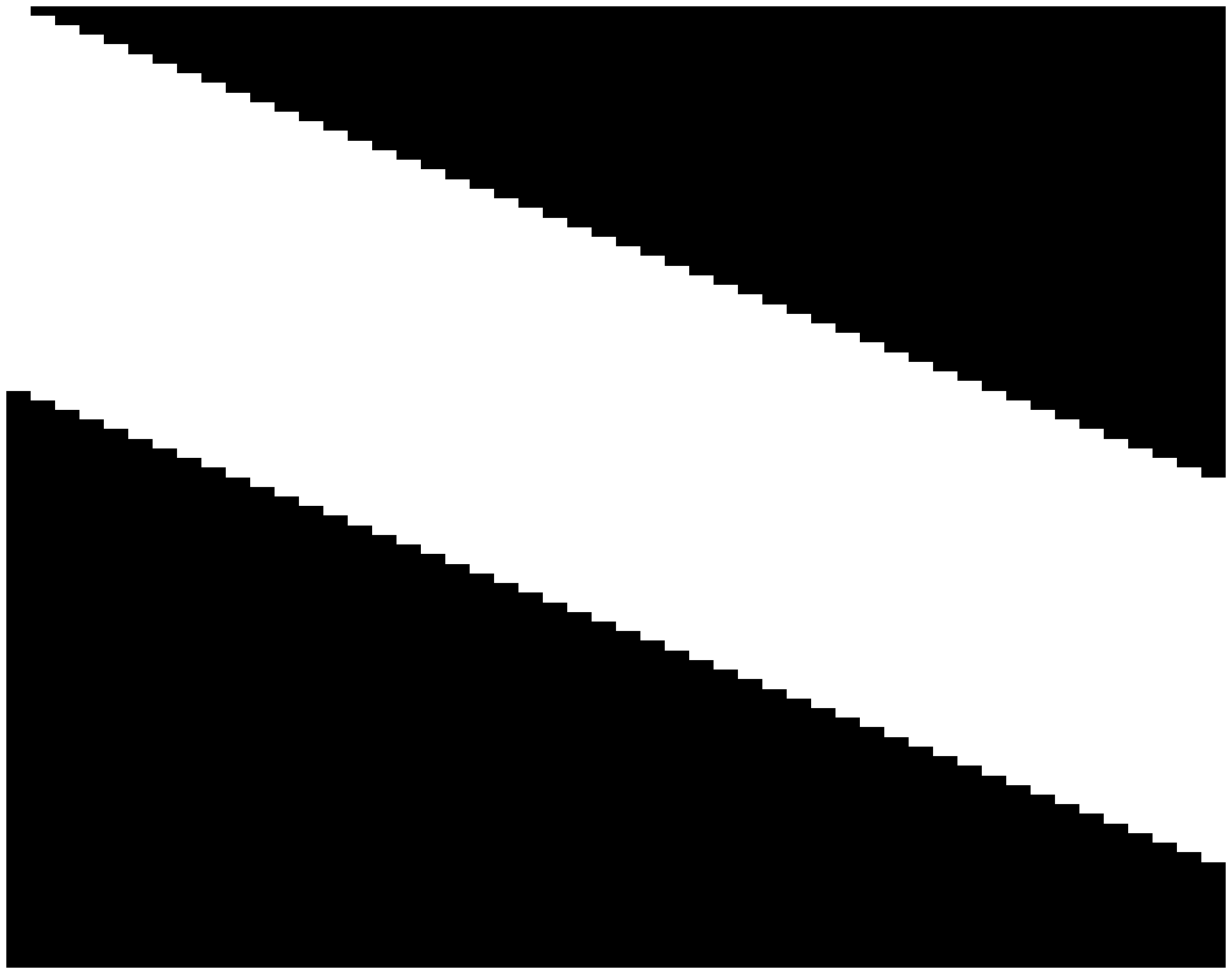}}
    \centerline{\includegraphics[width=1.0\linewidth,height=1\linewidth]{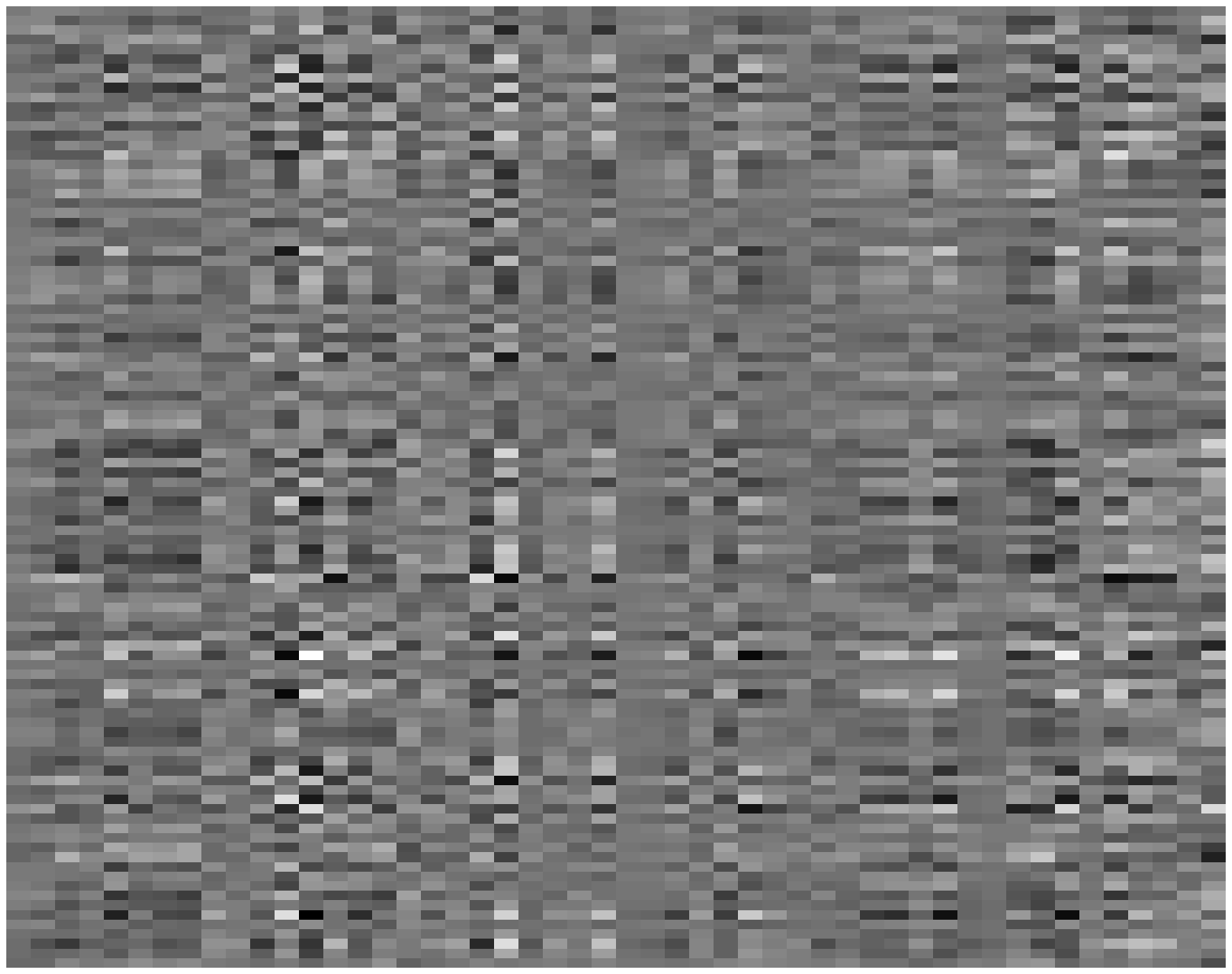}}
    \centerline{\small (b)Truth}\medskip
\end{minipage}
\begin{minipage}[b]{.12\linewidth}
    \centerline{\includegraphics[width=1.0\linewidth,height=1\linewidth]{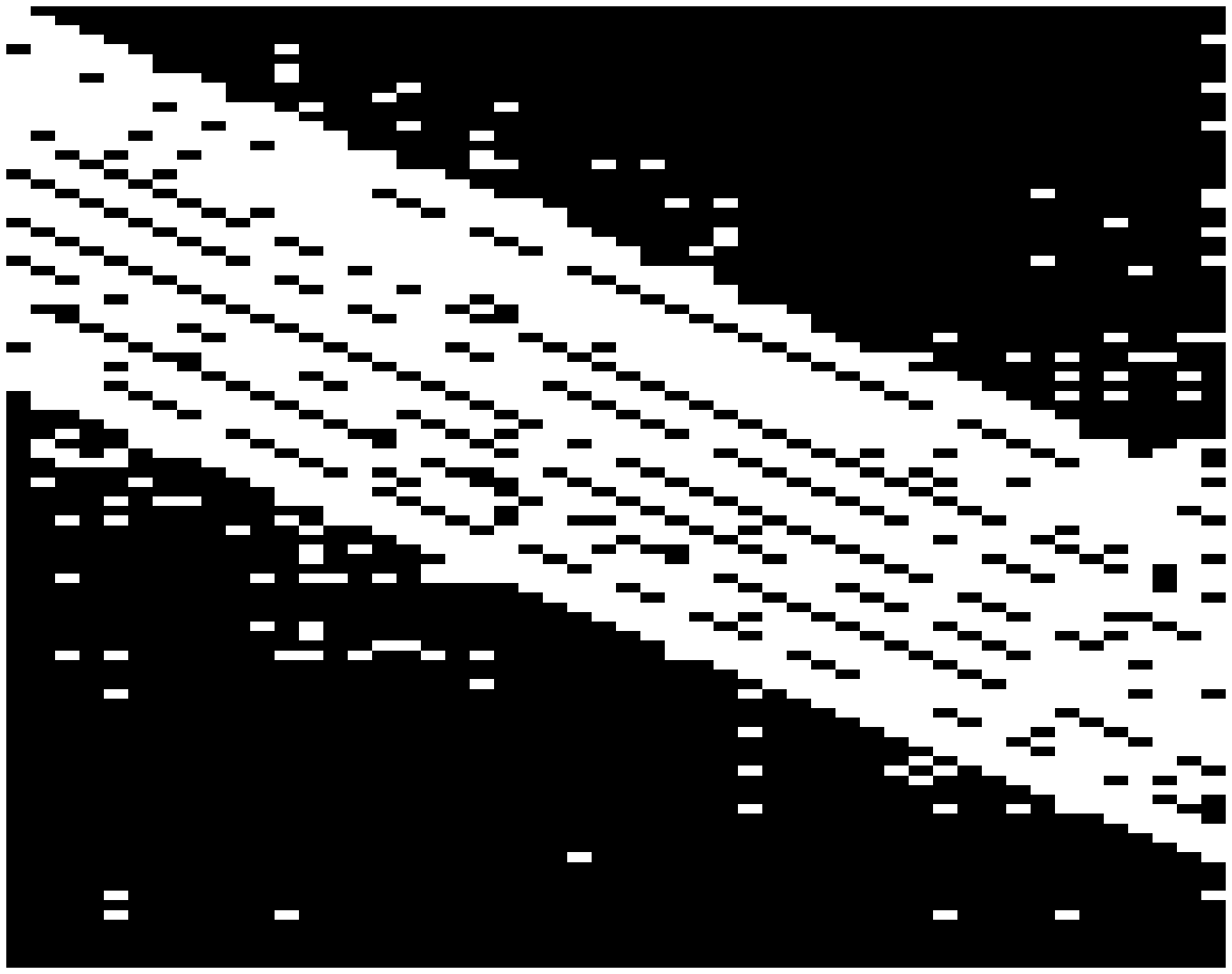}}
    \centerline{\includegraphics[width=1.0\linewidth,height=1\linewidth]{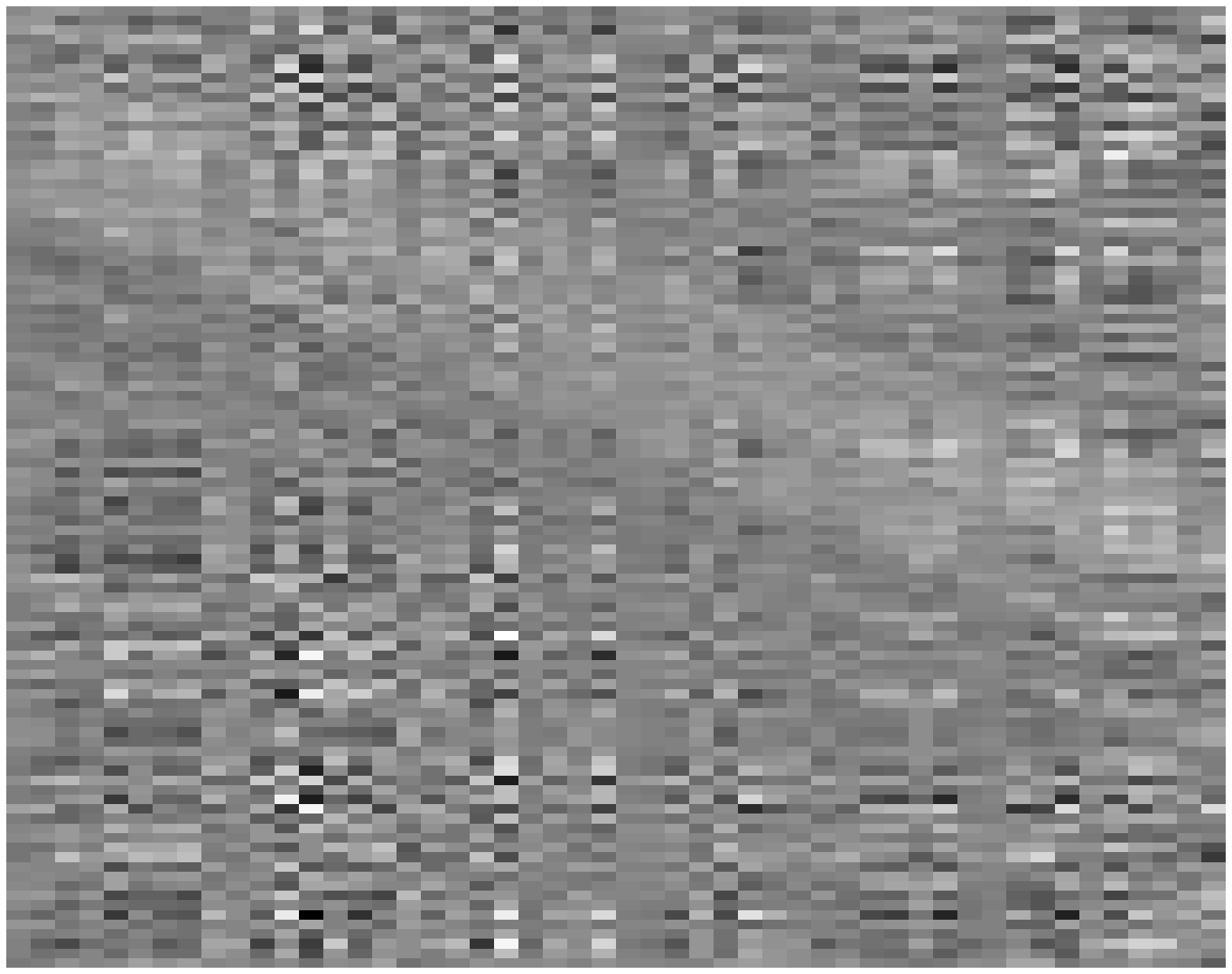}}
    \centerline{\small (c) PCP}\medskip
\end{minipage}
\begin{minipage}[b]{.12\linewidth}
    \centerline{\includegraphics[width=1.0\linewidth,height=1\linewidth]{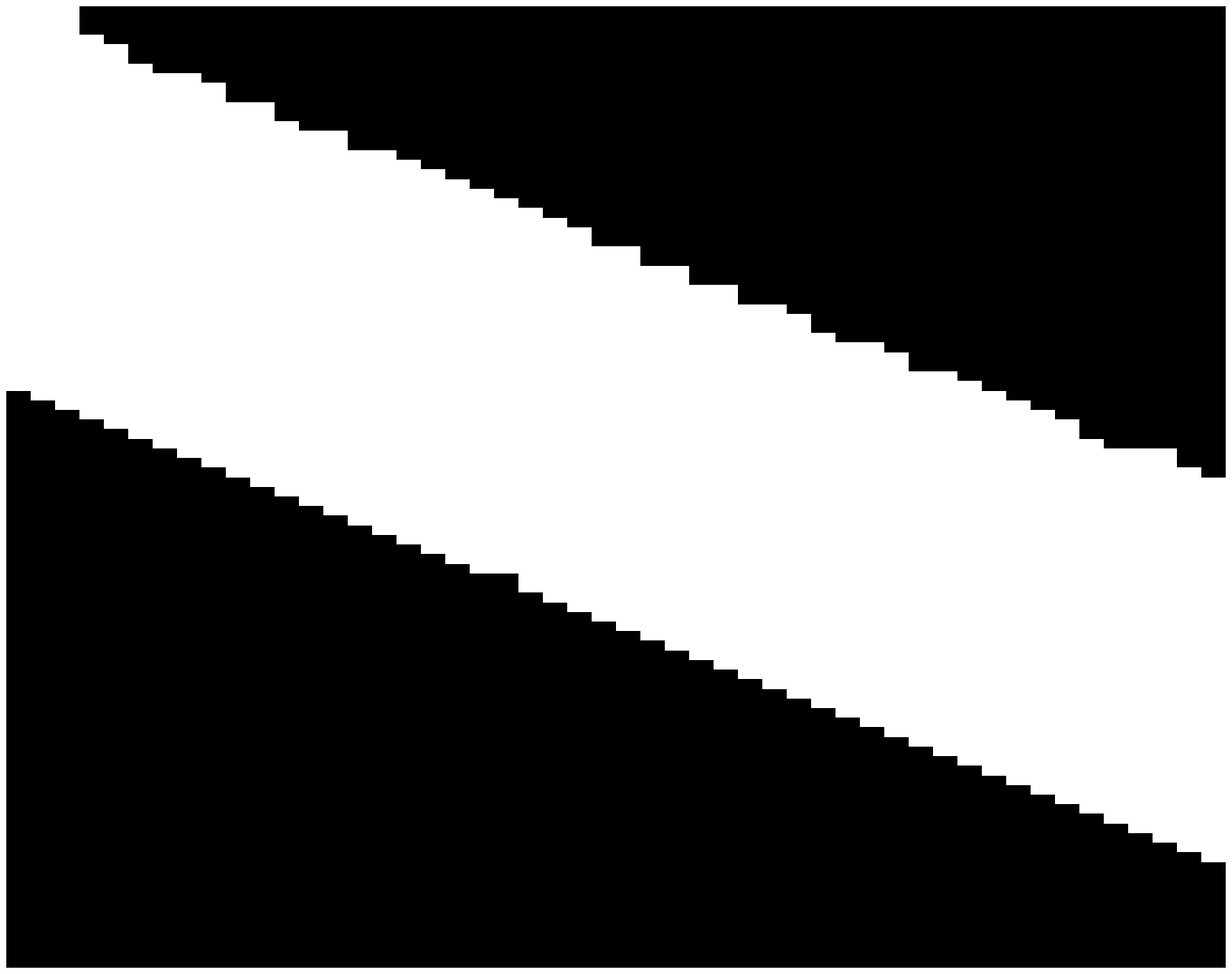}}
    \centerline{\includegraphics[width=1.0\linewidth,height=1\linewidth]{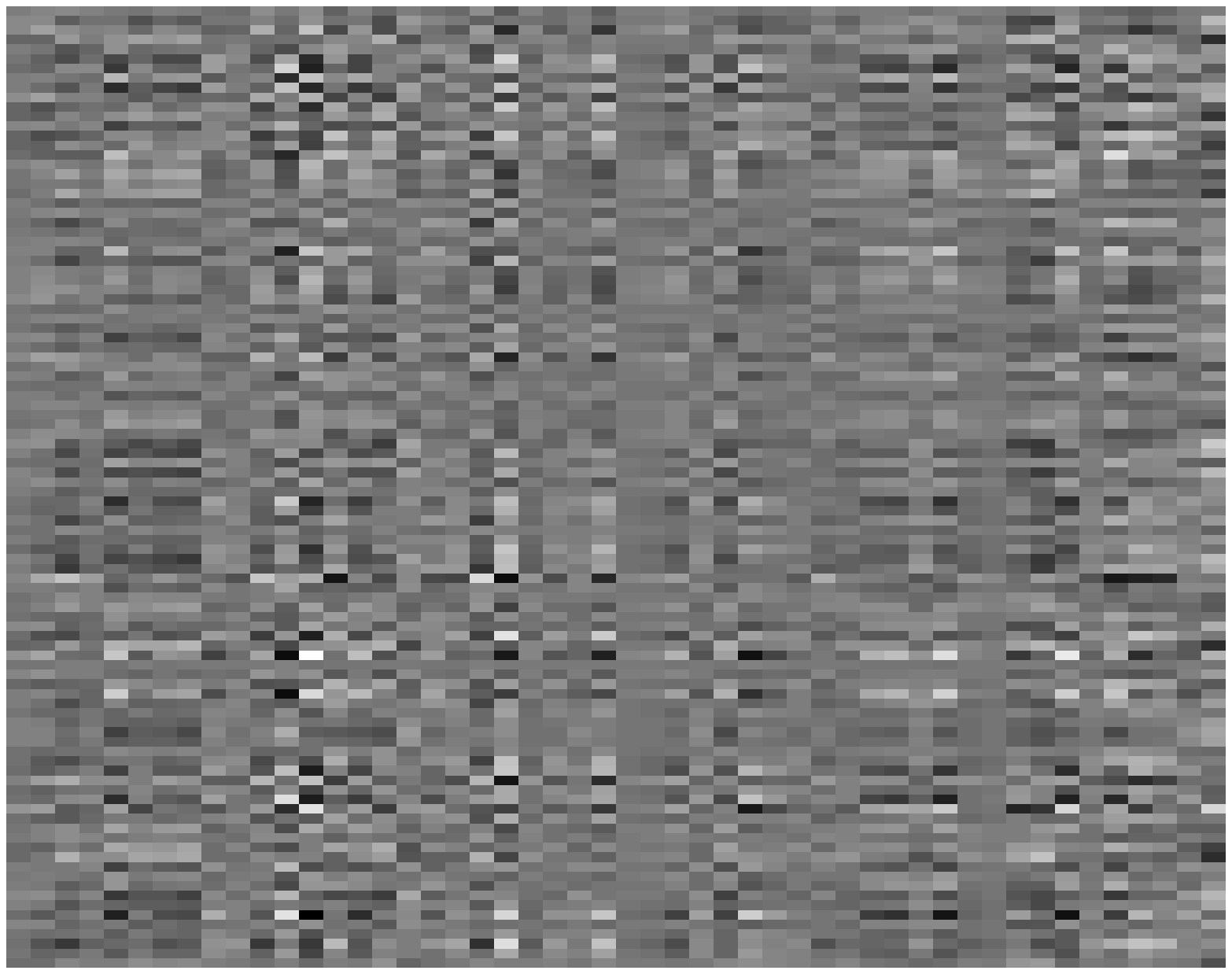}}
    \centerline{\small (d) DECOLOR}\medskip
\end{minipage}
\caption{\small \textbf{(a)} An example of synthesized data. Sequence $D\in\mathbb{R}^{100 \times 50}$ is a matrix composed of 50 frames of 1D images with 100 pixels per image. \textbf{(b)} The foreground support $S_0$ and underlying background images $B_0$. ${\rm rank}(B_0)=3$. $D$ is generated by adding a foreground object with width $W=40$ to each column of $B_0$, which moves downwards for 1 pixel per column. Also, i.i.d. Gaussian noise is added to each entry, and $\rm{SNR}=10$. {\textbf{(c)}} The results of PCP. The top panel is $\hat S$ and the bottom panel is $\hat B$. $\hat S$ of PCP is obtained by thresholding $|D_{ij}-\hat{B}_{ij}|$ with a threshold that gives the largest F-measure. Notice the artifacts in both $\hat S$ and $\hat B$ estimated by PCP. {\textbf{(d)}} The results of DECOLOR. Here $\hat S$ is directly output by DECOLOR without postprocessing. }\label{Fig_simuData}
\end{figure}

To better visualize the data, we use a simplified scenario: the video to be segmented is composed of 1D images. Thus, the image sequence and results can be displayed as 2D matrices. We generate the input $D$ by adding a foreground occlusion with support $S_0$ to a background matrix $B_0$. The background matrix $B_0$ with rank $r$ is generated as $B_0=UV^T$ where $U$ and $V$ are $m \times r$ and $n \times r$ matrices with entries independently sampled from a standard normal distribution. We choose $m=100$, $n=50$ and $r=3$ for all experiments. Then, an object with width $W$ is superposed on each column of $B_0$ and shifts downwards for 1 pixel per column. The intensity of this object is independently sampled from a uniform distribution $\mathcal{U}(-c,c)$, where $c$ is chosen to be the largest magnitude of entries in $B_0$. Also, we add i.i.d. Gaussian noise $\epsilon$ to $D$ with the corresponding signal-to-noise ratio (SNR) defined as:
\begin{align}
\rm{SNR} = \sqrt{\frac{\rm{var}(B_0)}{\rm{var}(\epsilon)}}.
\end{align}
Fig. \ref{Fig_simuData}(a) shows an example, where the moving foreground can be recognized as contiguous outliers superposed on a low-rank matrix. Our goal is to estimate $S_0$ and recover $B_0$ at the same time.

For quantitative evaluation, we measure the accuracy of outlier detection by comparing $\hat S$ with $S_0$. We regard it as a classification problem and evaluate the results using precision and recall, which are defined as:
\begin{align}
\rm{precision} = \frac{\rm{TP}}{\rm{TP+FP}}, \hspace{5mm} \rm{recall} = \frac{\rm{TP}}{\rm{TP+FN}},
\end{align}
where TP, FP, TN and FN mean the numbers of true positives, false positives, true negatives and false negatives, respectively. Precision and recall are widely used when the class distribution is skewed \cite{davis2006relationship}. For simplicity, instead of plotting precision/recall curves, we use a single measurement named F-measure that combines precision and recall:
\begin{align}
\mbox{F-measure} = 2\frac{\rm{precision}\cdot\rm{recall}}{\rm{precision}+\rm{recall}}.
\end{align}
The higher the F-measure is, the better the detection accuracy is. On our observation, PCP requires proper thresholding to generate a really sparse $\hat S$. For fair comparison, $\hat S$ of PCP is obtained by thresholding $|D_{ij} - \hat B_{ij}|$ with a threshold that gives the maximal F-measure. Furthermore, we measure the accuracy of low-rank recovery by calculating the difference between $\hat B$ and $B_0$. We use the Root Mean Square Error (RMSE) to measure the difference:
\begin{align}
RMSE = \frac{\|\hat B - B_0\|_F }{\|B_0\|_F}.
\end{align}

\subsubsection{Comparison to PCP}

\begin{figure*}
\centering
\begin{minipage}[b]{.3\linewidth}
 \centering
 \centerline{\includegraphics[width=1\linewidth]{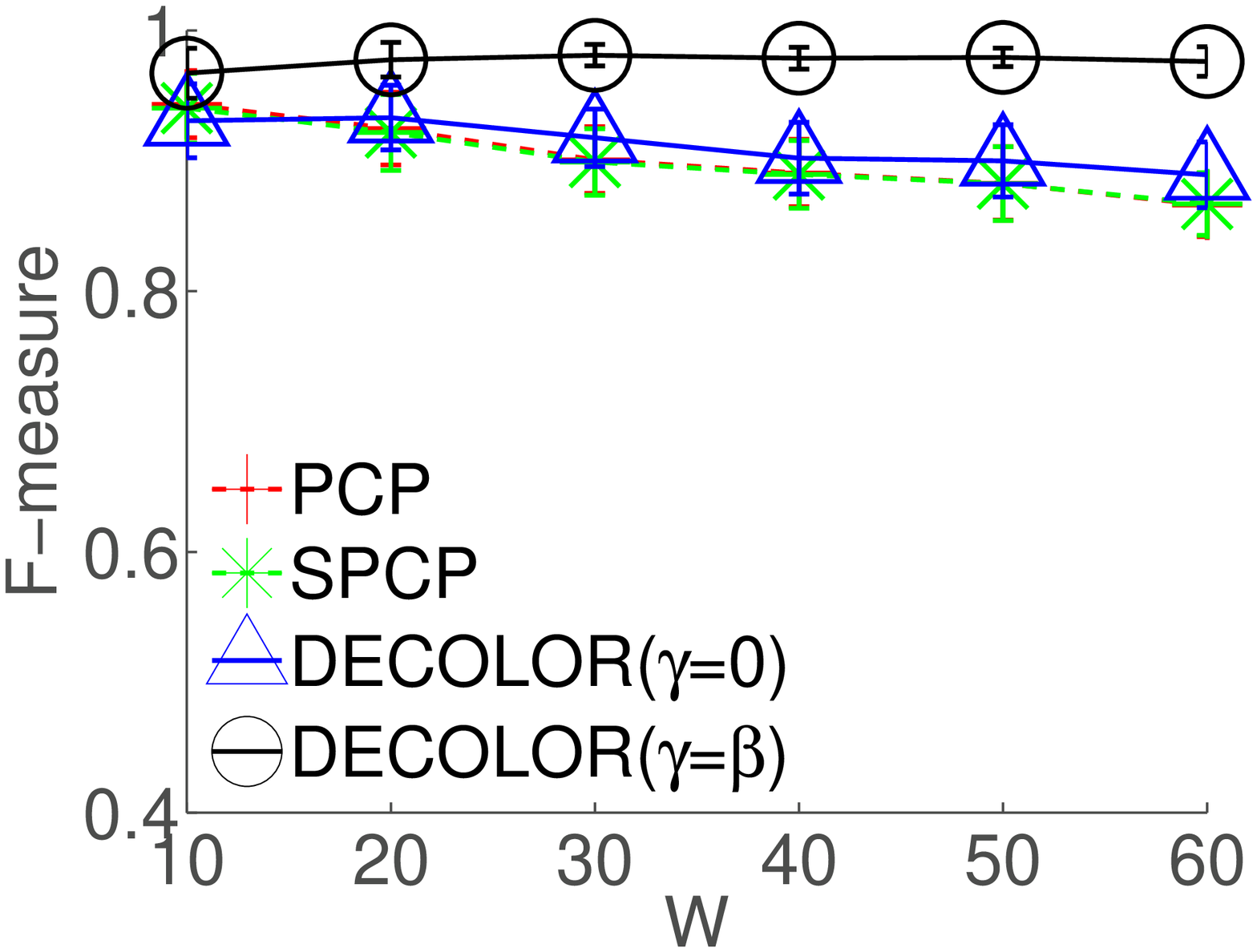}}
 \centerline{\includegraphics[width=1\linewidth]{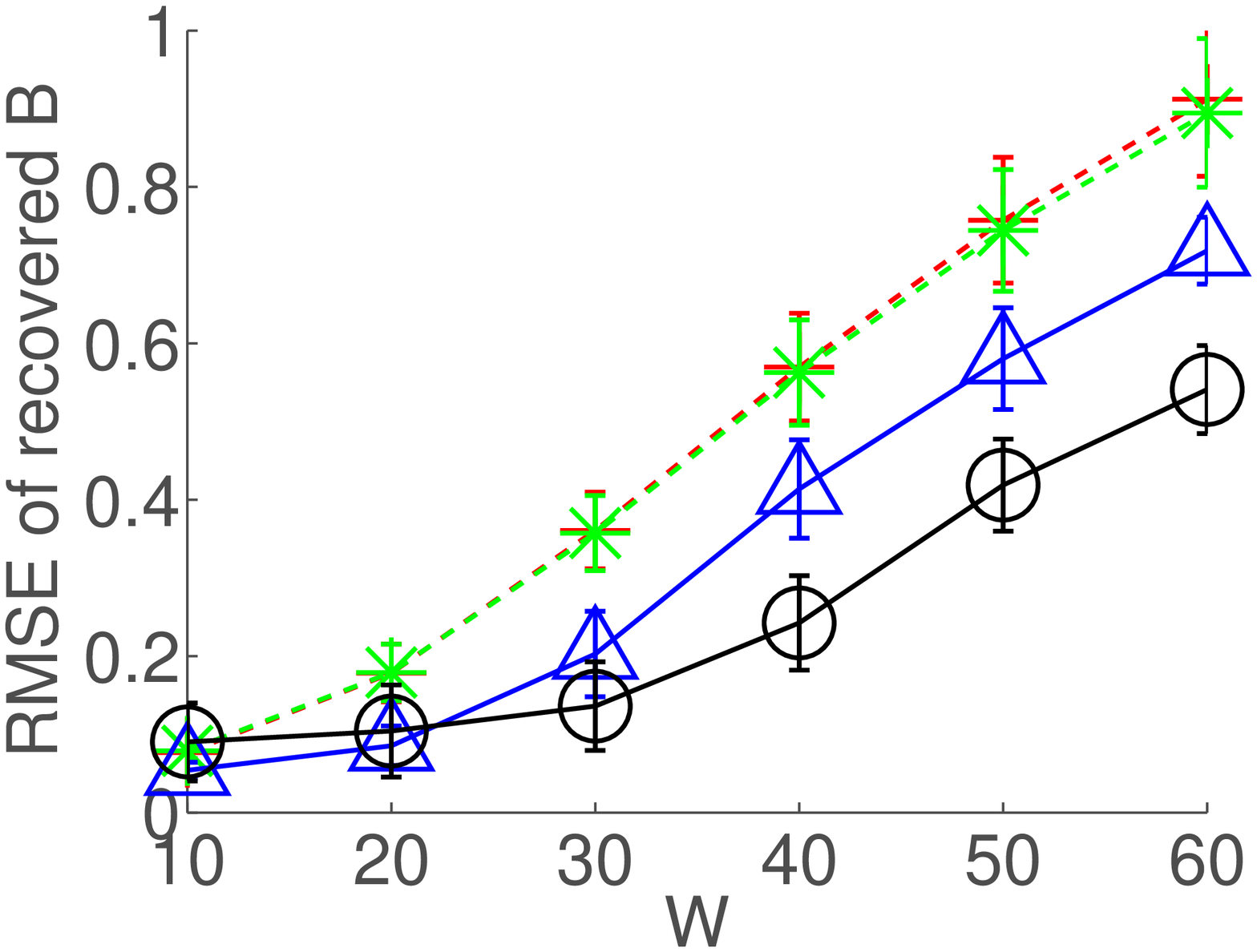}}
 \centerline{(a)}\medskip
\end{minipage}
\begin{minipage}[b]{.29\linewidth}
 \centering
 \centerline{\includegraphics[width=1\linewidth]{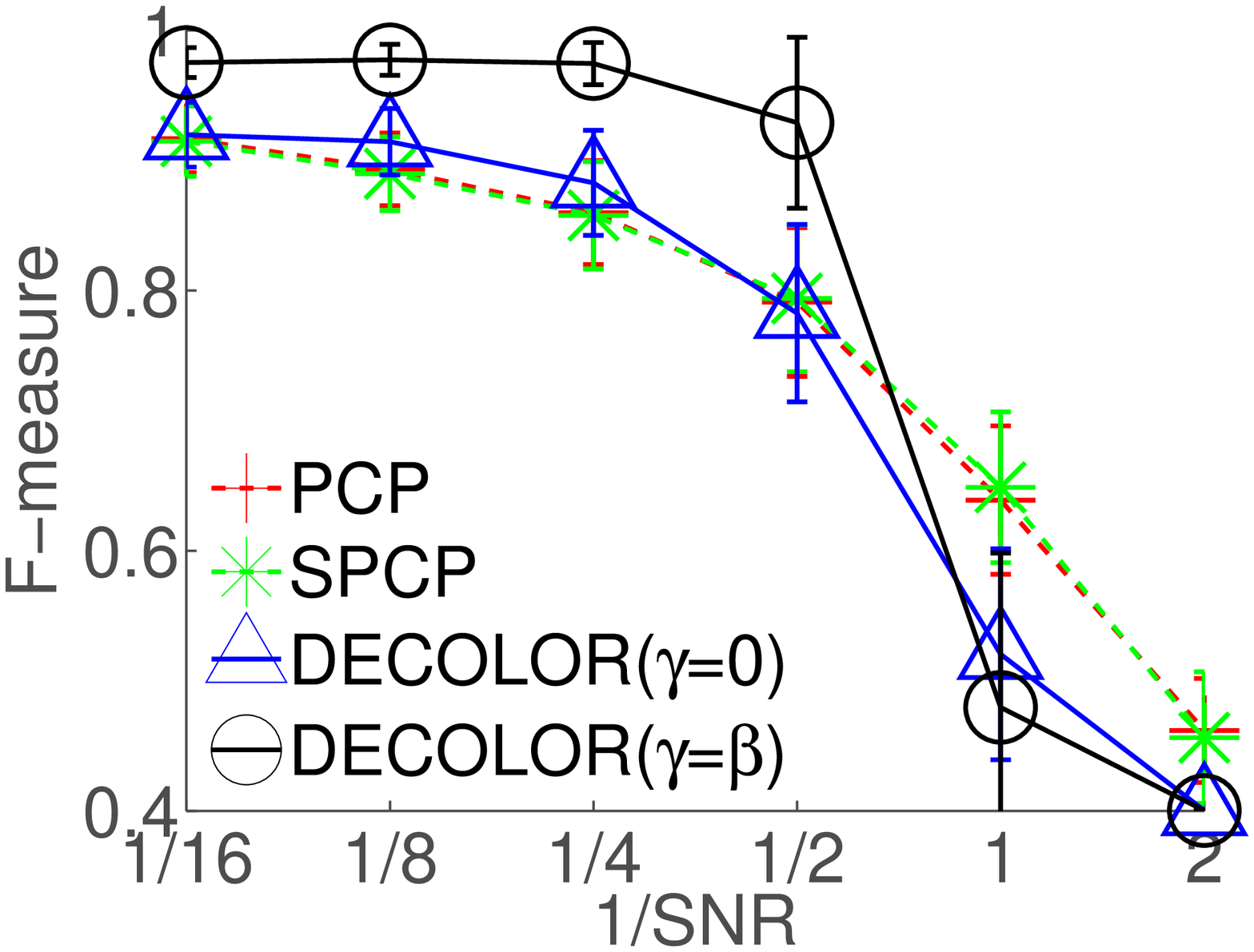}}
 \centerline{\includegraphics[width=1\linewidth]{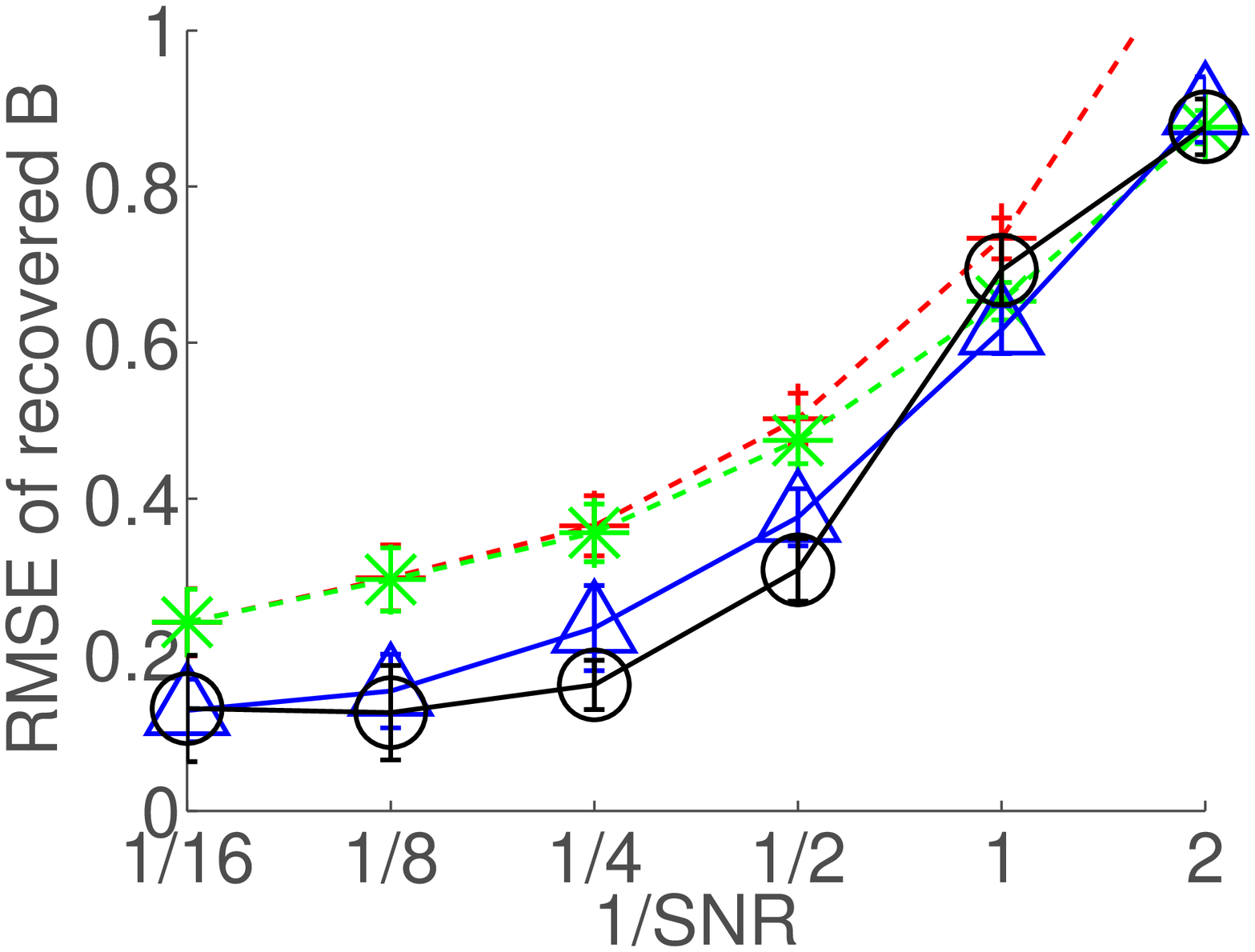}}
 \centerline{(b)}\medskip
\end{minipage}
\hspace{5mm}
\begin{minipage}[b]{.29\linewidth}
 \centerline{\includegraphics[width=1\linewidth]{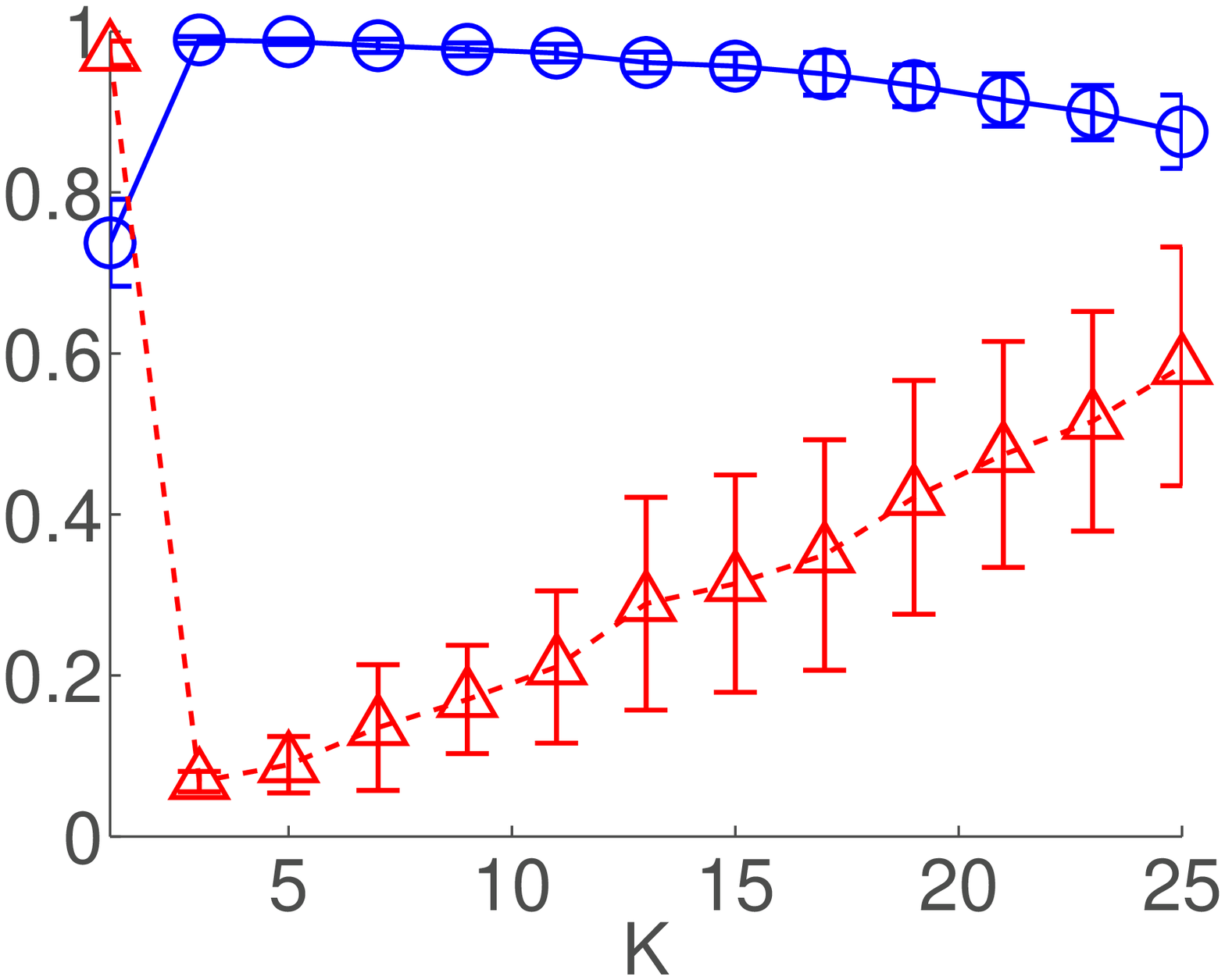}}
 \centerline{\includegraphics[width=1\linewidth]{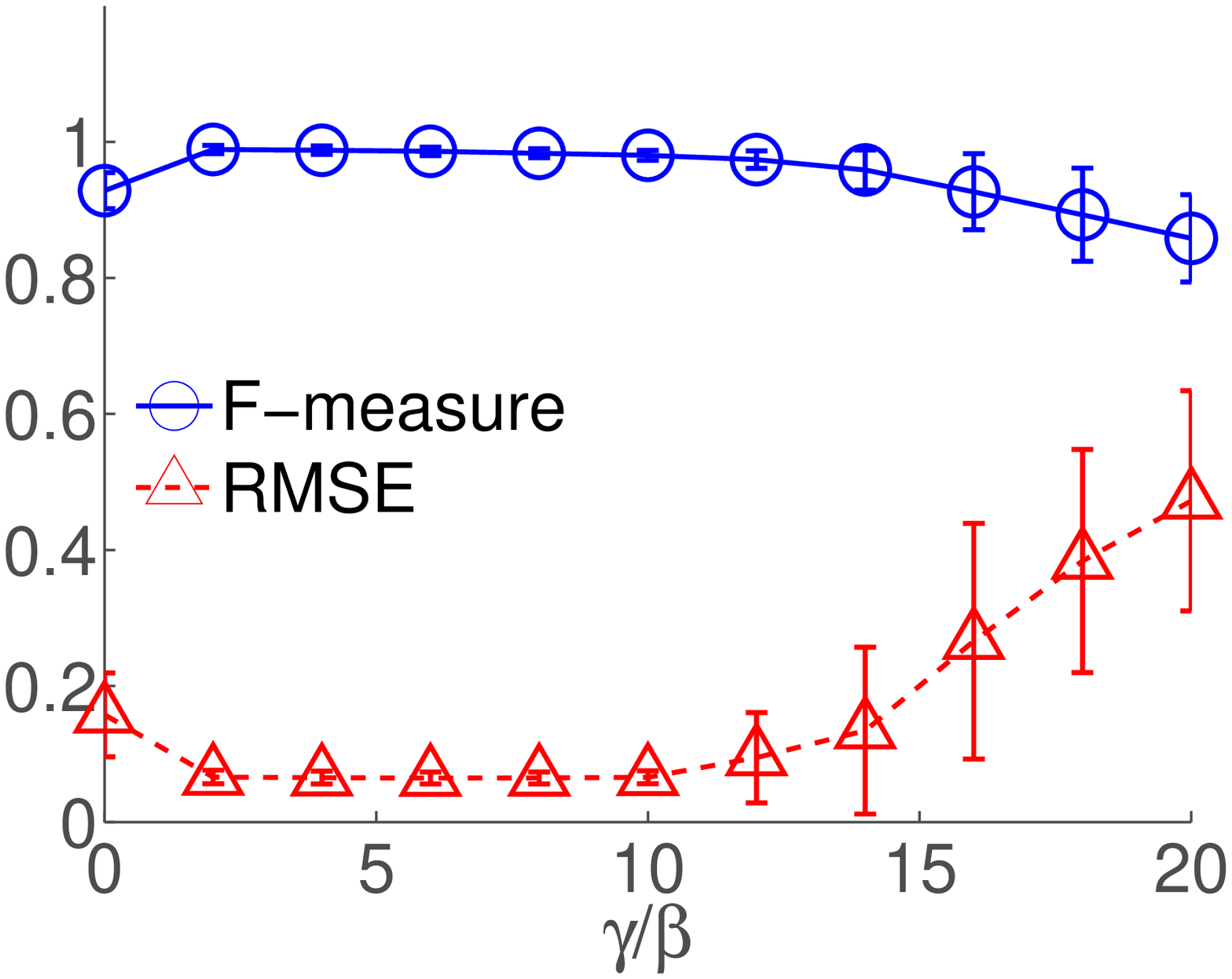}}
 \centerline{(c)}\medskip
\end{minipage}
\caption{\small Quantitative evaluation. \textbf{(a)} F-measure and RMSE as functions of $W$, when $\rm{SNR}=10$.
\textbf{(b)} F-measure and RMSE as functions of SNR, when $W=25$. \textbf{(c)} The effects of parameters, \ie $K$ and $\gamma$. The results are averaged over 50 random trials with $W=25$ and $SNR=10$. The top panel shows the effect of $K$. The true rank of $B_0$ is 3. The accuracy increases sharply when $K$ changes from 1 to 3 and decreases smoothly after $K$ is larger than 3. The bottom panel shows the effect of $\gamma$. The accuracy keeps stable within $[\beta,10\beta]$. }\label{Fig_simu_quant}
\end{figure*}

Fig. \ref{Fig_simuData} gives a qualitative comparison between PCP and DECOLOR. Fig. \ref{Fig_simuData}(c) presents the results of PCP. Notice the artifacts in $\hat B$ that spatially coincide with $S_0$, which shows that the $\ell_1$-penalty is not robust enough for relatively dense errors distributed in a contiguous region. Fig. \ref{Fig_simuData}(d) shows the results of DECOLOR. We see less false detections in estimated $\hat S$ compared with PCP. Also, the recovered $\hat B$ is less corrupted by outliers.

For quantitative evaluation, we perform random experiments with different object width $W$ and SNR. Fig. \ref{Fig_simu_quant}(a) reports the numerical results as functions of $W$. We can see that all methods achieve a high accuracy when $W=10$, which means all of them work well when outliers are really sparse. As $W$ increases, the performance of PCP degrades significantly, while that of DECOLOR keeps less affected. This demonstrates the robustness of DECOLOR. The result of DECOLOR with $\gamma=0$ falls in between those of PCP and DECOLOR with $\gamma=\beta$, and it has a larger variance. This shows the importance of the contiguity prior. Moreover, we can find that DECOLOR gives a very stable performance for outlier detection (F-measure), while the accuracy of matrix recovery (inverse to RMSE) drops obviously as $W$ increases. The reason is that some background pixels are always occluded when the foreground is too large, such that they can not be recovered even when the foreground can be detected accurately.

Fig. \ref{Fig_simu_quant}(b) shows the results under different noise levels. DECOLOR maintains better performance than PCP if SNR is relatively high, but drops dramatically after $\rm{SNR}<2$. This can be interpreted by the property of non-convex penalties. Compared with $\ell_1$-norm, non-convex penalties are more robust to gross errors \cite{candes2008enhancing} but more sensitive to entrywise perturbations \cite{mazumder2011sparsenet}. In general cases of natural video analysis, SNR is much larger than 1. Thus, DECOLOR can work stably.

\subsubsection{Effects of parameters}\label{section_simu_par}

Fig. \ref{Fig_simu_quant}(c) demonstrates the effects of parameters in Algorithm \ref{Alg_overall}, \ie $K$ and $\gamma$.

The parameter $K$ is the rough estimate of ${\rm rank}(B_0)$, which controls the complexity of the background model. Here, the true rank of $B_0$ is 3. From the top plot in Fig. \ref{Fig_simu_quant}(c), we can see that the optimal result is achieved at the turning point where $K=3$. After that, the accuracy decreases very smoothly as $K$ increases. This insensitivity to $K$ is attributed to the shrinkage effect of the nuclear norm in (\ref{Eq_energy_S_1}), which plays an important role to prevent overfitting when estimating $B$. Specifically, given parameters $K$ and $\alpha$, the singular values of $\hat B$ are always shrunk by $\alpha$ due to the soft-thresholding operator in (\ref{softshrinksolution}). Thus, our model overfits slowly when $K$ is larger than the true rank. Similar results can be found in \cite{Mazumder2010spectral}.

The parameter $\gamma$ controls the strength of mutual interaction between neighboring pixels. From the bottom plot in Fig. \ref{Fig_simu_quant}(c), we can see that the performance keeps very stable when $\gamma \in [\beta,10\beta]$.

\subsubsection{Inseparable cases}\label{section_Separable}

\begin{figure}
\begin{minipage}[b]{.33\linewidth}
\centering
\centerline{\includegraphics[width=1\linewidth]{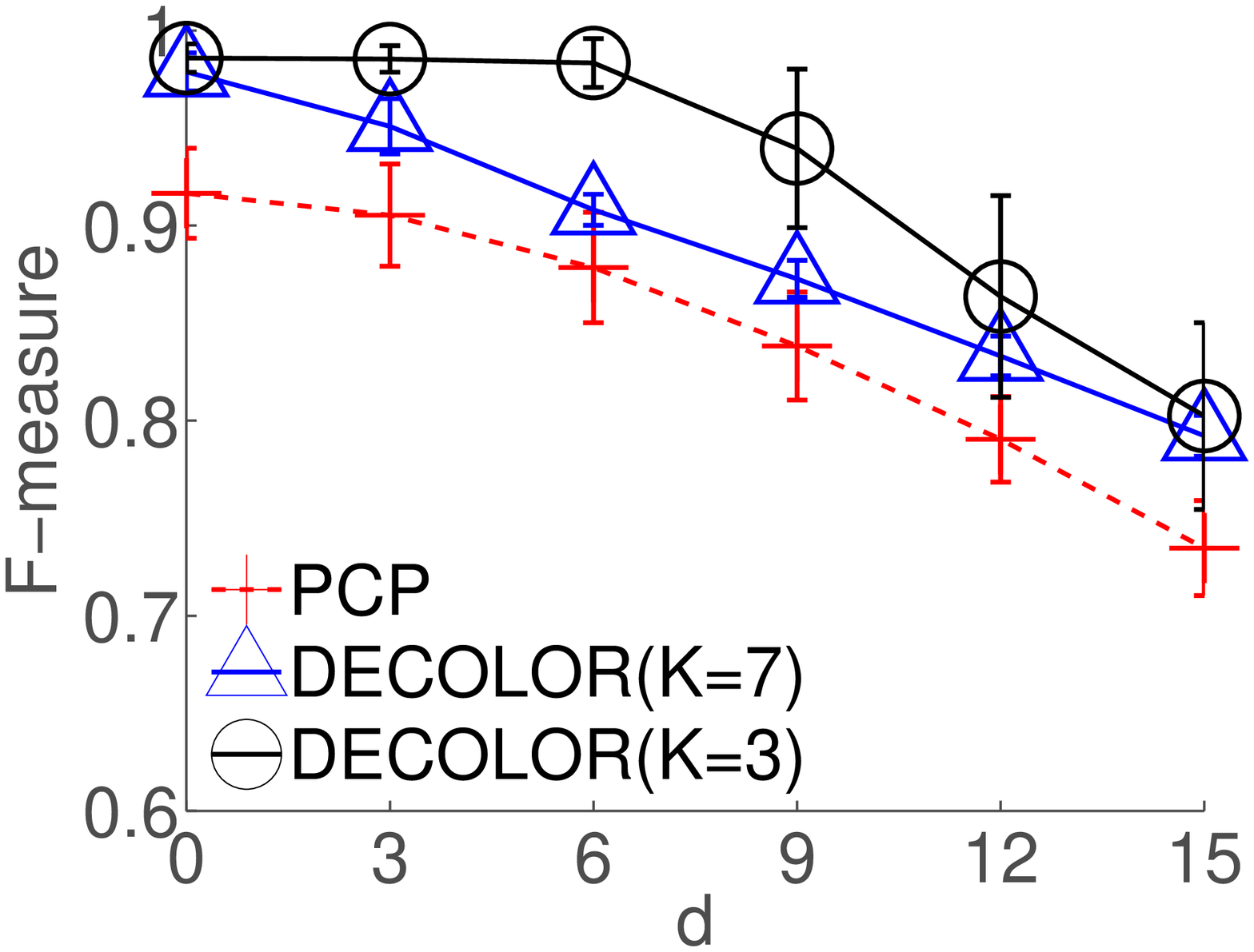}}
\centerline{(a)}
\end{minipage}
\hfill
\begin{minipage}[b]{.6\linewidth}
\begin{minipage}[b]{.45\linewidth}
 \centerline{DECOLOR}\medskip
 \centerline{\includegraphics[width=1\linewidth]{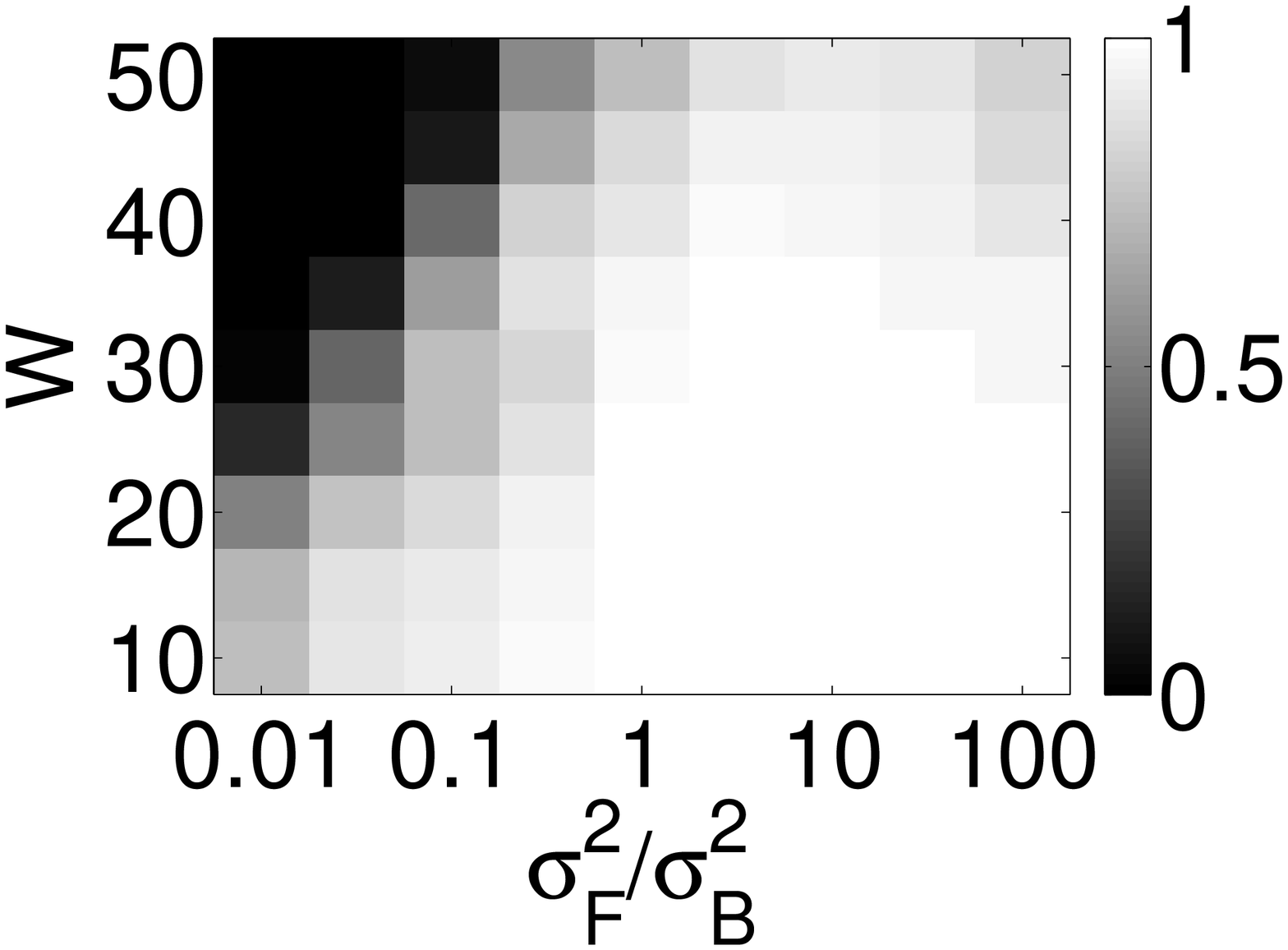}}
\end{minipage}
\begin{minipage}[b]{.45\linewidth}
 \centerline{PCP}\medskip
 \centerline{\includegraphics[width=1\linewidth]{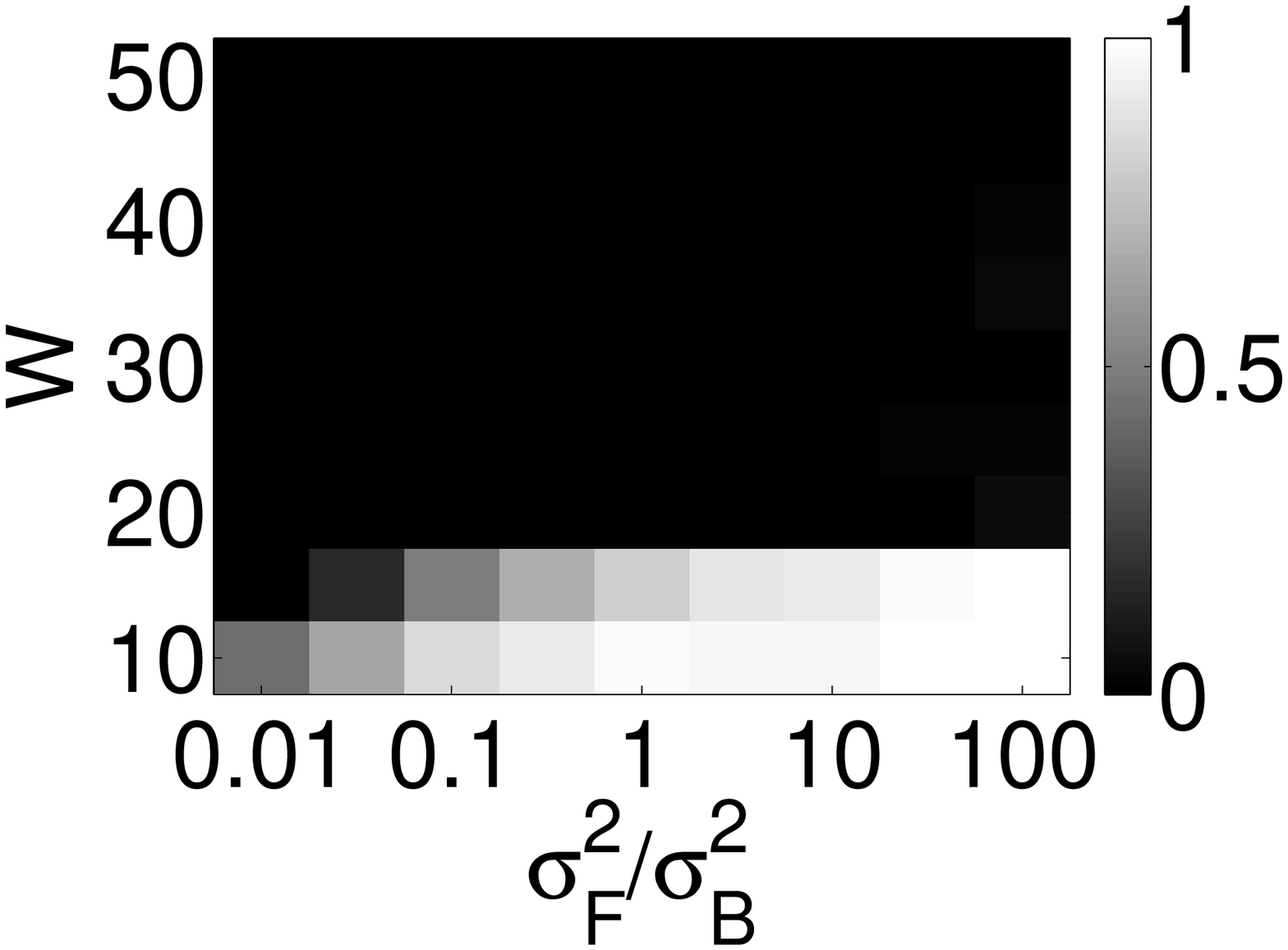}}
\end{minipage}
\centerline{(b)}
\end{minipage}
\caption{\small Simulation to illustrate inseparable cases of DECOLOR. \textbf{(a)} F-measure as a function of $d$, where $d$ is the number of frames within which the foreground stops moving. The true rank of $B_0$ is 3. \textbf{(b)} Fraction of trials of accurate foreground detection (F-measure$>$0.95) over 200 trials, as a function of $\sigma_F$ and $W$. Here, $\sigma_F$ represents the standard deviation of foreground intensities and $W$ denotes the foreground width. $\sigma_B$ is the standard deviation of $B_0$.
}\label{Fig_simu_quant_F}
\end{figure}

In previous simulations, the foreground is always moving and the foreground entries are sampled from a uniform distribution with a relatively large variance. Under these conditions, DECOLOR performs effectively and stably for foreground detection (F-measure) unless SNR is too bad. Next, we would like to study the cases when DECOLOR can not separate the foreground from the background correctly.

Firstly, we let the foreground not move for $d$ frames when generating the data. Fig. \ref{Fig_simu_quant_F}(a) shows the averaged F-measure as a function of $d$. Here, ${\rm rank}(B_0)=3$. We can see that, with the default parameter $K=7$, the accuracy of DECOLOR will decrease dramatically as long as $d>0$. This is because DECOLOR overfits the static foreground into the background model, as the model dimension $K$ is larger than its actual value. When we decrease $K$ to 3, DECOLOR performs more stably until $d>6$, which means that DECOLOR can tolerate temporary stopping of foreground motion. In short, when the object is not always moving, DECOLOR becomes more sensitive to $K$, and it can not work when the object stops for a long time.

Next, to investigate the influence of foreground texture, we also run DECOLOR on random problems with outlier entries sampled from uniform distributions with random mean and different variances $\sigma_F^2$. Fig. \ref{Fig_simu_quant_F}(b) displays the fraction of trials in which DECOLOR gives a high accuracy of foreground detection (F-measure$>$0.95) over 200 trials, as a 2D function of $\sigma_F^2$ and $W$. The result of PCP is also shown for comparison. As we can see, DECOLOR can achieve accurate detection with a high probability over a wide range of conditions, except for the upper left corner where $W$ is large and $\sigma_F^2$ is small, which represents the case of large and textureless foreground. In practice, the interior motion of a textureless object is undetectable. Thus, its interior region will keep unchanged for a relatively long time if the object is large or moving slowly. In this case, the interior part of the foreground may fit into the low-rank model, which makes DECOLOR fail.

\subsection{Real Sequences}

We test DECOLOR on real sequences from public datasets for background subtraction, motion segmentation and dynamic texture detection. Please refer to Table \ref{Tab_data} for the details of each sequence.

\begin{table}[H]
\renewcommand{\arraystretch}{1.1}
\caption{Information of the sequences used in experiments.}
\label{Tab_data}
\centering
\begin{tabular}{lllll}
\toprule
Fig. & Size$\times$\#frames & Ref. & Description \\ [0.5ex]
\hline 
Fig.  \ref{Fig_BackSub}(a) & $[160,120]\times48$& \cite{wang2006novel}& Crowded scene \\
Fig.  \ref{Fig_BackSub}(b) &$[238,158]\times24$ & \cite{chan2009layered}& Crowded scene \\
Fig.  \ref{Fig_BackSub}(c)&$[160,128]\times24$ &\cite{li2004statistical}& Crowded scene \\
Fig.  \ref{Fig_BackSub}(d)&$[160,128]\times48$ & \cite{li2004statistical}& Dynamic background \\
Fig.  \ref{Fig_BackSub}(e)&$[160,128]\times48$ & \cite{li2004statistical}& Dynamic background \\
Fig.  \ref{Fig_MotionSeg}(a)&$[320,240]\times40$ & \cite{brox2010object} & Moving cameras \\
Fig.  \ref{Fig_MotionSeg}(b) &$[320,240]\times30$ & \cite{brox2010object}& Moving cameras \\
Fig.  \ref{Fig_MotionSeg}(c) &$[320,240]\times30$ & \cite{brox2010object}& Moving cameras \\
Fig.  \ref{Fig_MotionSeg}(d) &$[320,240]\times24$ & \cite{brox2010object}& Moving cameras \\
Fig.  \ref{Fig_smoke} &$[180,144]\times48$ & \cite{fazekas2009dynamic}& Dynamic foreground \\
\bottomrule
\end{tabular}
\end{table}

\subsubsection{Comparison to sparse signal recovery}\label{section_exp_regression}

\begin{figure}
\centering
\begin{minipage}[t]{0.18\linewidth}
    \centerline{\includegraphics[width=\linewidth]{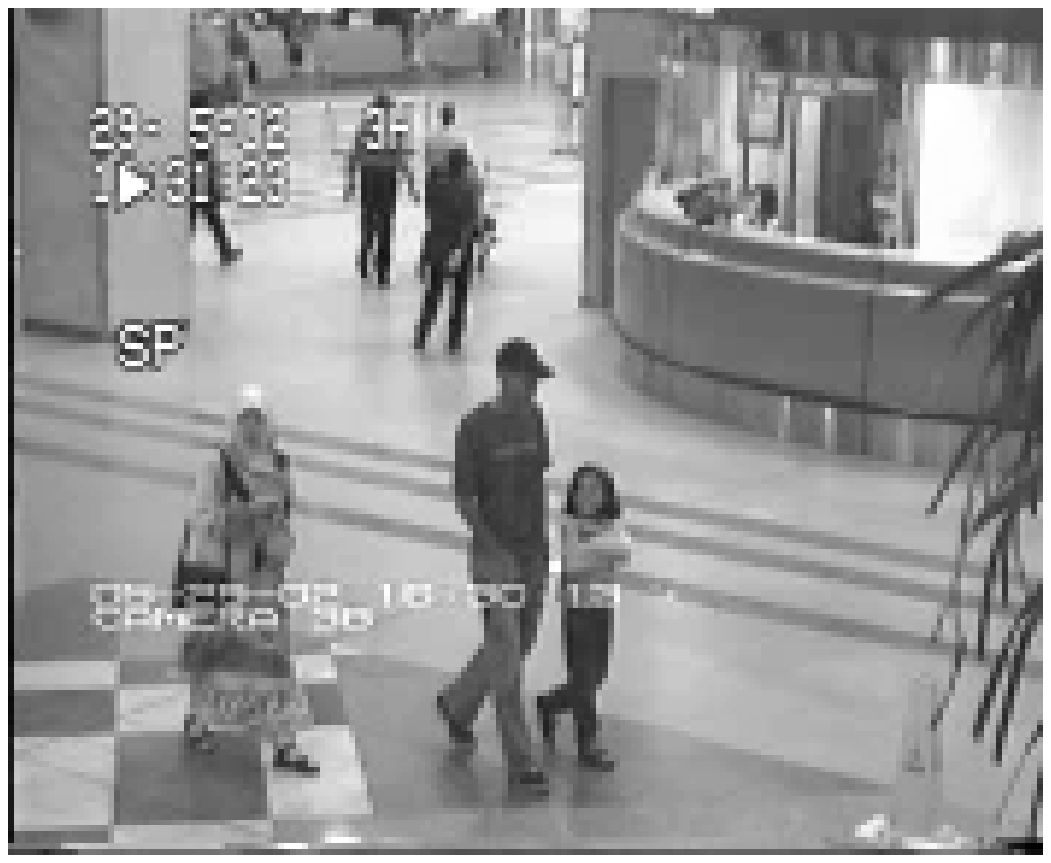}}
    \centerline{\includegraphics[width=\linewidth]{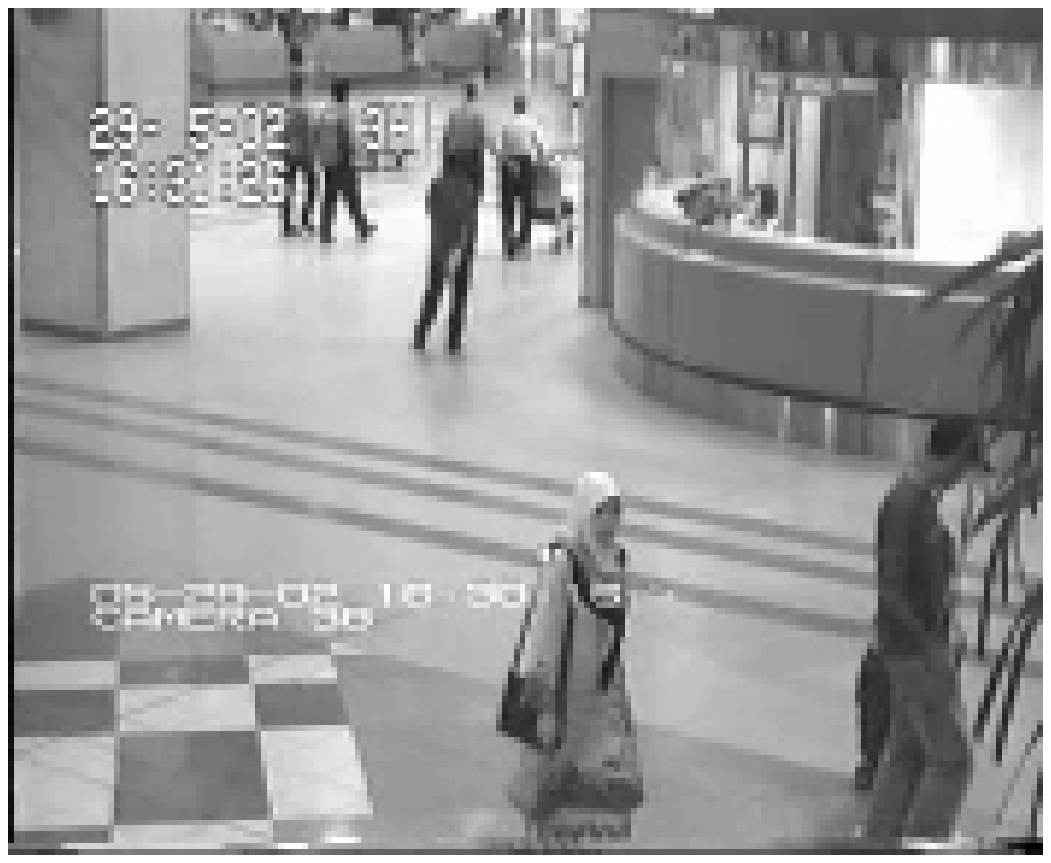}}
    \centerline{Data}
    \centerline{(a)}
\end{minipage}
\hspace{5mm}
\begin{minipage}[t]{0.6\linewidth}
\begin{minipage}[t]{0.3\linewidth}
    \centerline{\includegraphics[width=\linewidth]{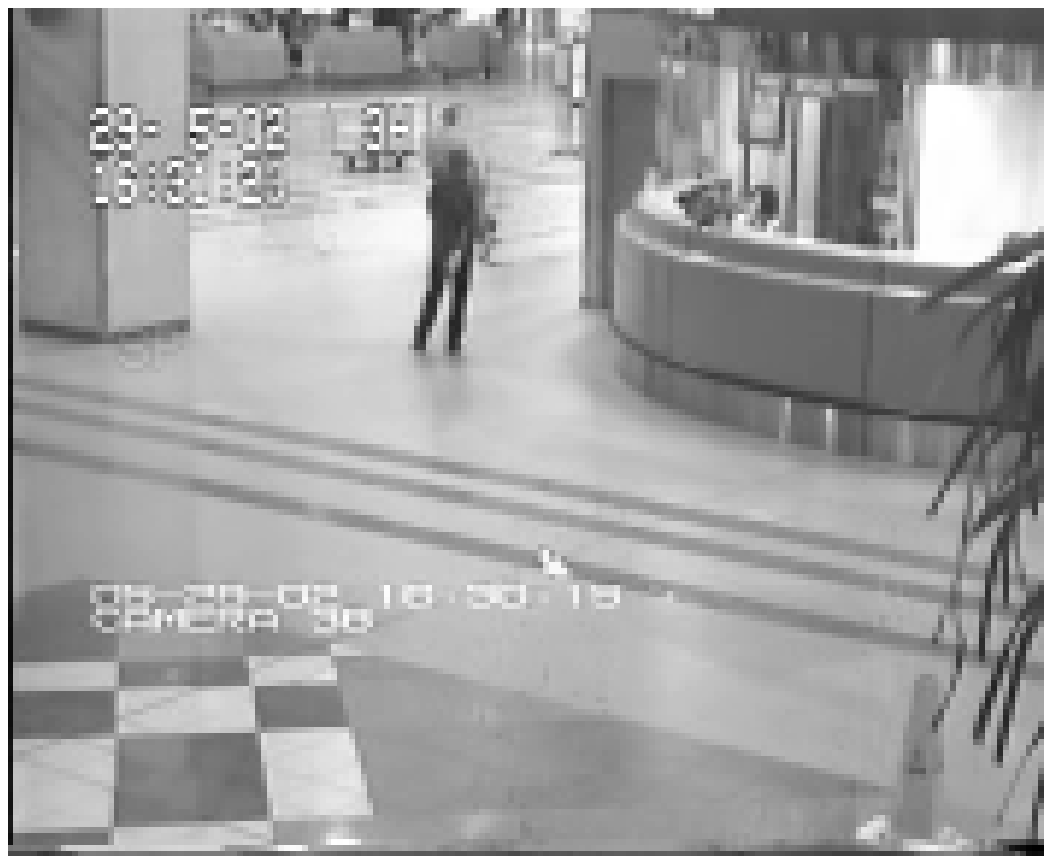}}
    \centerline{\includegraphics[width=\linewidth]{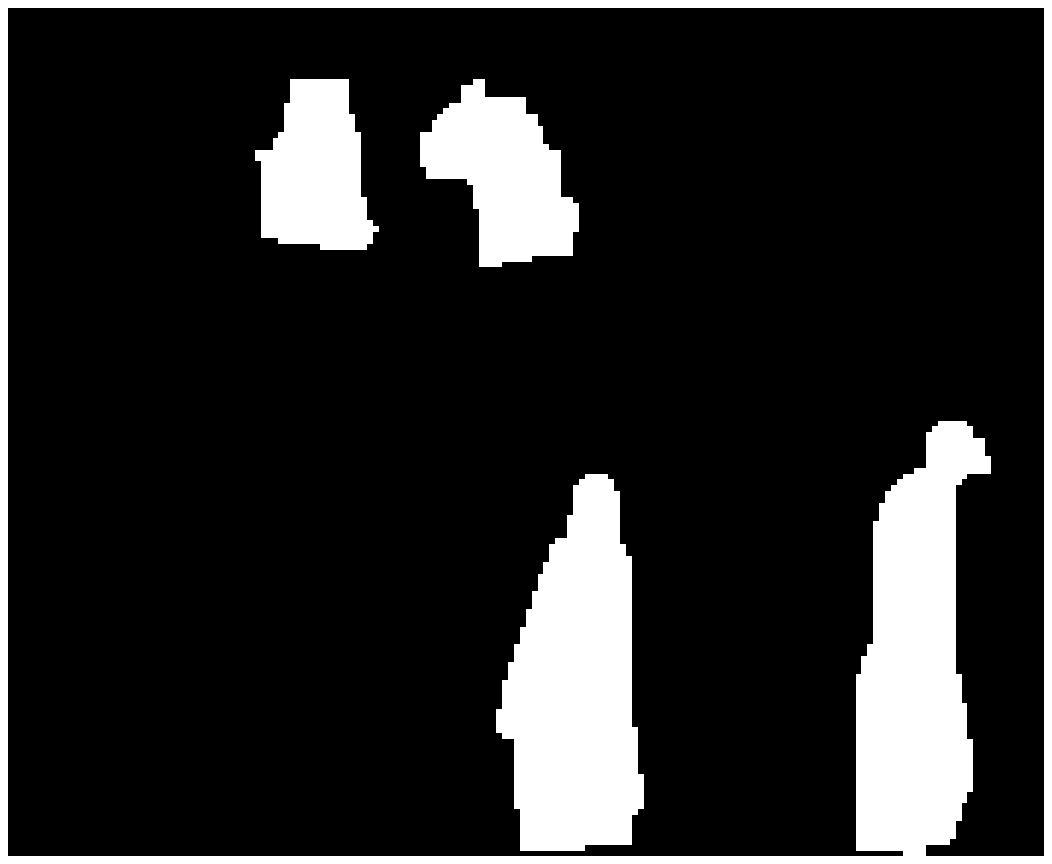}}
    \centerline{DECOLOR}
\end{minipage}
\begin{minipage}[t]{0.3\linewidth}
    \centerline{\includegraphics[width=\linewidth]{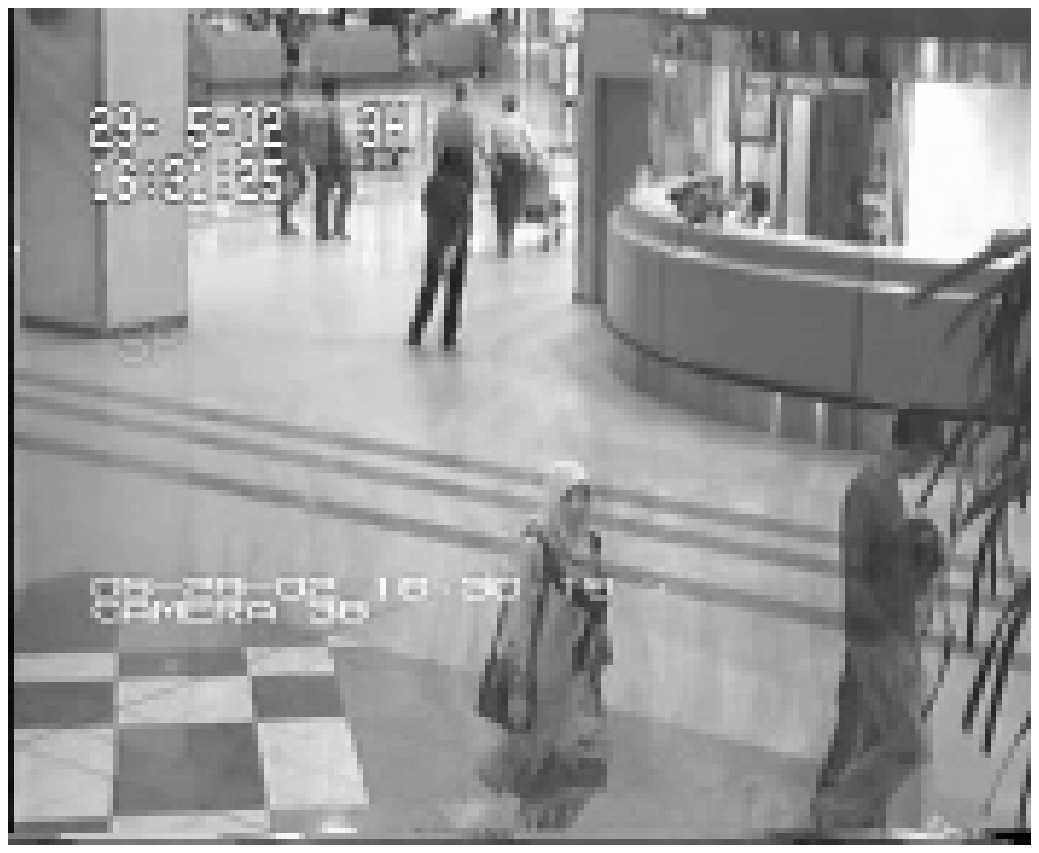}}
    \centerline{\includegraphics[width=\linewidth]{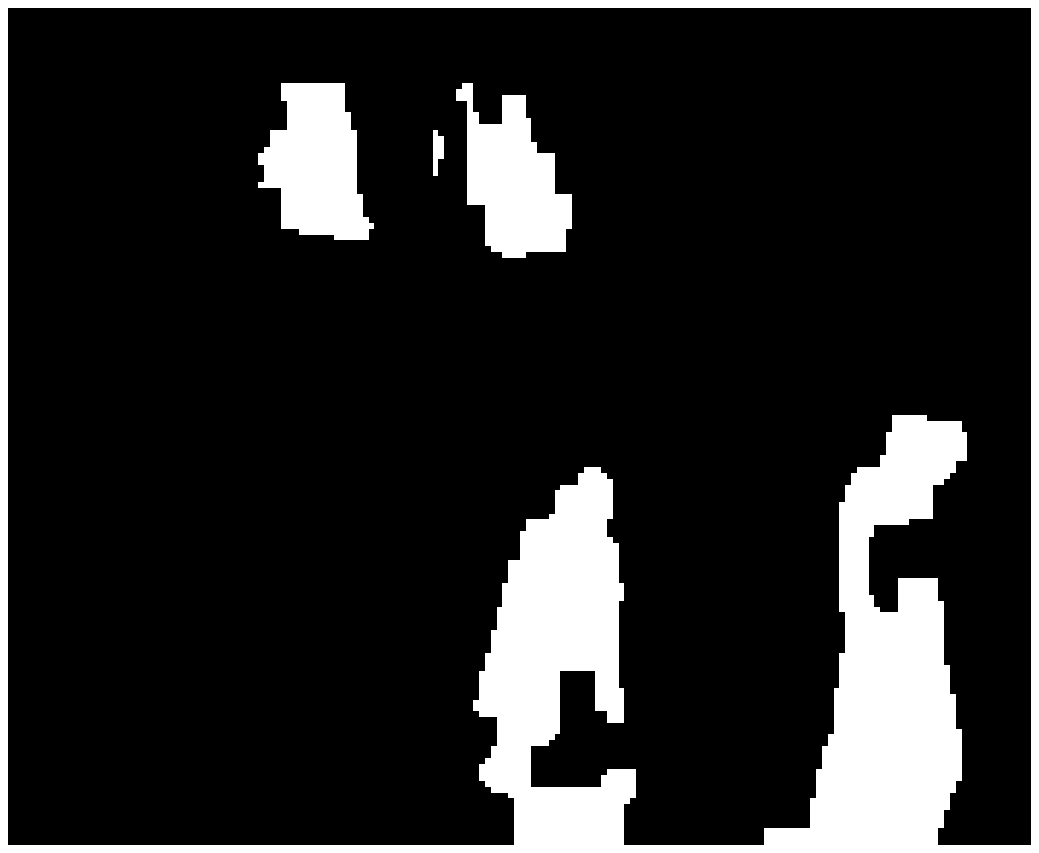}}
    \centerline{ProxFlow}
    \centerline{(b)}
\end{minipage}
\begin{minipage}[t]{0.3\linewidth}
    \centerline{\includegraphics[width=\linewidth]{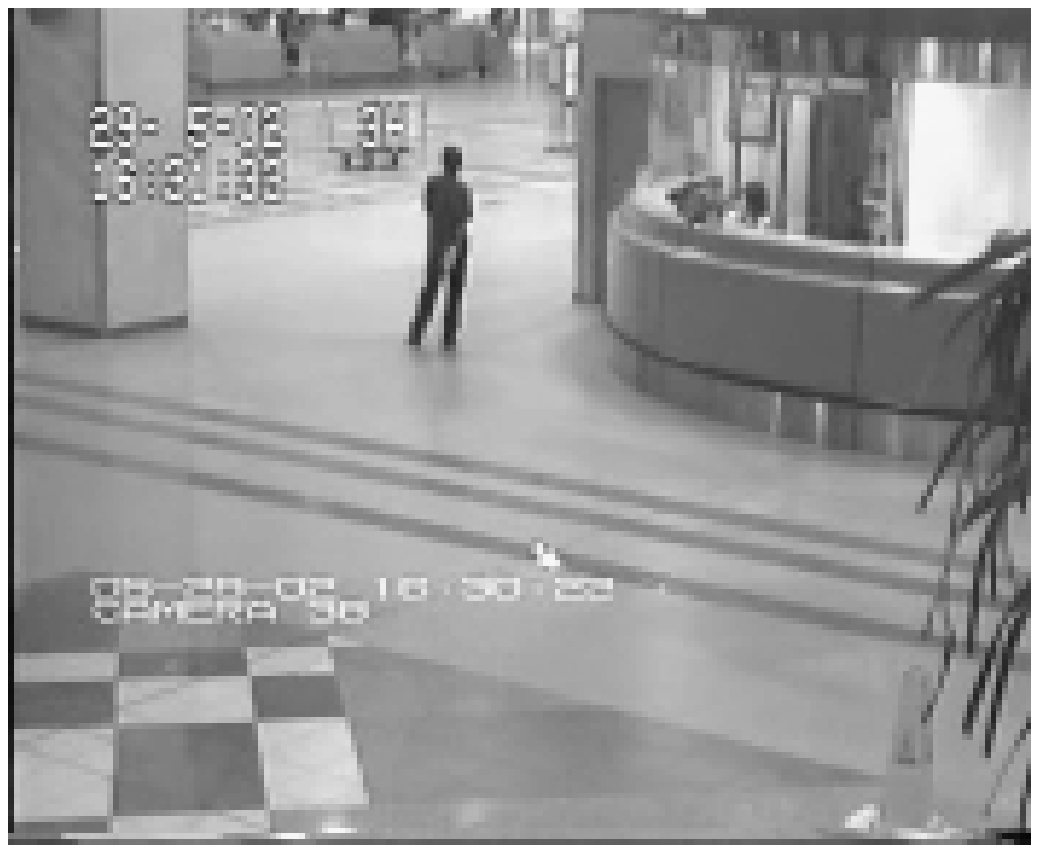}}
    \centerline{\includegraphics[width=\linewidth]{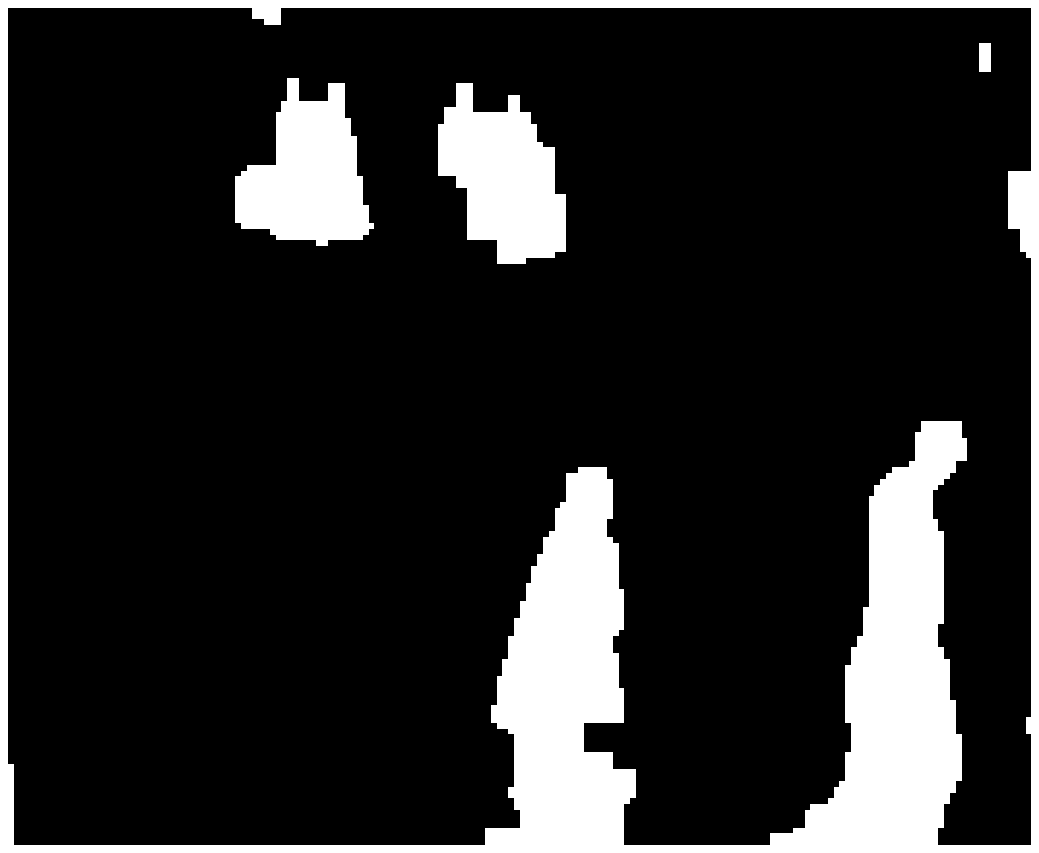}}
    \centerline{ProxFlow+}\vspace{2mm}
\end{minipage}
\end{minipage}
\caption{\small An example illustrating the difference between DECOLOR and sparse signal recovery. \textbf{(a)} The first and the last frames of a sequence of 24 images. Several people are walking and continuously presented in the scene. \textbf{(b)} The estimated background (top) and segmentation (bottom) corresponding to the last frame. ProxFlow means sparse signal recovery by solving (\ref{Eq_ProxFlow}) with the ProxFlow algorithm \cite{mairal2010network}, where the first 23 frames are used as the basis matrix $\Phi$ in (\ref{Eq_ProxFlow}). ProxFlow+ means applying ProxFlow with bases $\Phi$ being the low-rank matrix $\hat B$ learnt by DECOLOR.}\label{Fig_ProxFlow}
\end{figure}

As discussed in Section \ref{section_regression}, a key difference between DECOLOR and sparse signal recovery is the assumption on availability of training sequences. Background subtraction via sparse signal recovery requires a set of background images without foreground, which is not always available especially for surveillance of crowded scenes. Fig. \ref{Fig_ProxFlow}(a) gives such a sequence clipped from the start of an indoor surveillance video, where the couple is always in the scene.

Fig. \ref{Fig_ProxFlow}(b) shows the results of the 24th frame. For sparse signal recovery, we apply the ProxFlow algorithm\footnote{The code is available at http://www.di.ens.fr/willow/SPAMS/} \cite{mairal2010network} to solve the model in (\ref{Eq_ProxFlow}). The previous 23 frames are used as the bases ($\Phi$ in (\ref{Eq_ProxFlow})). Since the subspace spanned by previous frames also includes foreground objects, ProxFlow can not recover the background and gives inaccurate segmentation. Instead, DECOLOR can estimate a clean background from occluded data. In practice, DECOLOR can be used for background initialization. For example, the last column in Fig. \ref{Fig_ProxFlow}(b) shows the results of running ProxFlow with $\Phi$ being low-rank $\hat B$ learnt by DECOLOR. That is, we use the background images recovered by DECOLOR as the training images for background subtraction. We can see that the results are improved apparently.

\subsubsection{Background estimation}

\begin{figure*}
\centering
\begin{minipage}[t]{0.15\linewidth}
    \centerline{Data}
    \vspace{3mm}
\end{minipage}
\hspace{5mm}
\begin{minipage}[t]{0.15\linewidth}
    \centerline{DECOLOR}
\end{minipage}
\begin{minipage}[t]{0.15\linewidth}
    \centerline{PCP}
\end{minipage}
\begin{minipage}[t]{0.15\linewidth}
    \centerline{Median}
\end{minipage}
\begin{minipage}[t]{0.15\linewidth}
    \centerline{MoG}
\end{minipage}
\\
\begin{minipage}[t]{0.15\linewidth}
    \centerline{\includegraphics[width=\linewidth]{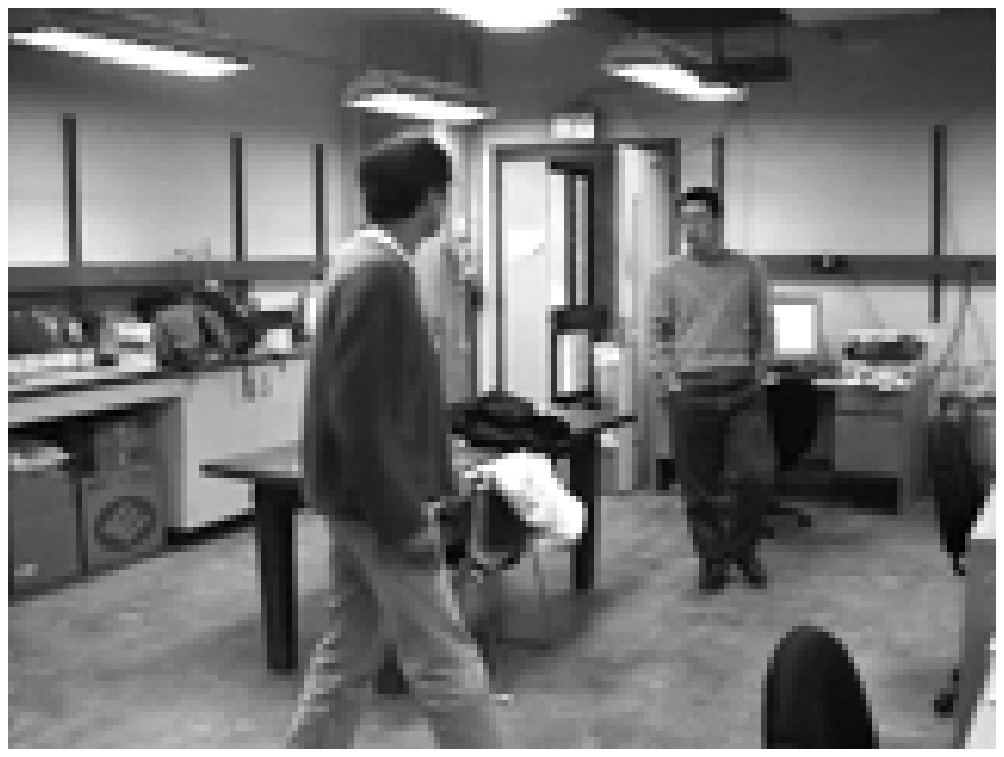}}
    \centerline{\includegraphics[width=\linewidth]{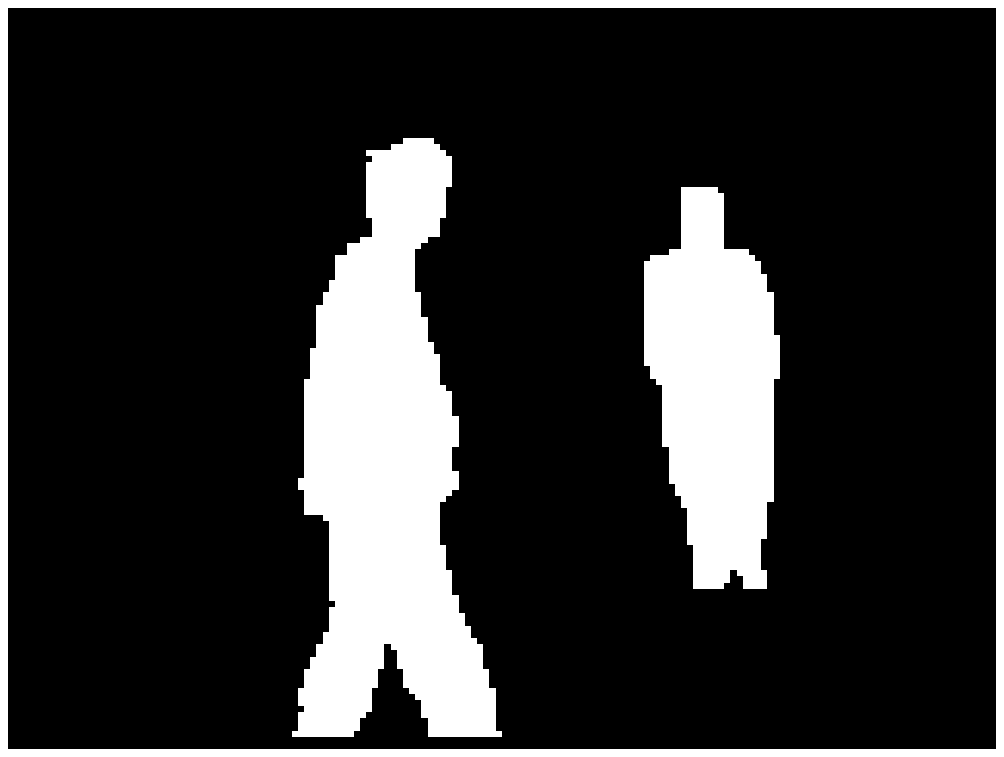}}
\end{minipage}
\hspace{5mm}
\begin{minipage}[t]{0.15\linewidth}
    \centerline{\includegraphics[width=\linewidth]{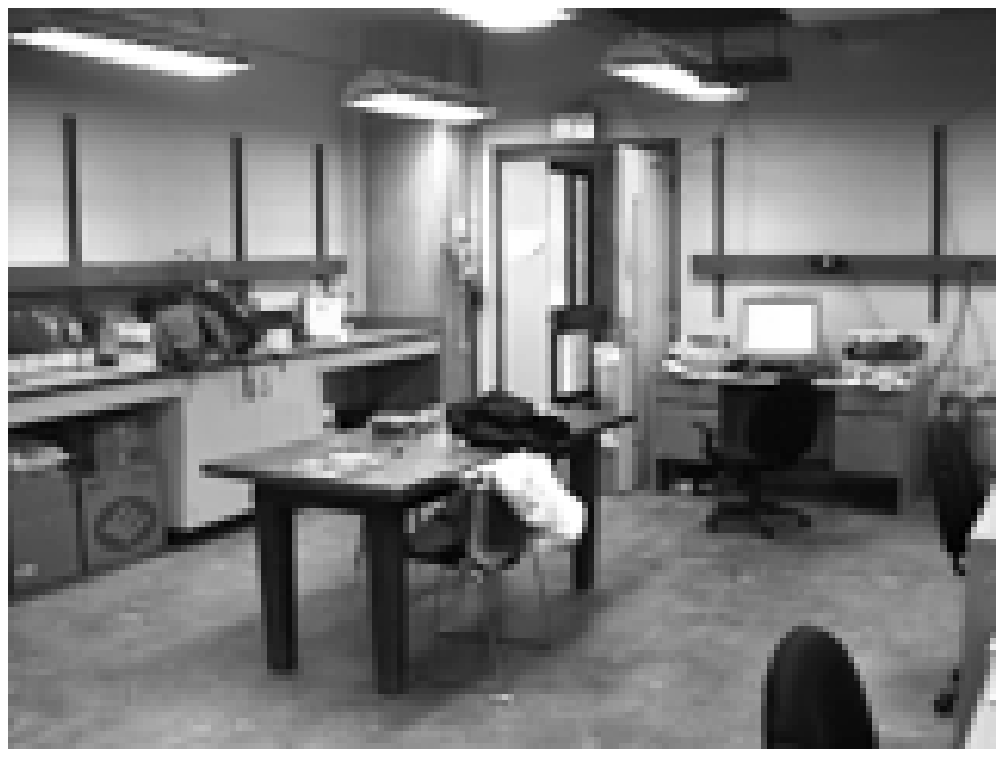}}
    \centerline{\includegraphics[width=\linewidth]{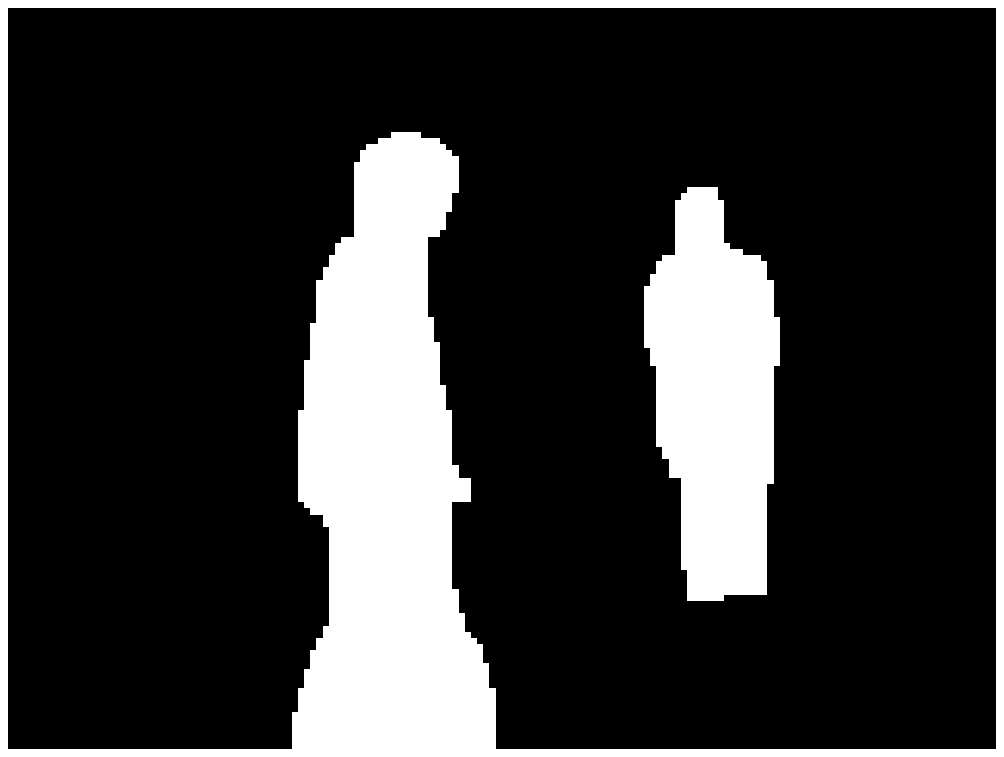}}
\end{minipage}
\begin{minipage}[t]{0.15\linewidth}
    \centerline{\includegraphics[width=\linewidth]{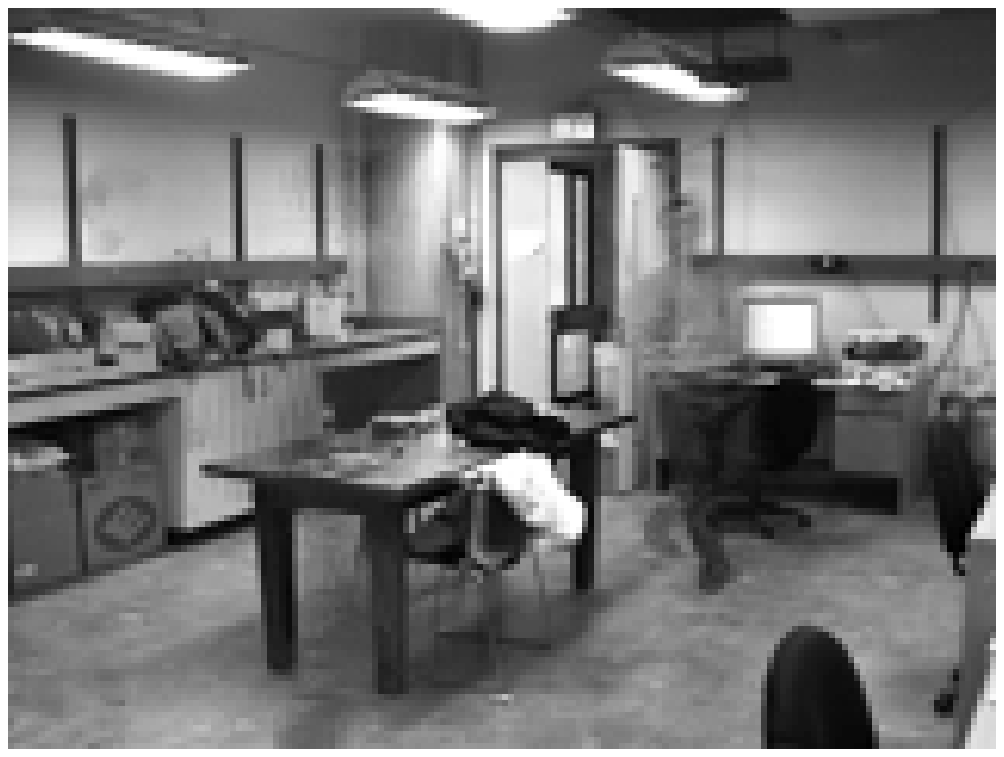}}
    \centerline{\includegraphics[width=\linewidth]{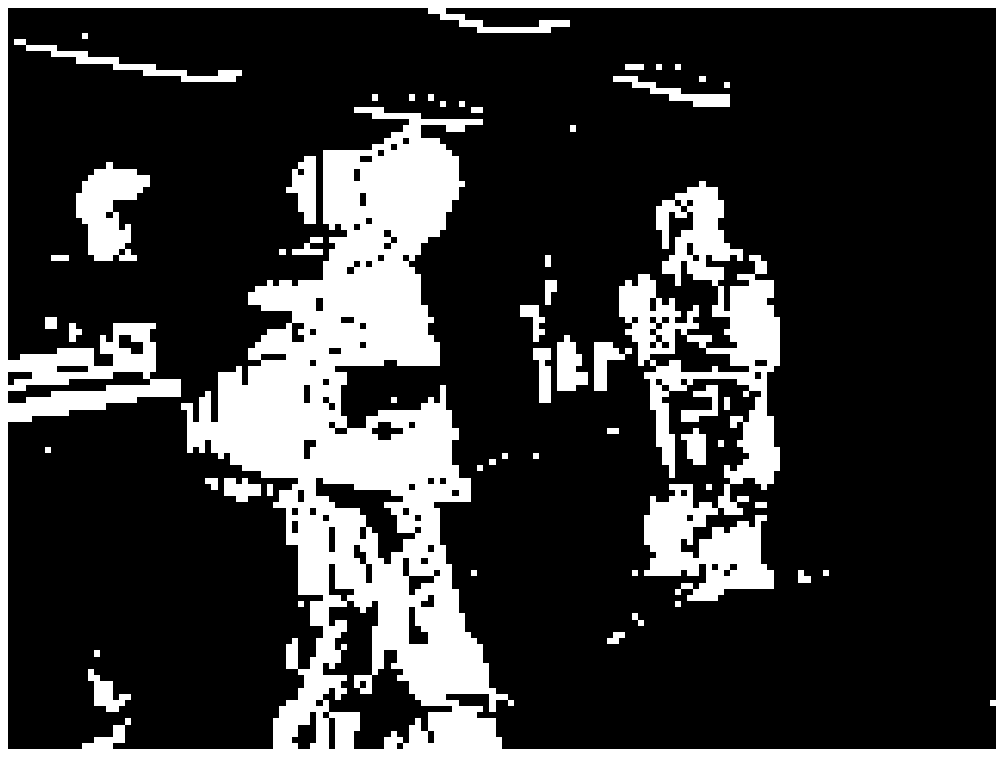}}
\end{minipage}
\begin{minipage}[t]{0.15\linewidth}
    \centerline{\includegraphics[width=\linewidth]{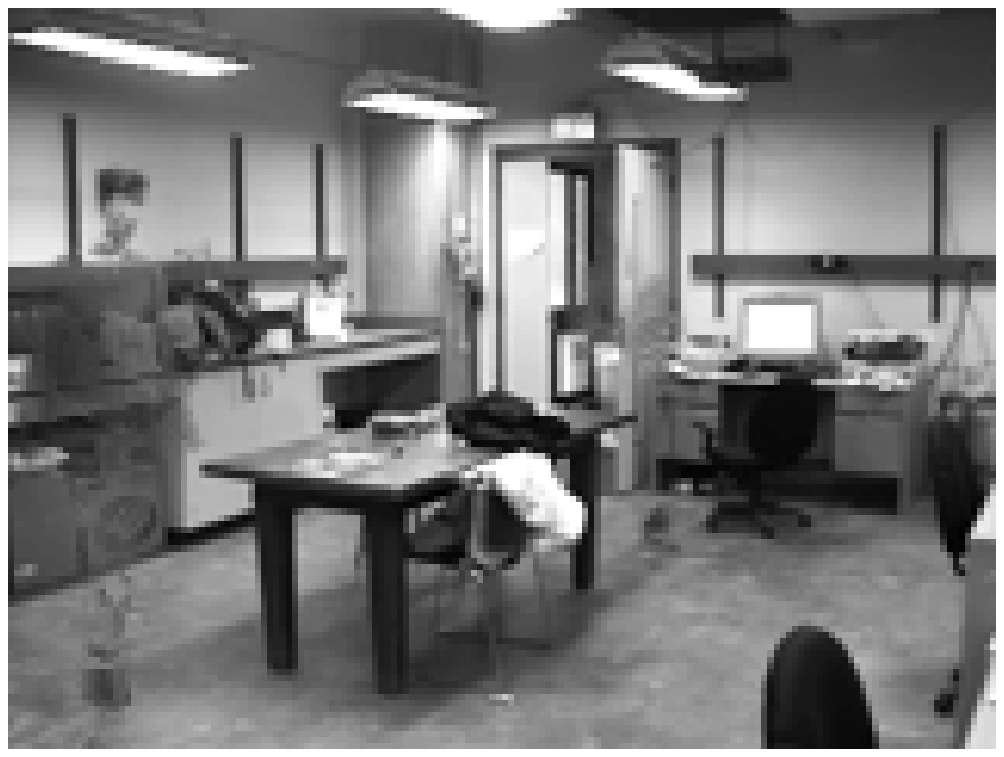}}
    \centerline{\includegraphics[width=\linewidth]{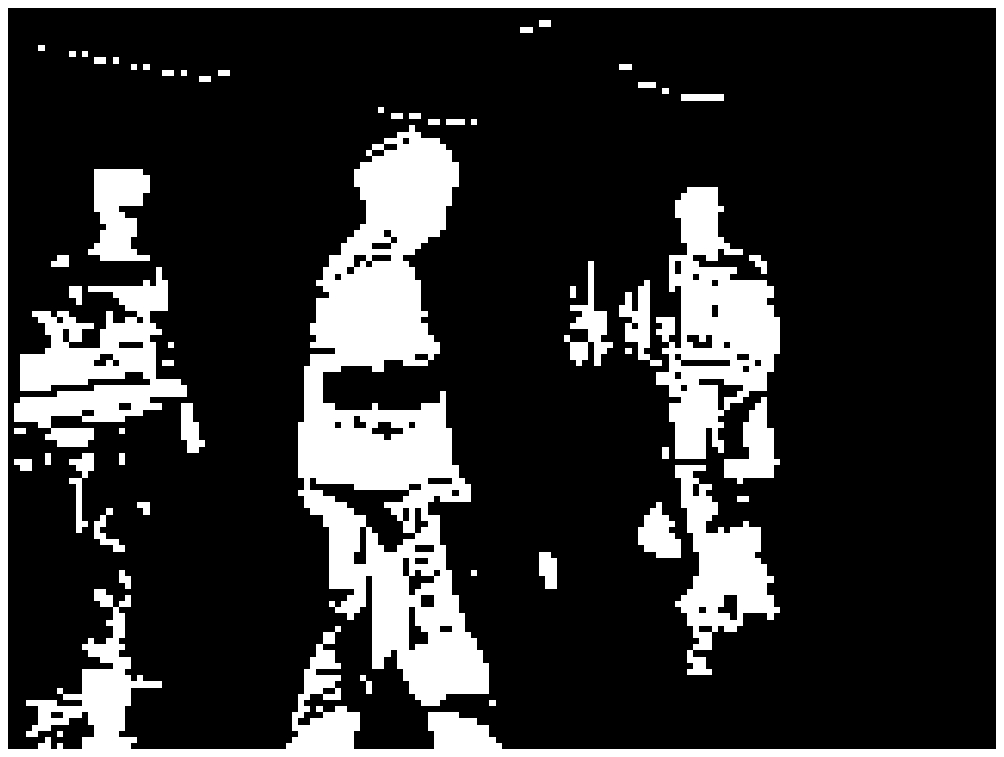}}
\end{minipage}
\begin{minipage}[t]{0.15\linewidth}
    \centerline{\includegraphics[width=\linewidth]{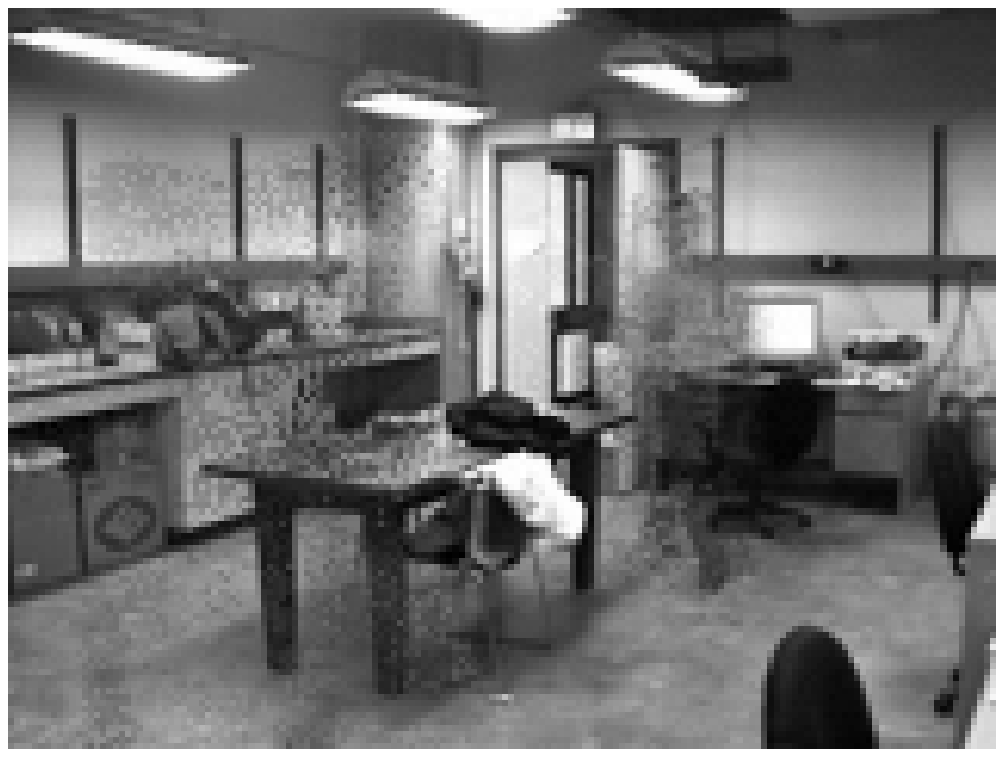}}
    \centerline{\includegraphics[width=\linewidth]{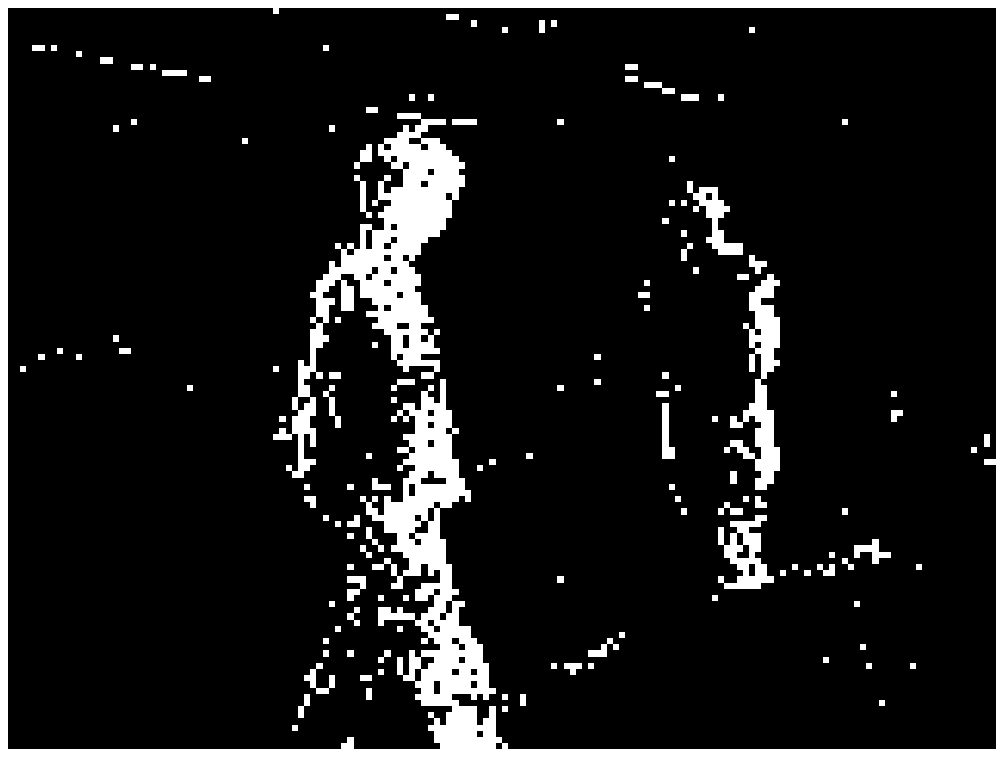}}
\end{minipage}
\centerline{(a)} 
\begin{minipage}[t]{0.15\linewidth}
    \centerline{\includegraphics[width=\linewidth]{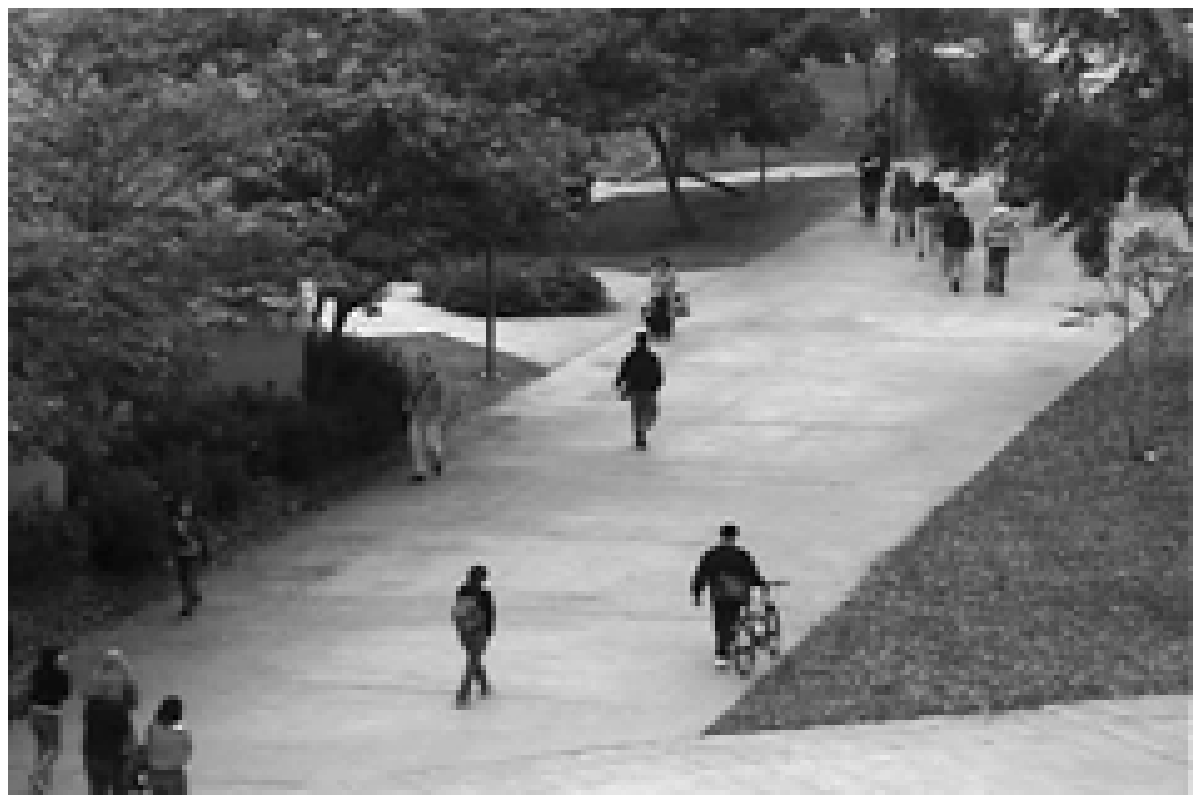}}
    \centerline{\includegraphics[width=\linewidth]{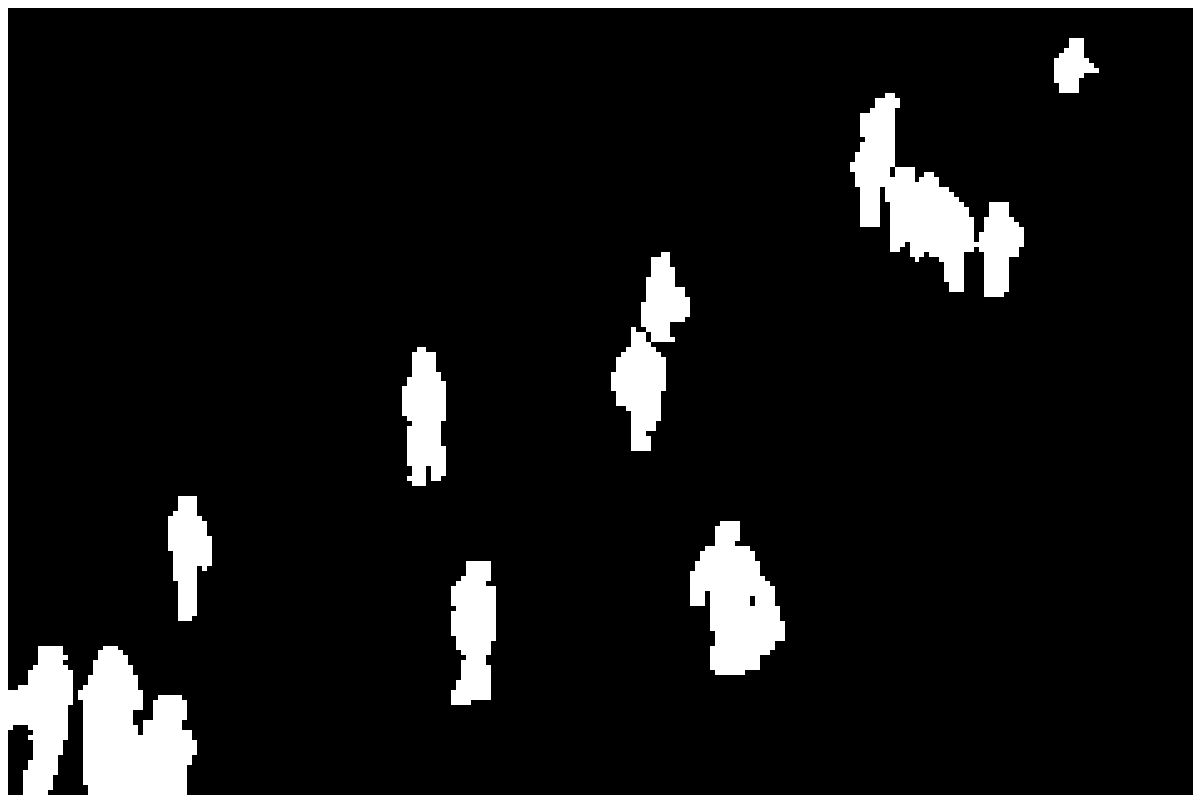}}
\end{minipage}
\hspace{5mm}
\begin{minipage}[t]{0.15\linewidth}
    \centerline{\includegraphics[width=\linewidth]{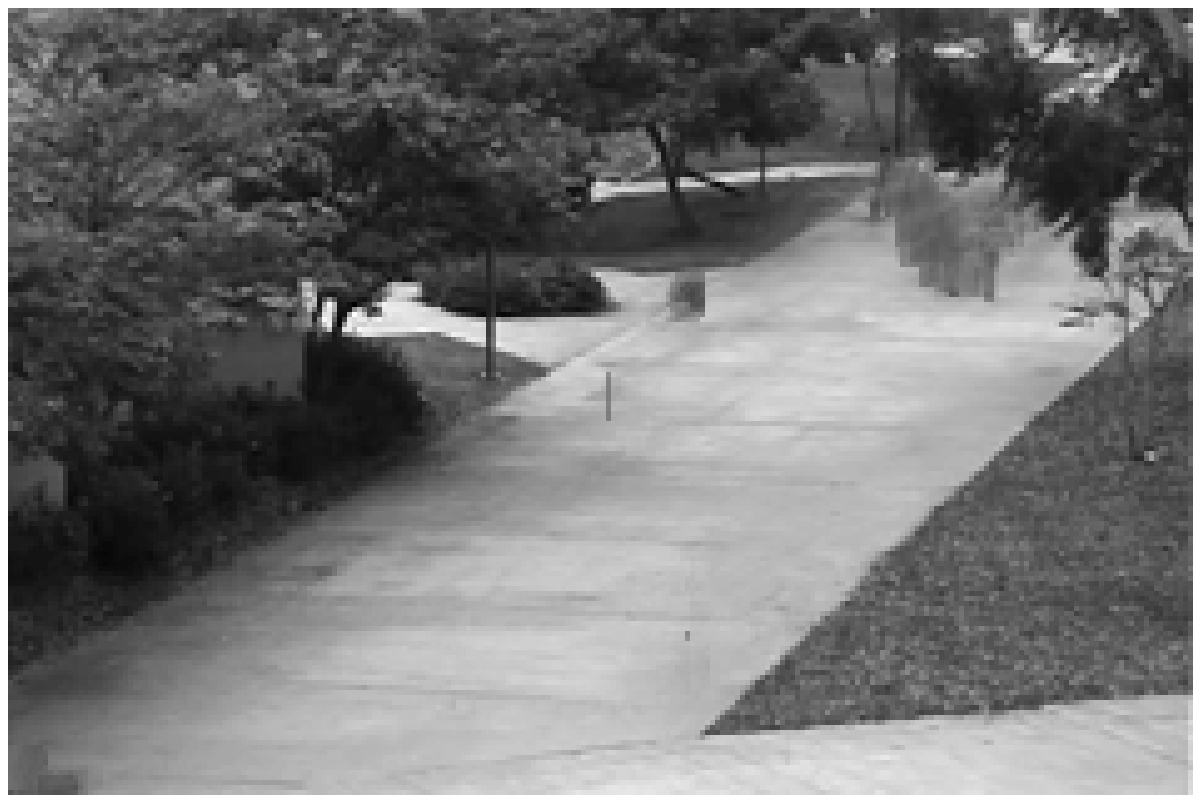}}
    \centerline{\includegraphics[width=\linewidth]{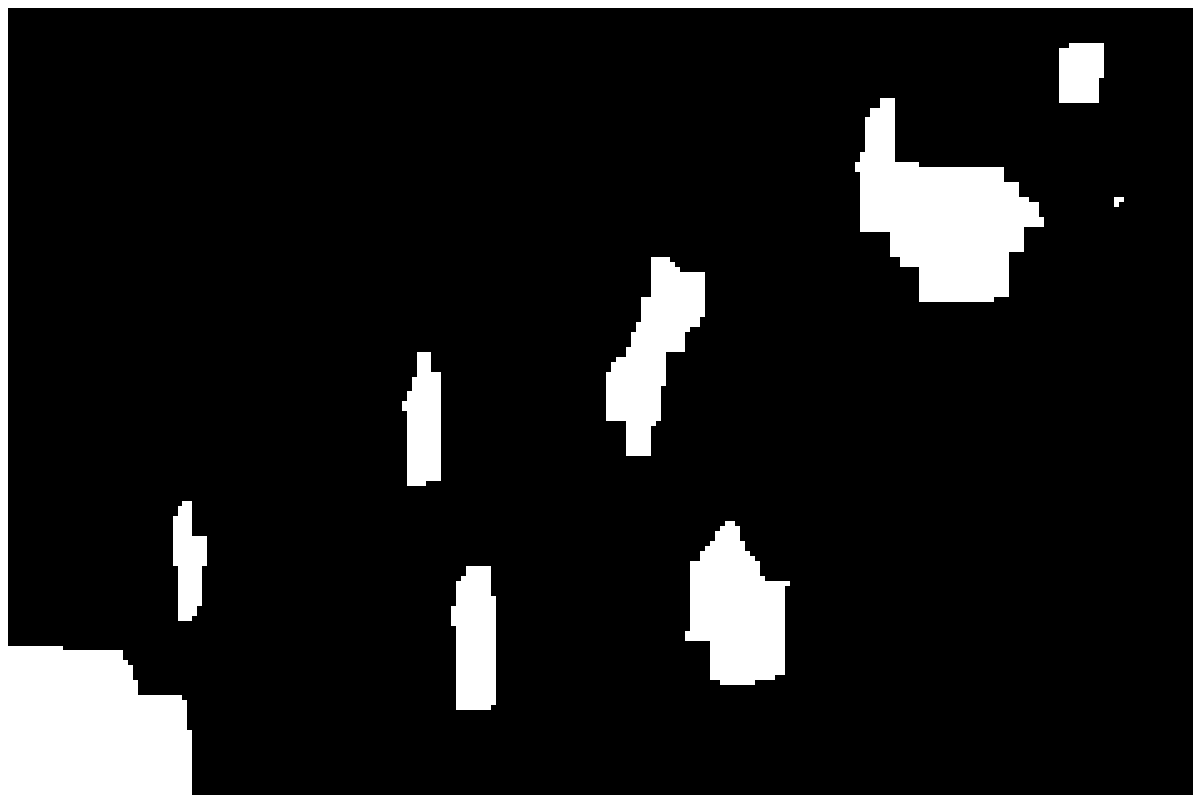}}
\end{minipage}
\begin{minipage}[t]{0.15\linewidth}
    \centerline{\includegraphics[width=\linewidth]{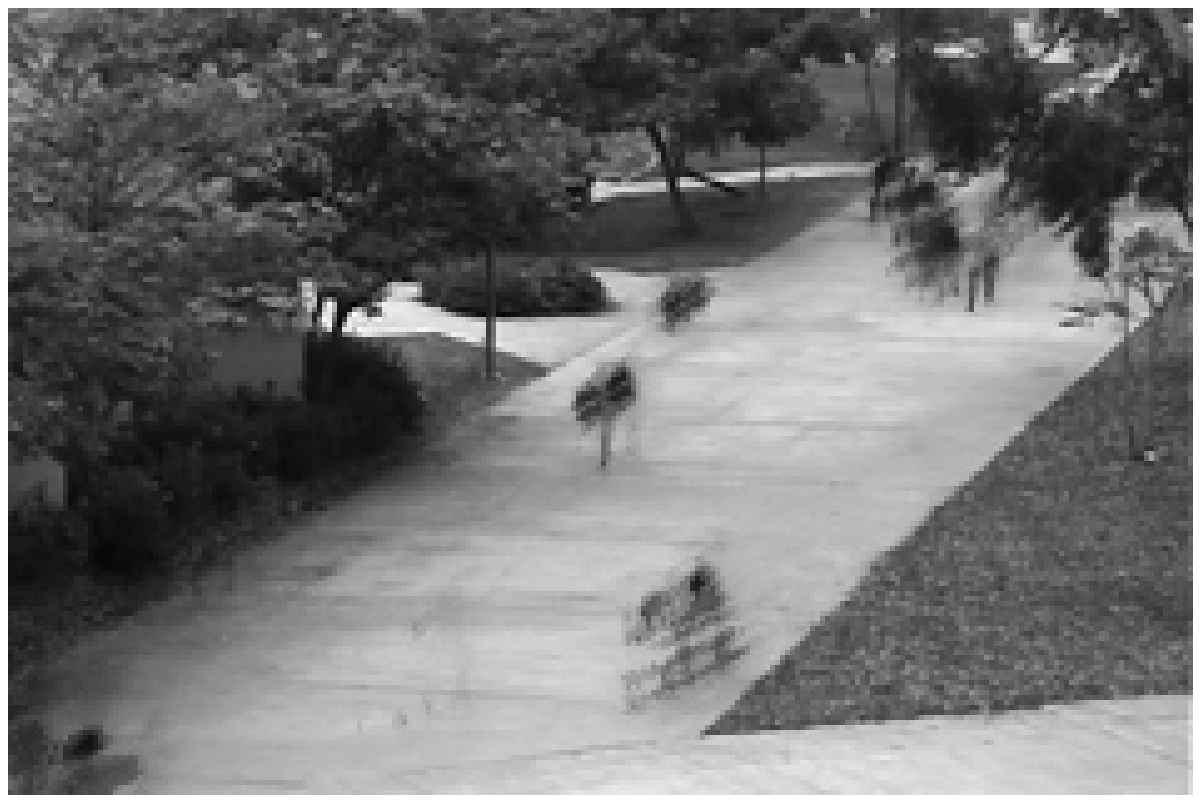}}
    \centerline{\includegraphics[width=\linewidth]{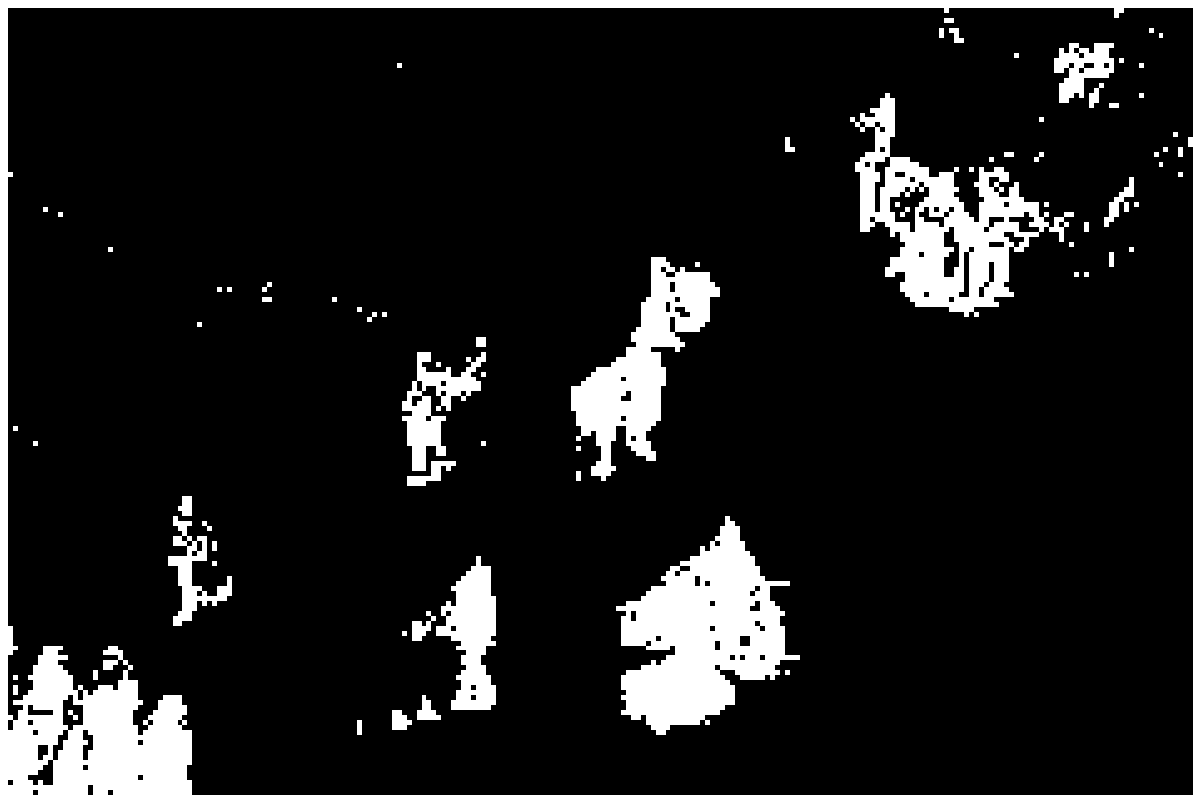}}
\end{minipage}
\begin{minipage}[t]{0.15\linewidth}
    \centerline{\includegraphics[width=\linewidth]{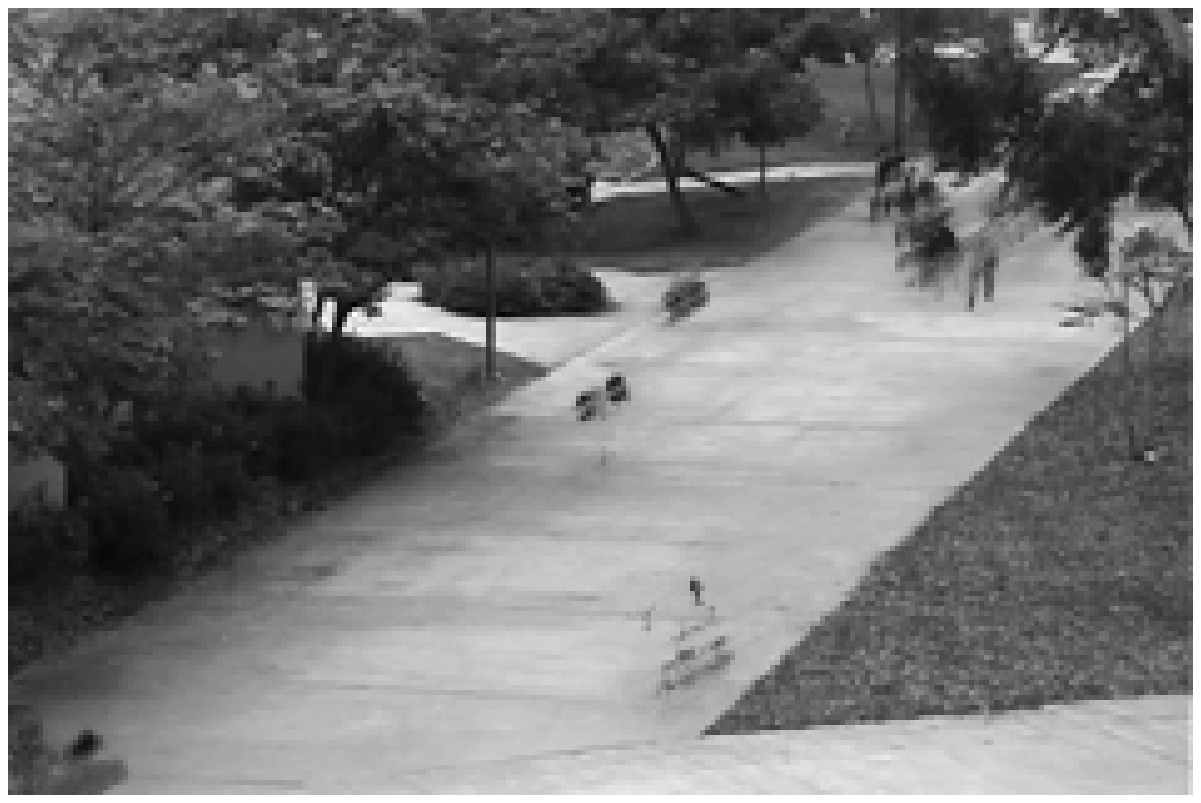}}
    \centerline{\includegraphics[width=\linewidth]{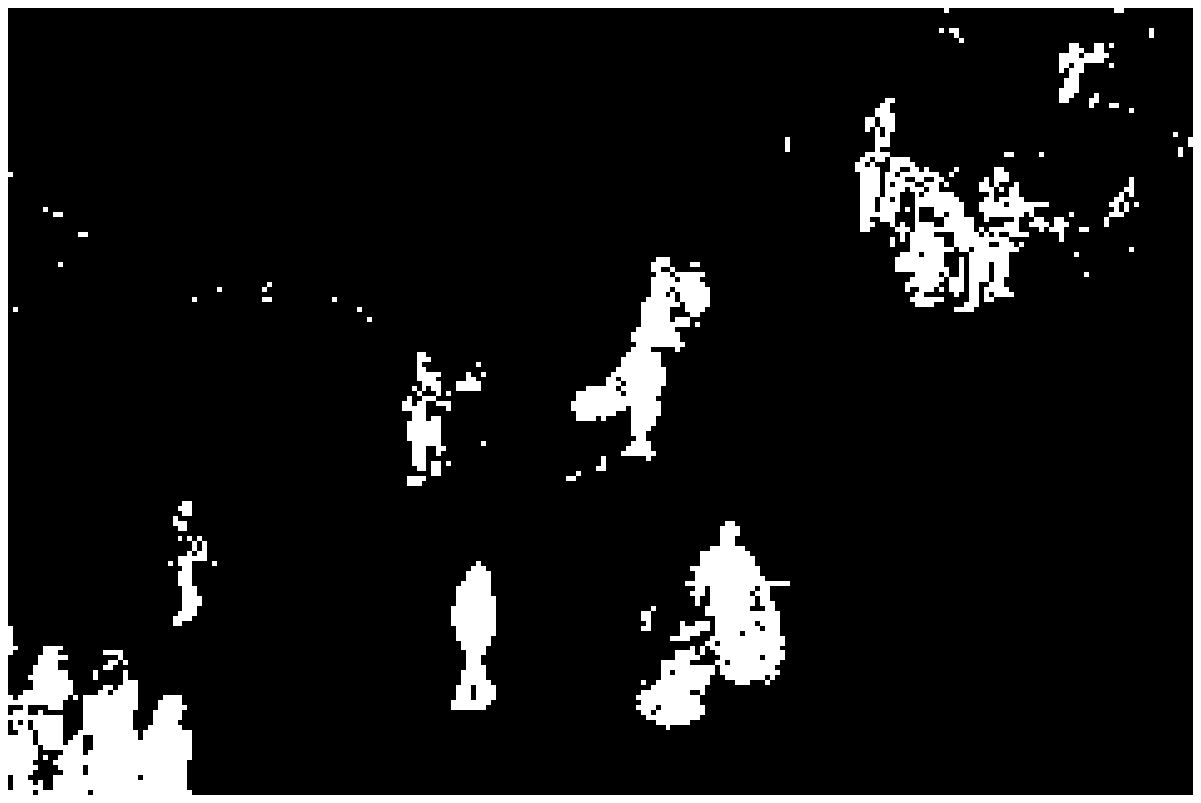}}
\end{minipage}
\begin{minipage}[t]{0.15\linewidth}
    \centerline{\includegraphics[width=\linewidth]{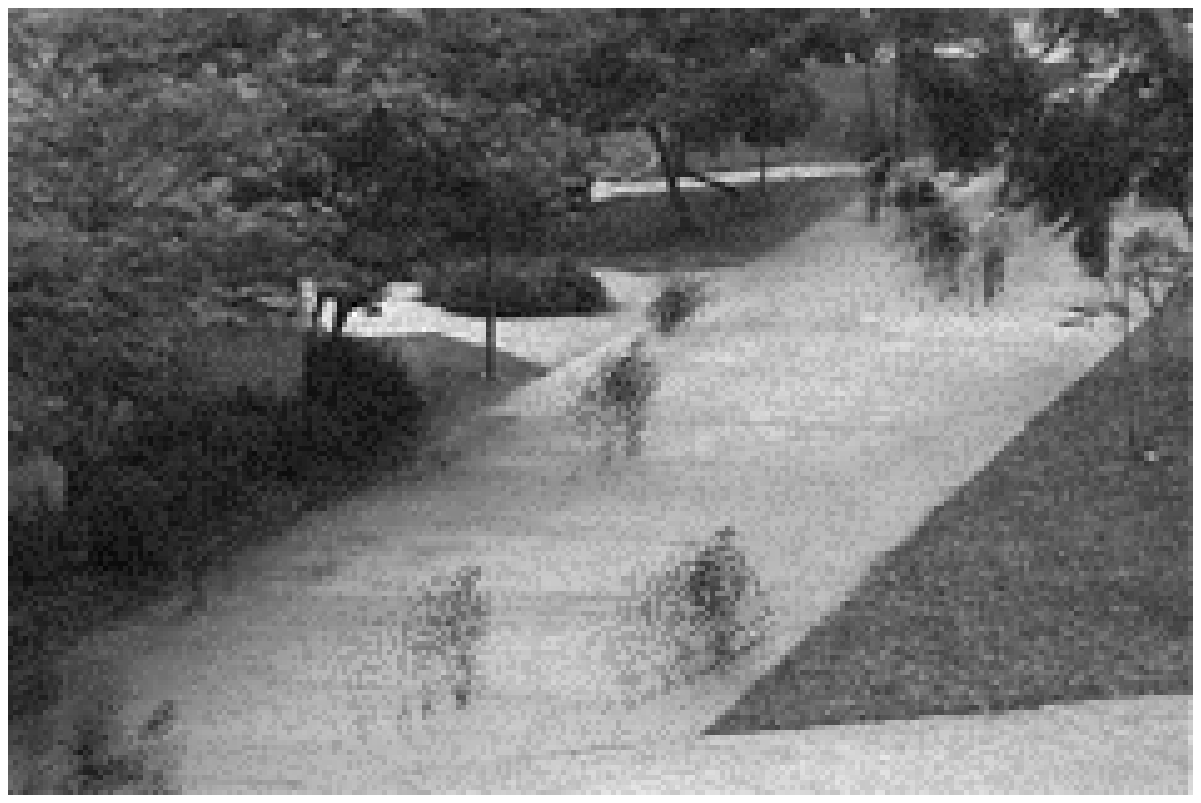}}
    \centerline{\includegraphics[width=\linewidth]{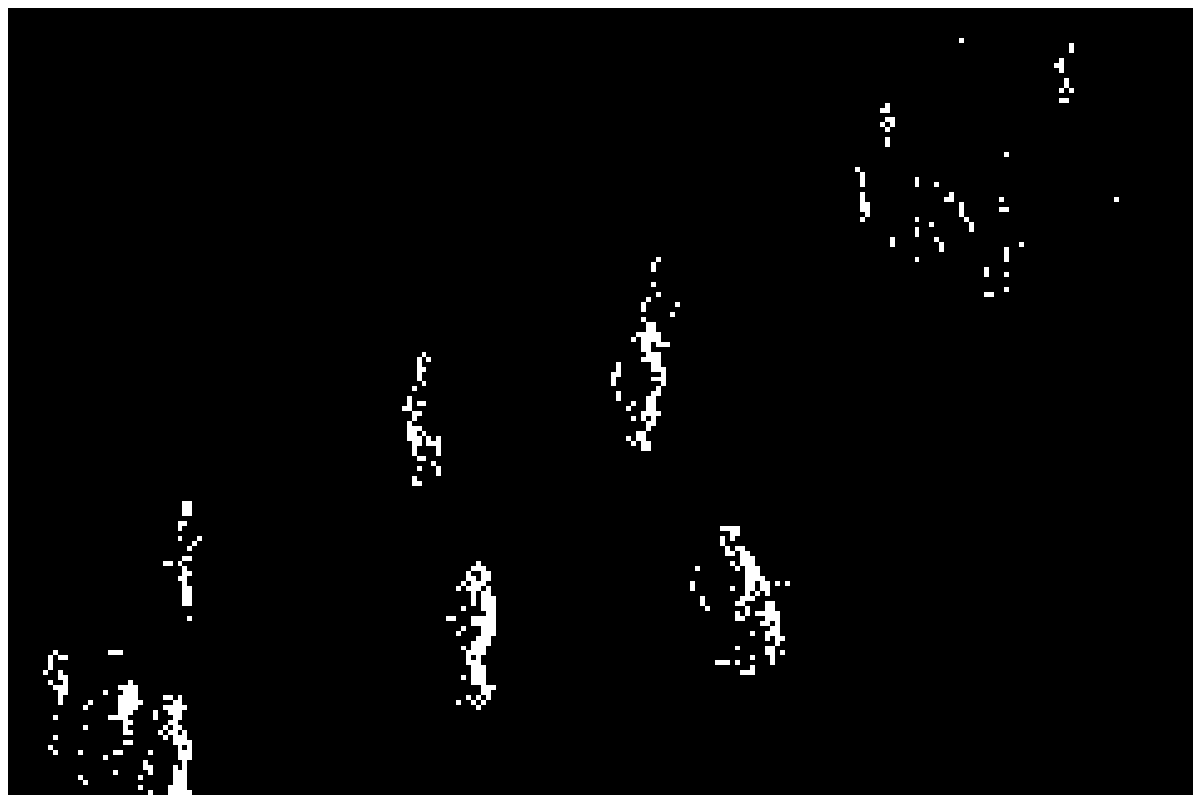}}
\end{minipage}
\centerline{(b)} 
\begin{minipage}[t]{0.15\linewidth}
    \centerline{\includegraphics[width=\linewidth]{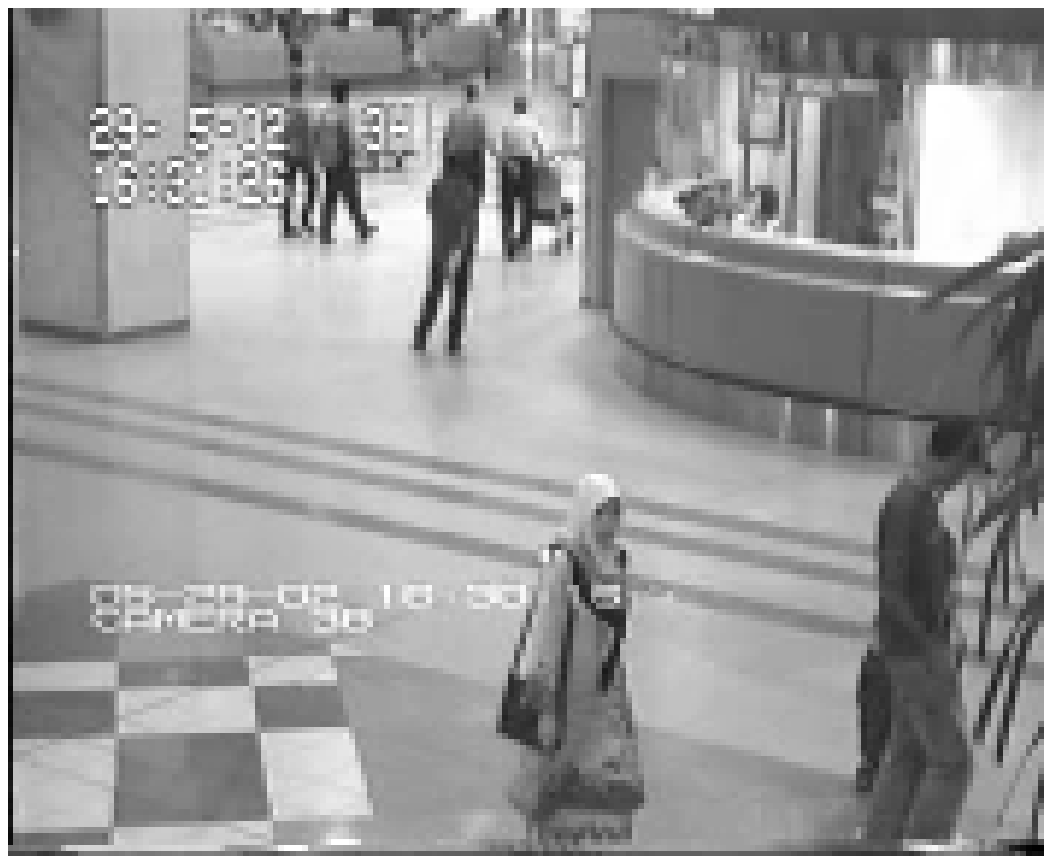}}
    \centerline{\includegraphics[width=\linewidth]{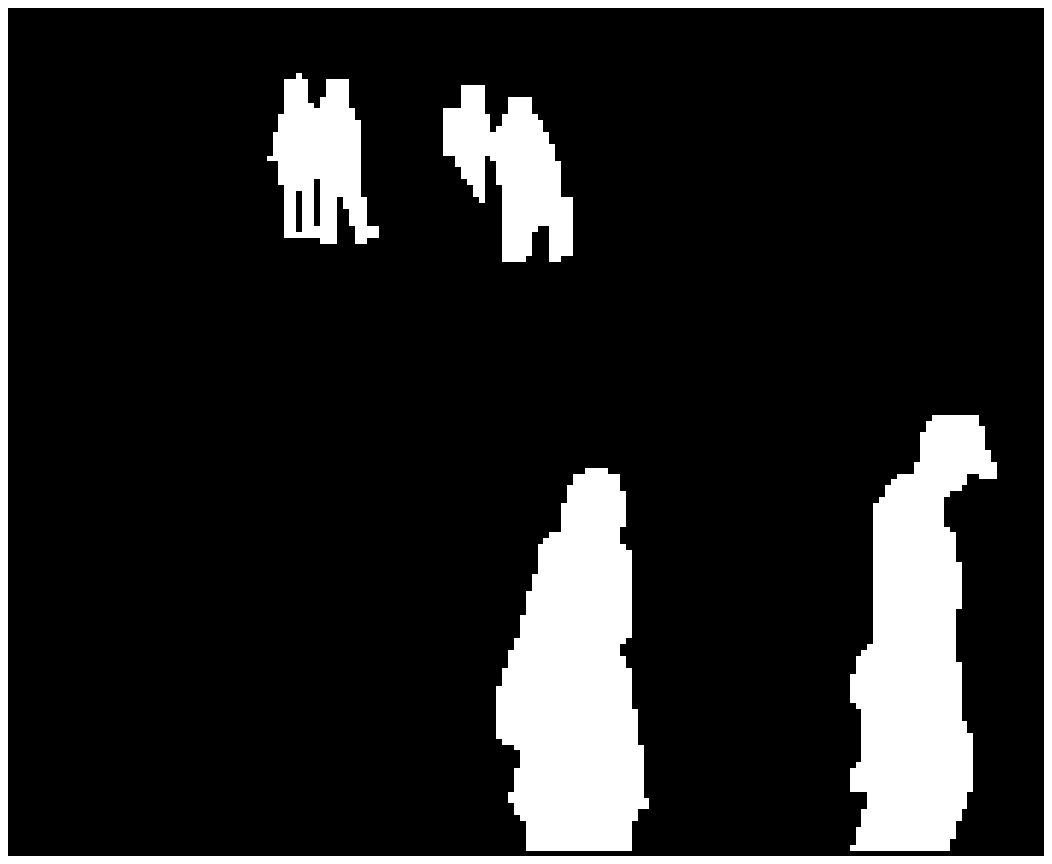}}
\end{minipage}
\hspace{5mm}
\begin{minipage}[t]{0.15\linewidth}
    \centerline{\includegraphics[width=\linewidth]{figures/hall_B3_DECOLOR.eps}}
    \centerline{\includegraphics[width=\linewidth]{figures/hall_S3_DECOLOR.eps}}
\end{minipage}
\begin{minipage}[t]{0.15\linewidth}
    \centerline{\includegraphics[width=\linewidth]{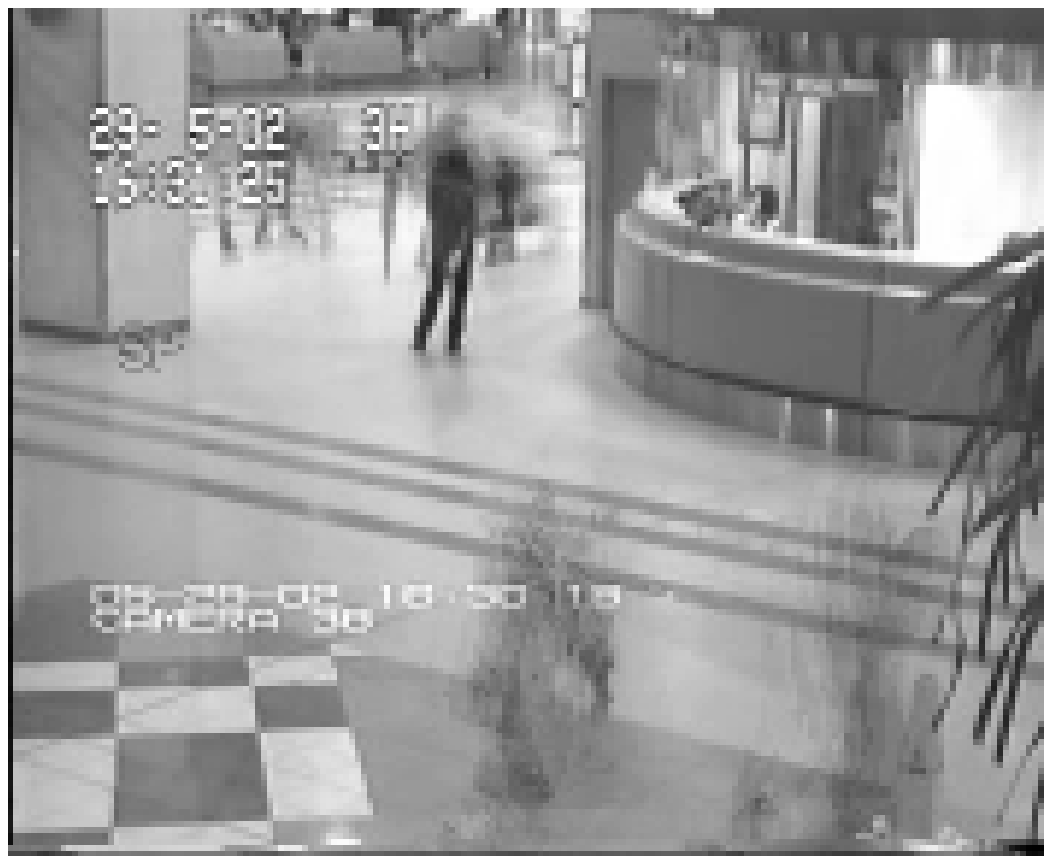}}
    \centerline{\includegraphics[width=\linewidth]{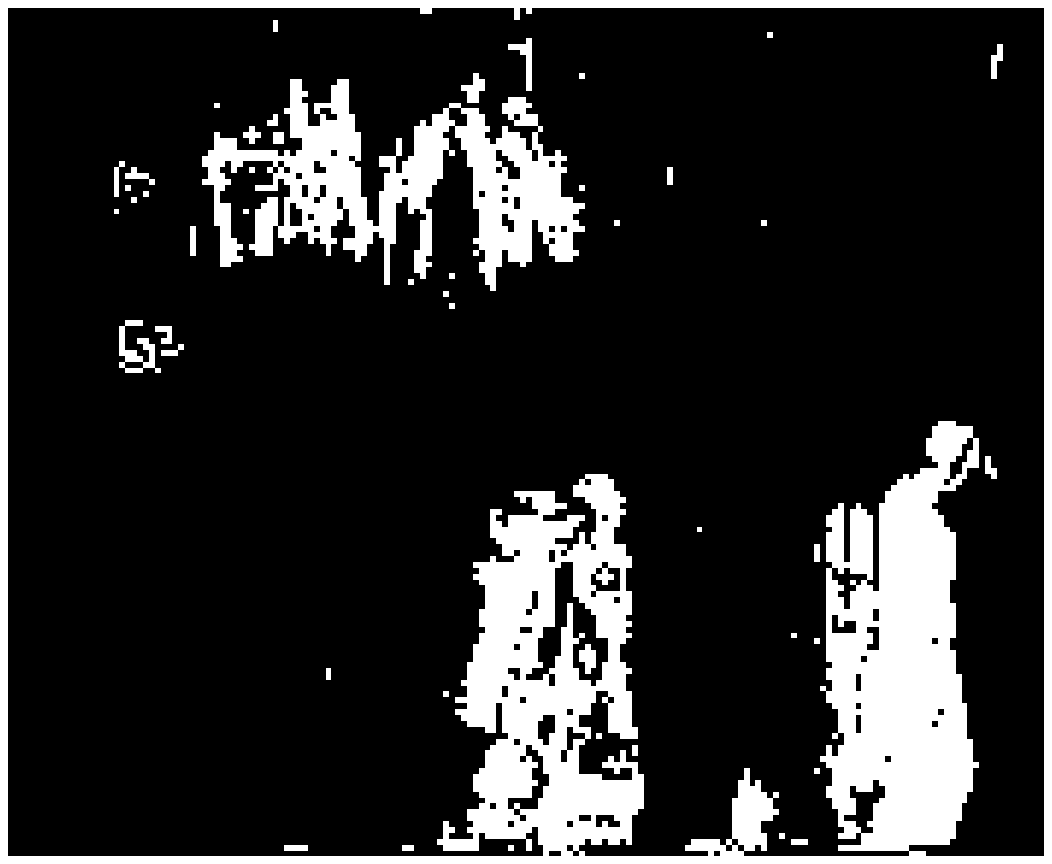}}
\end{minipage}
\begin{minipage}[t]{0.15\linewidth}
    \centerline{\includegraphics[width=\linewidth]{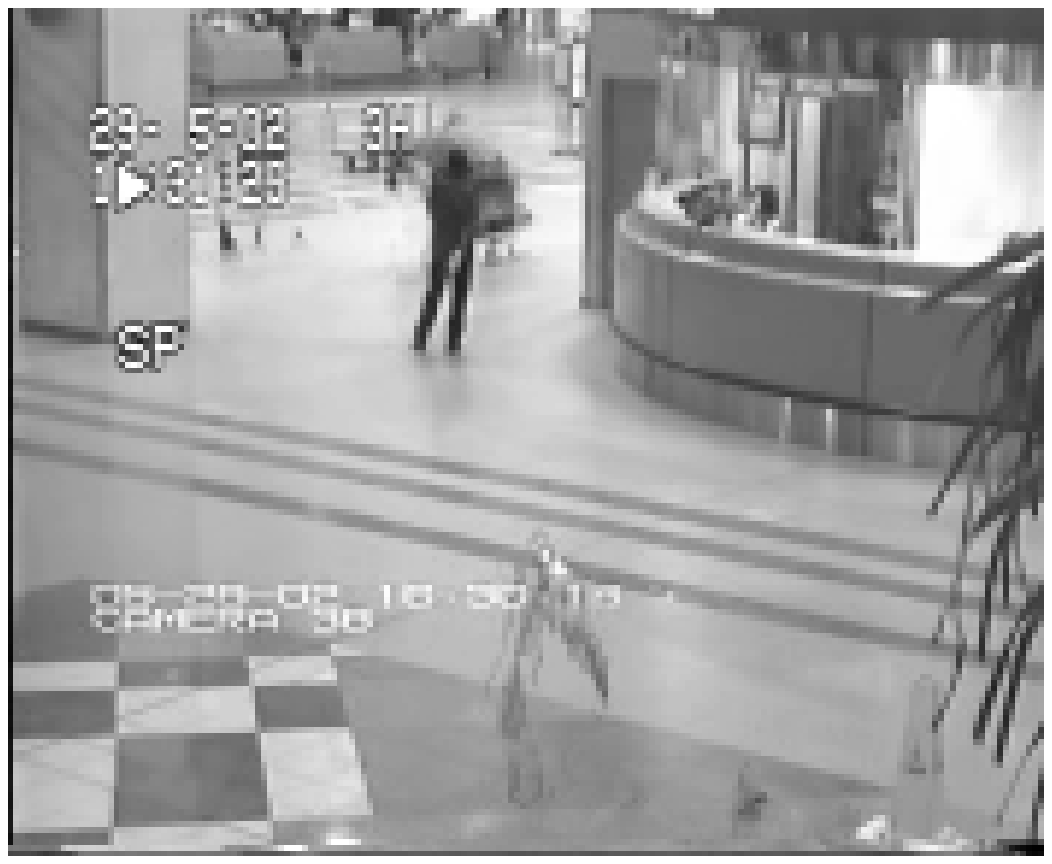}}
    \centerline{\includegraphics[width=\linewidth]{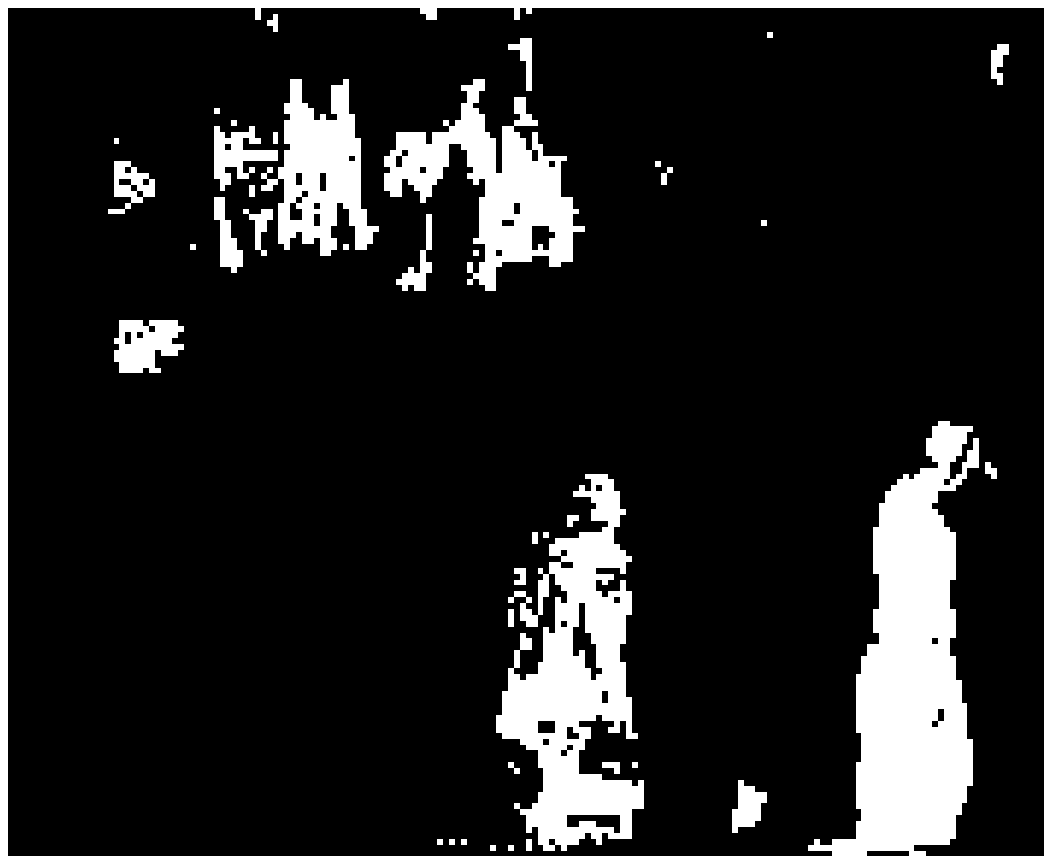}}
\end{minipage}
\begin{minipage}[t]{0.15\linewidth}
    \centerline{\includegraphics[width=\linewidth]{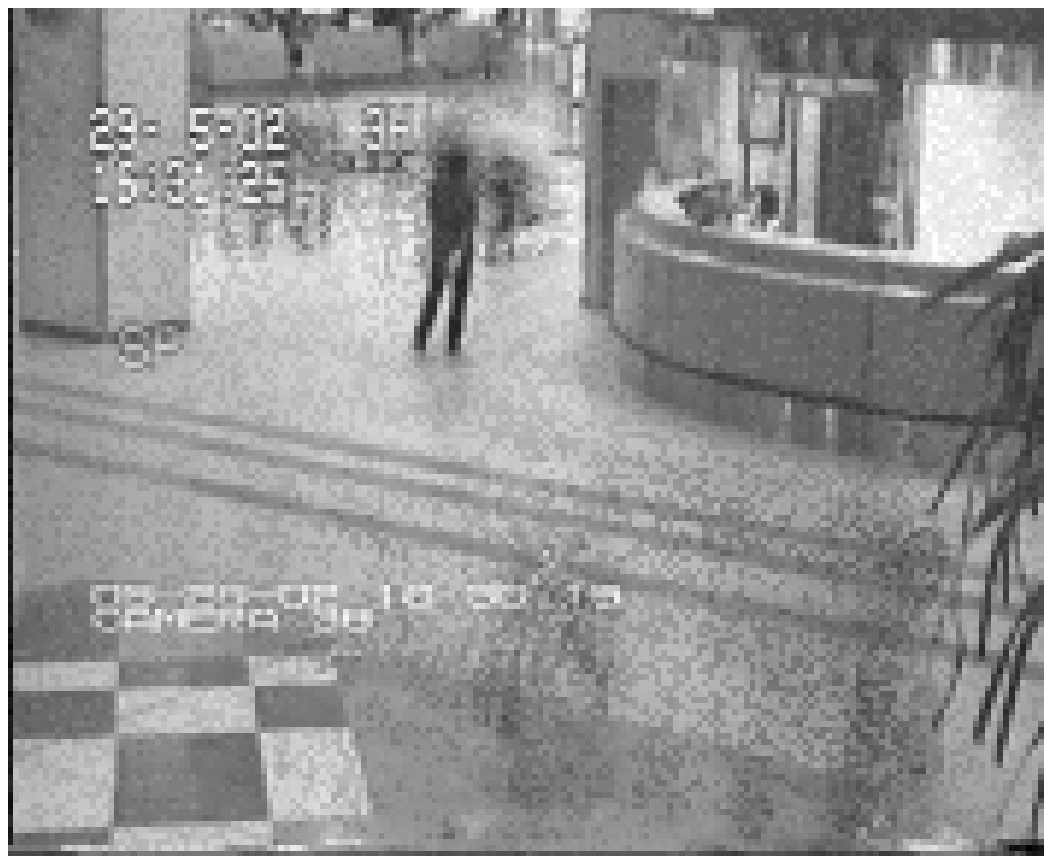}}
    \centerline{\includegraphics[width=\linewidth]{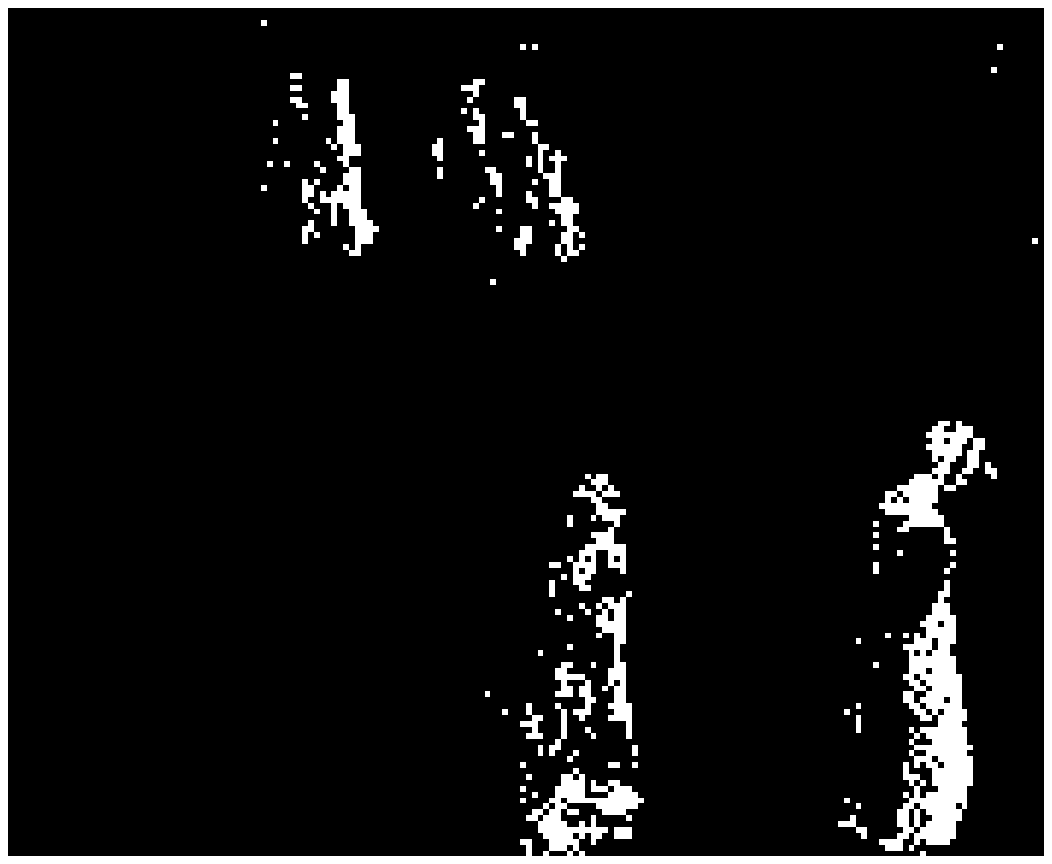}}
\end{minipage}
\centerline{(c)} 
\begin{minipage}[t]{0.15\linewidth}
    \centerline{\includegraphics[width=\linewidth]{figures/airport_D3.eps}}
    \centerline{\includegraphics[width=\linewidth]{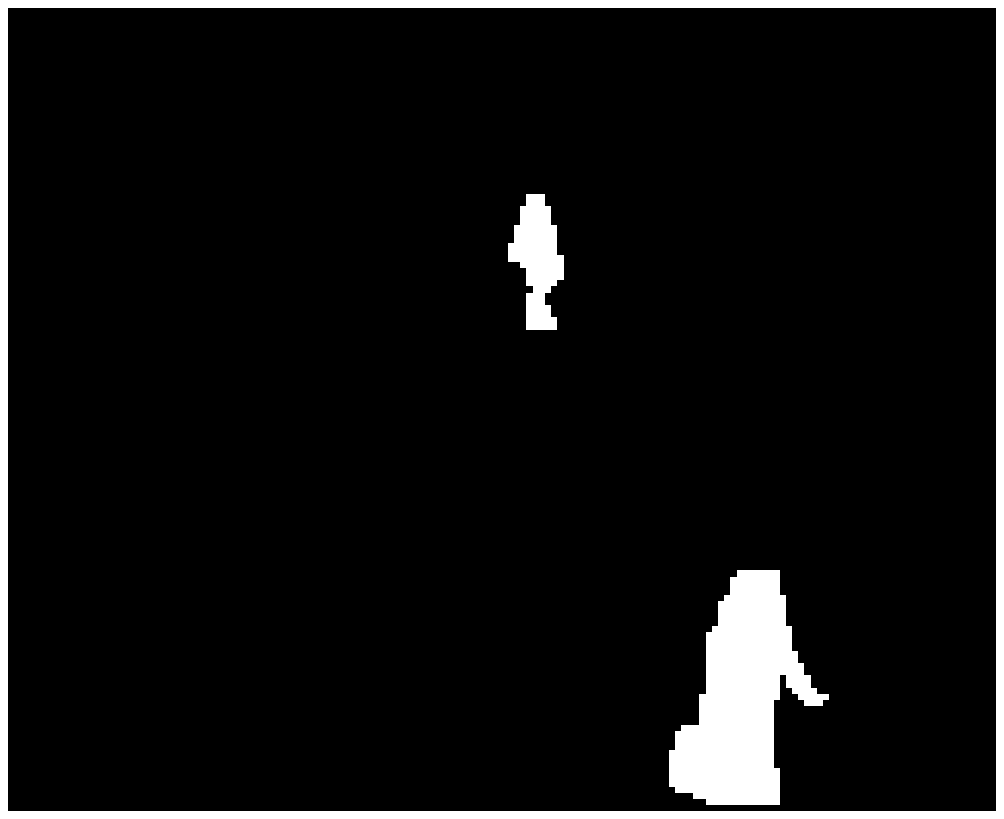}}
\end{minipage}
\hspace{5mm}
\begin{minipage}[t]{0.15\linewidth}
    \centerline{\includegraphics[width=\linewidth]{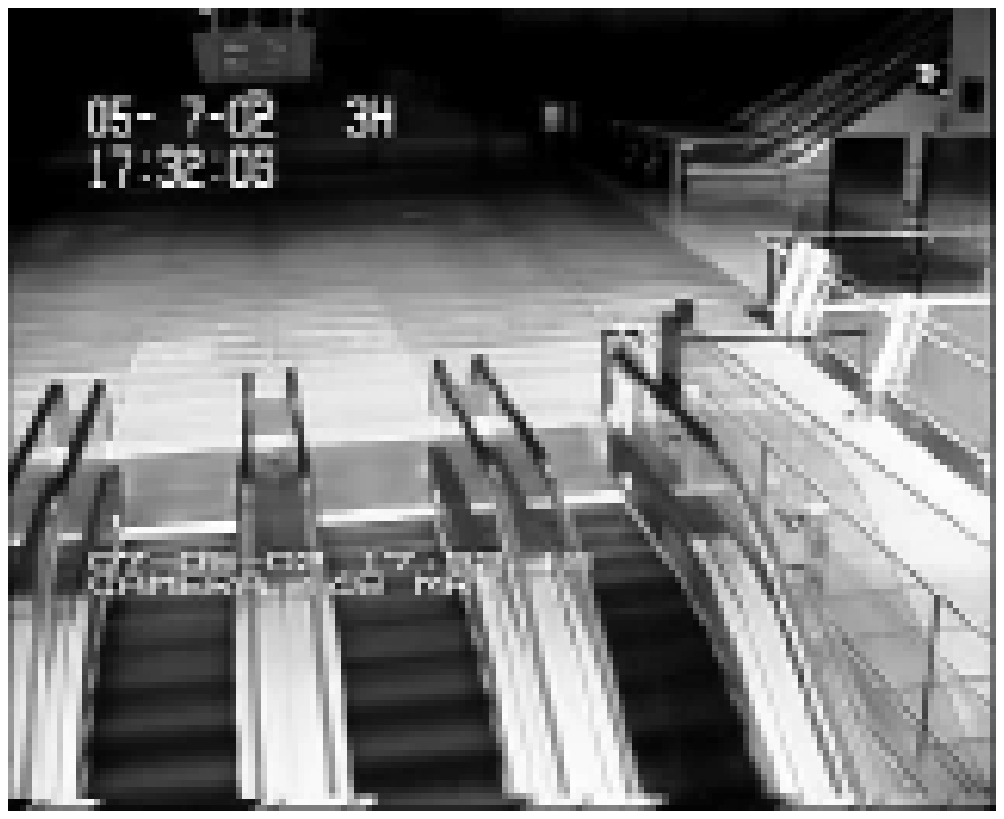}}
    \centerline{\includegraphics[width=\linewidth]{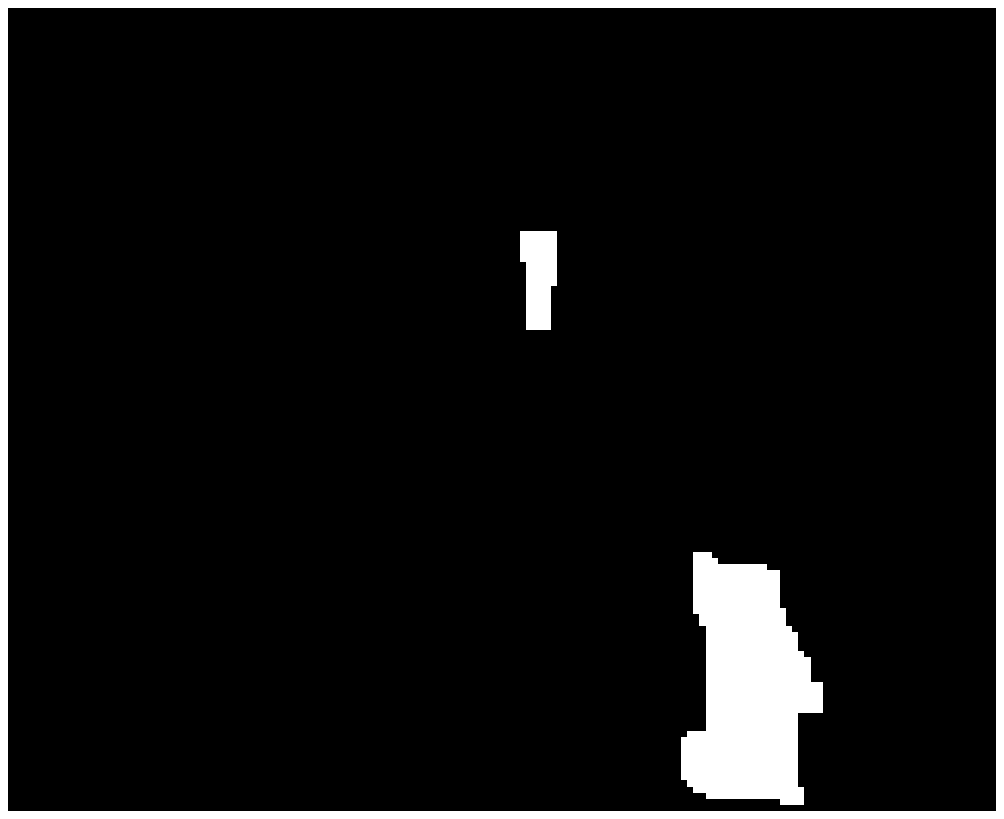}}
\end{minipage}
\begin{minipage}[t]{0.15\linewidth}
    \centerline{\includegraphics[width=\linewidth]{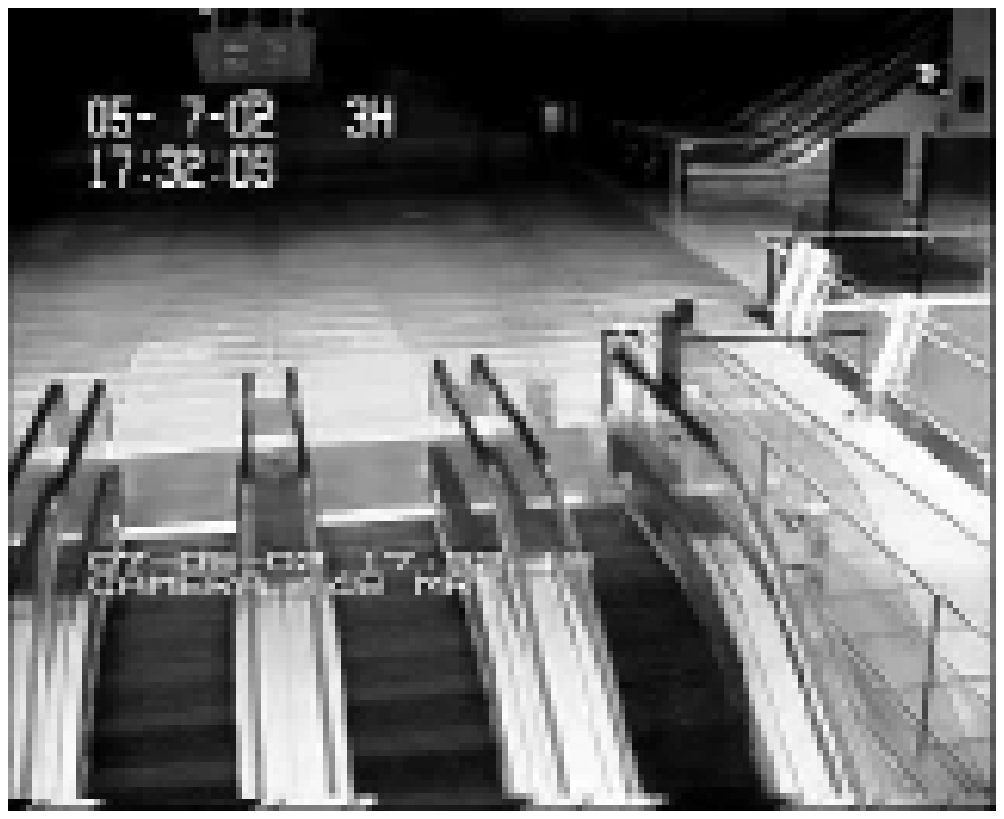}}
    \centerline{\includegraphics[width=\linewidth]{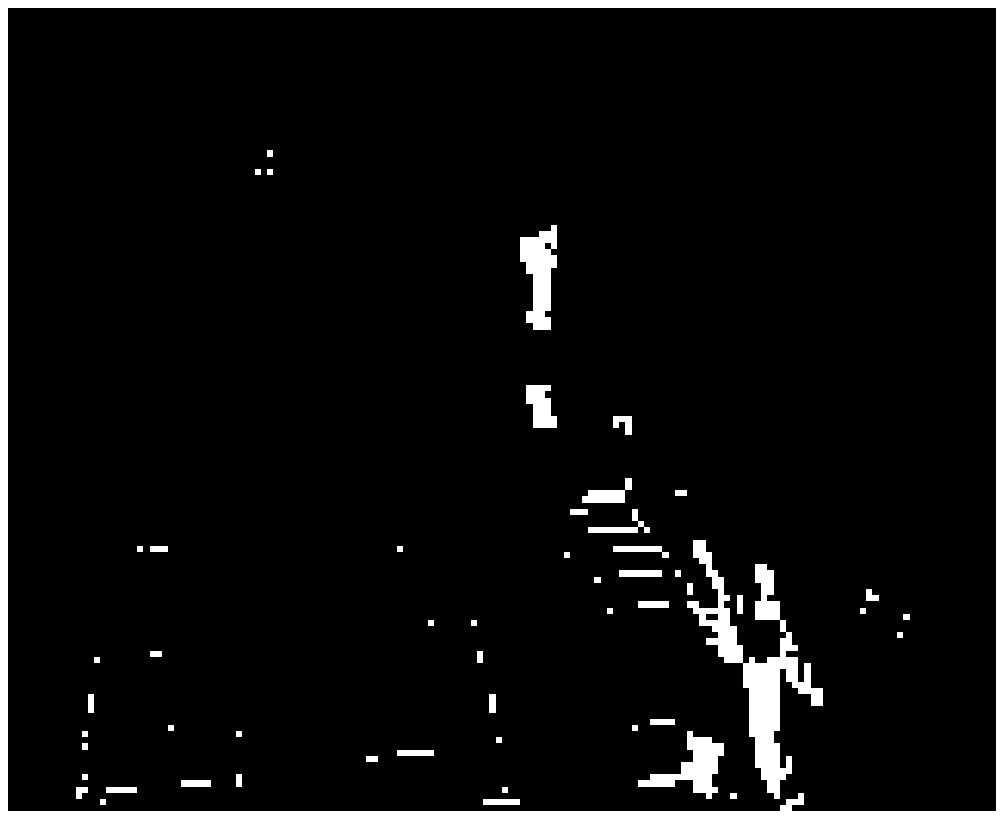}}
\end{minipage}
\begin{minipage}[t]{0.15\linewidth}
    \centerline{\includegraphics[width=\linewidth]{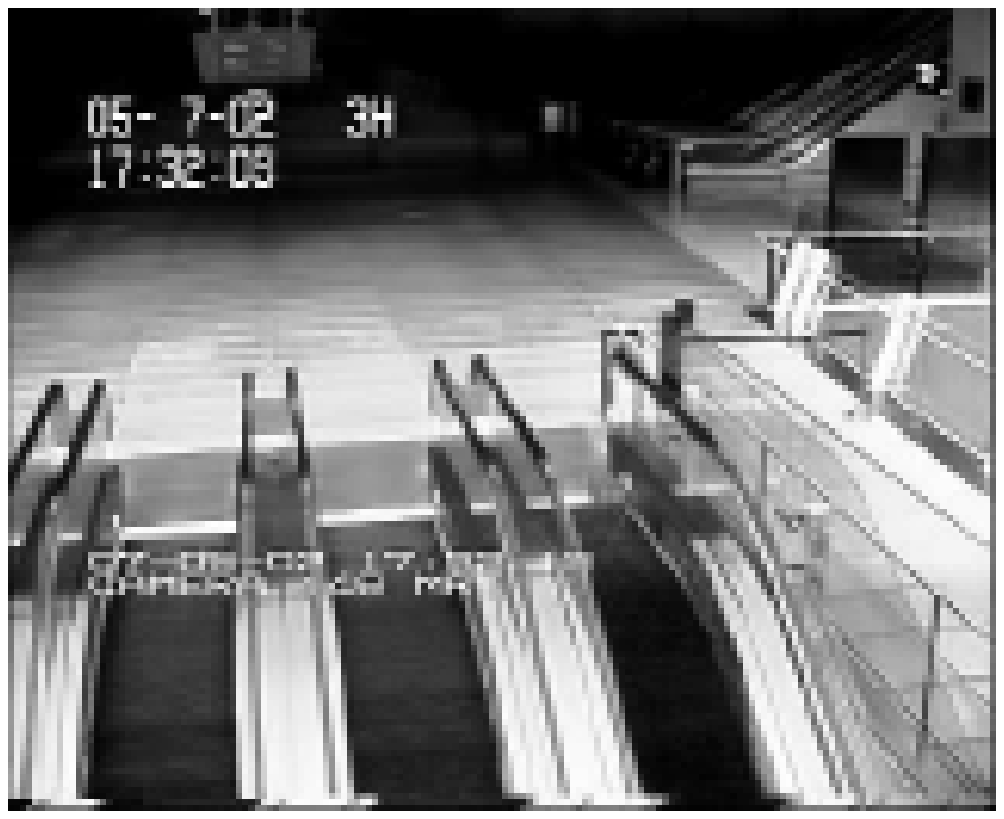}}
    \centerline{\includegraphics[width=\linewidth]{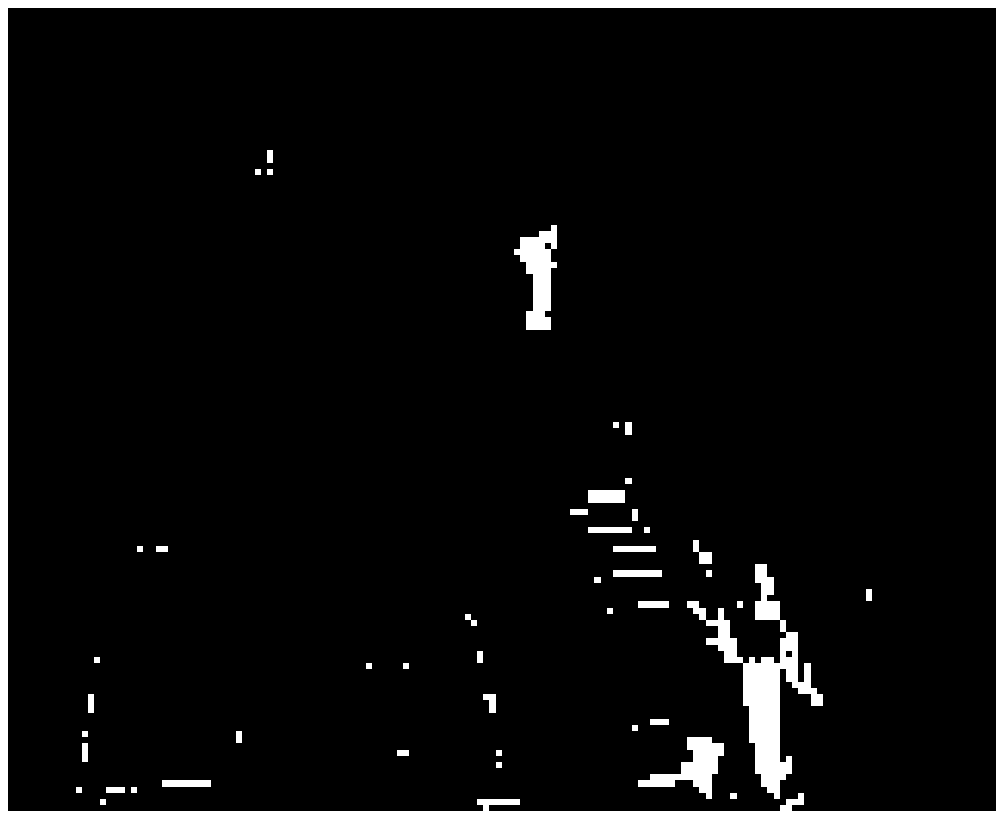}}
\end{minipage}
\begin{minipage}[t]{0.15\linewidth}
    \centerline{\includegraphics[width=\linewidth]{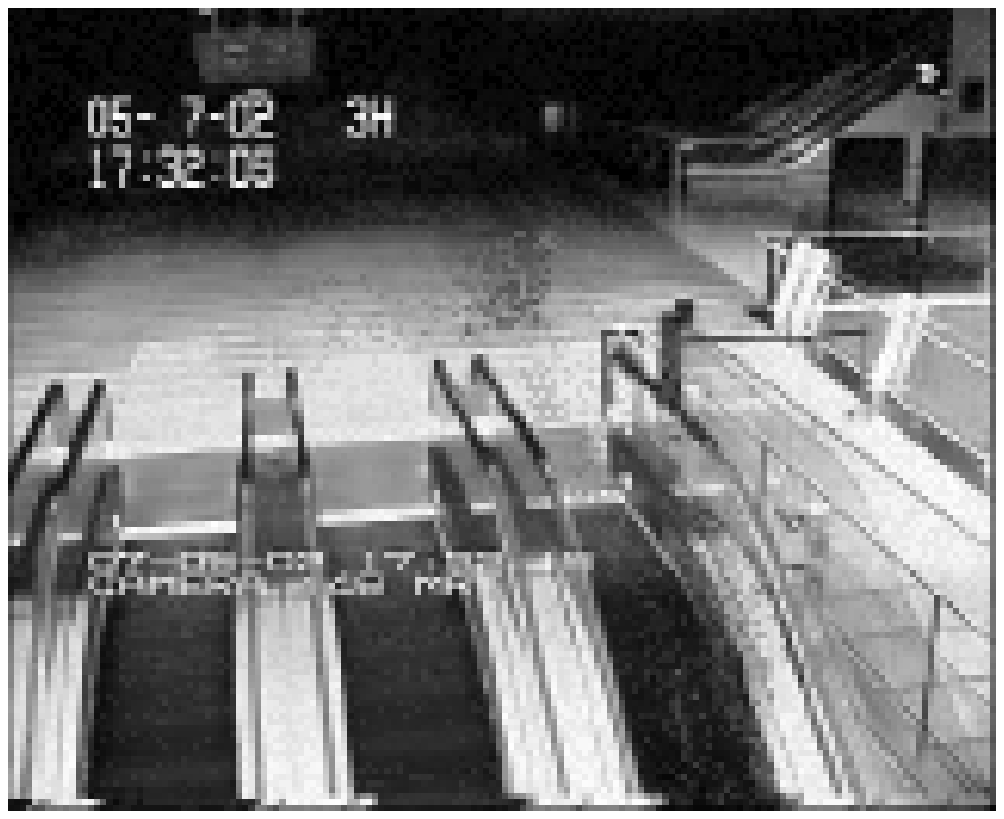}}
    \centerline{\includegraphics[width=\linewidth]{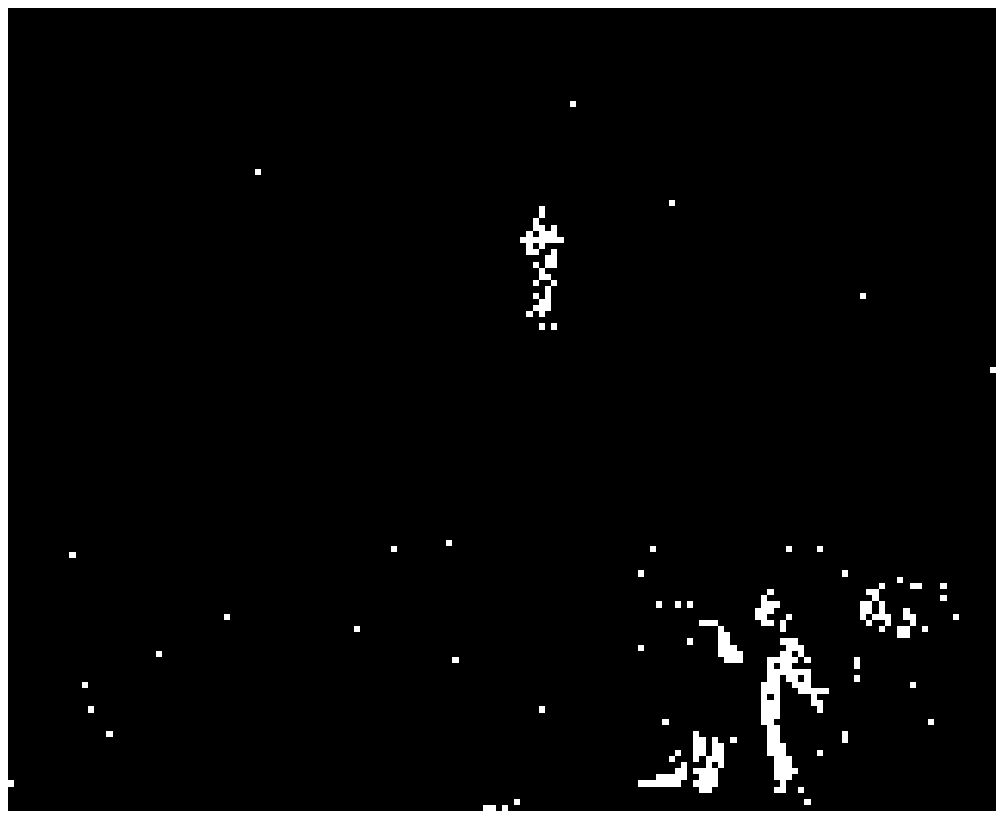}}
\end{minipage}
\centerline{(d)} 
\begin{minipage}[t]{0.15\linewidth}
    \centerline{\includegraphics[width=\linewidth]{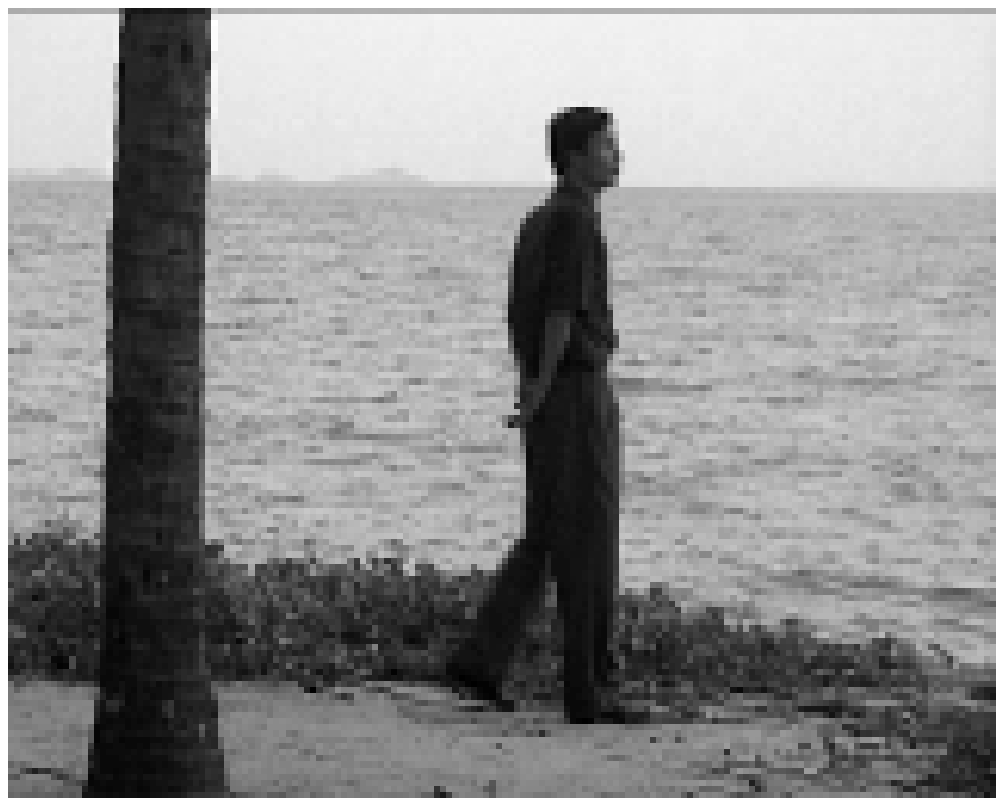}}
    \centerline{\includegraphics[width=\linewidth]{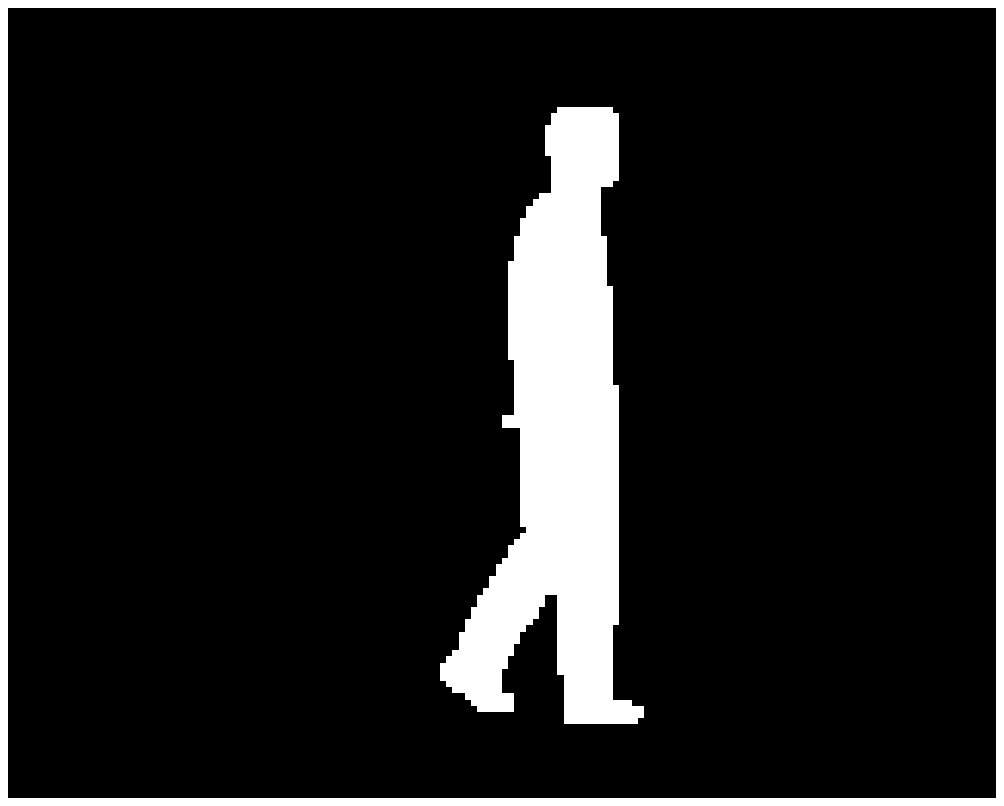}}
\end{minipage}
\hspace{5mm}
\begin{minipage}[t]{0.15\linewidth}
    \centerline{\includegraphics[width=\linewidth]{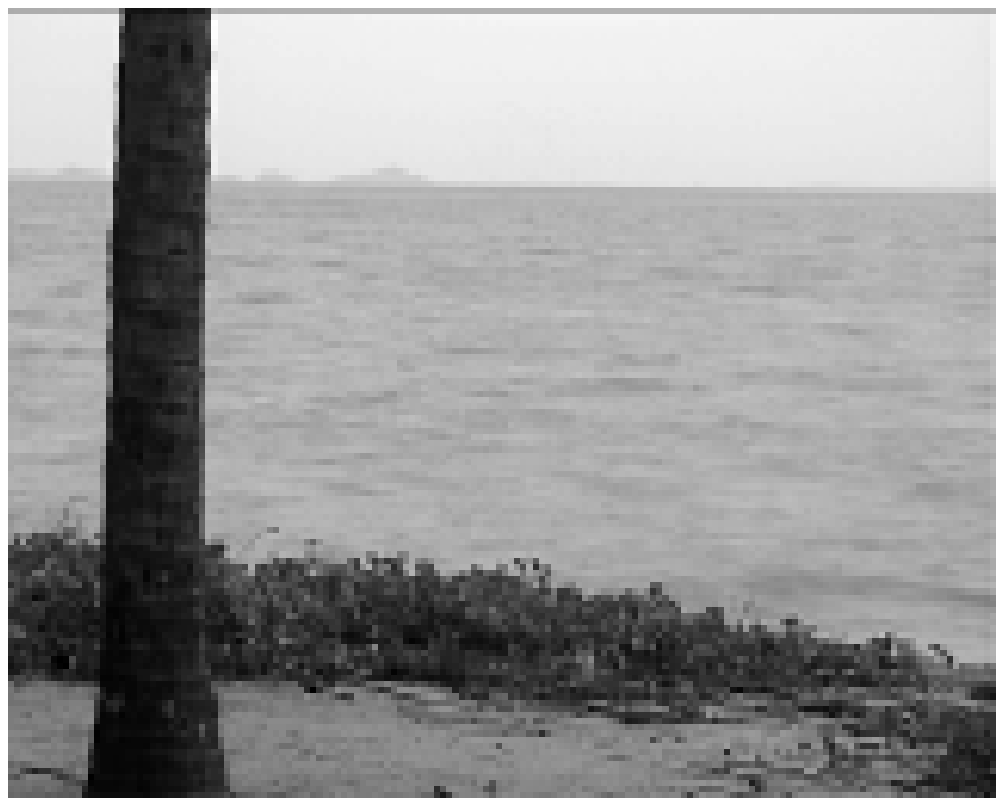}}
    \centerline{\includegraphics[width=\linewidth]{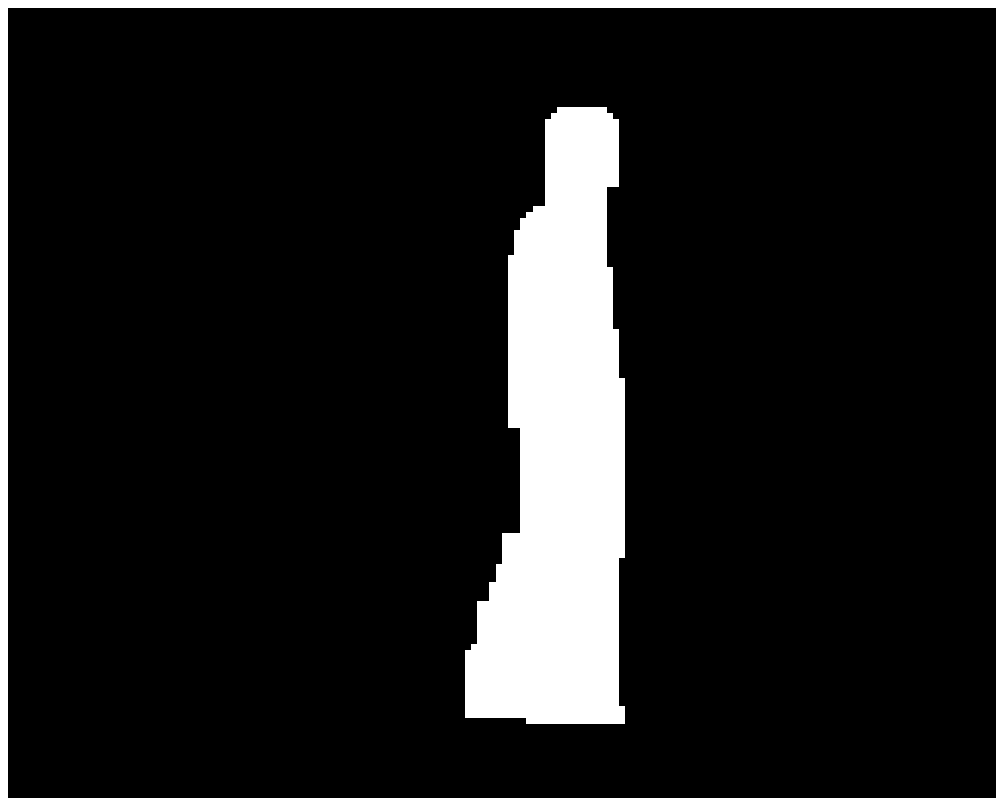}}
\end{minipage}
\begin{minipage}[t]{0.15\linewidth}
    \centerline{\includegraphics[width=\linewidth]{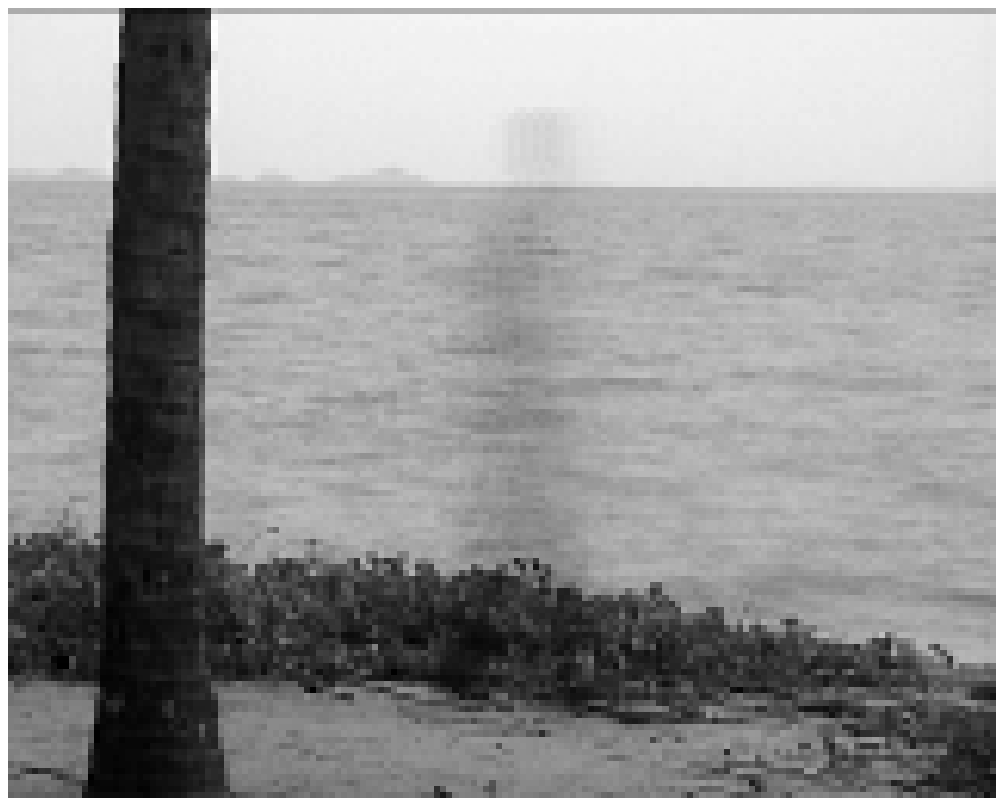}}
    \centerline{\includegraphics[width=\linewidth]{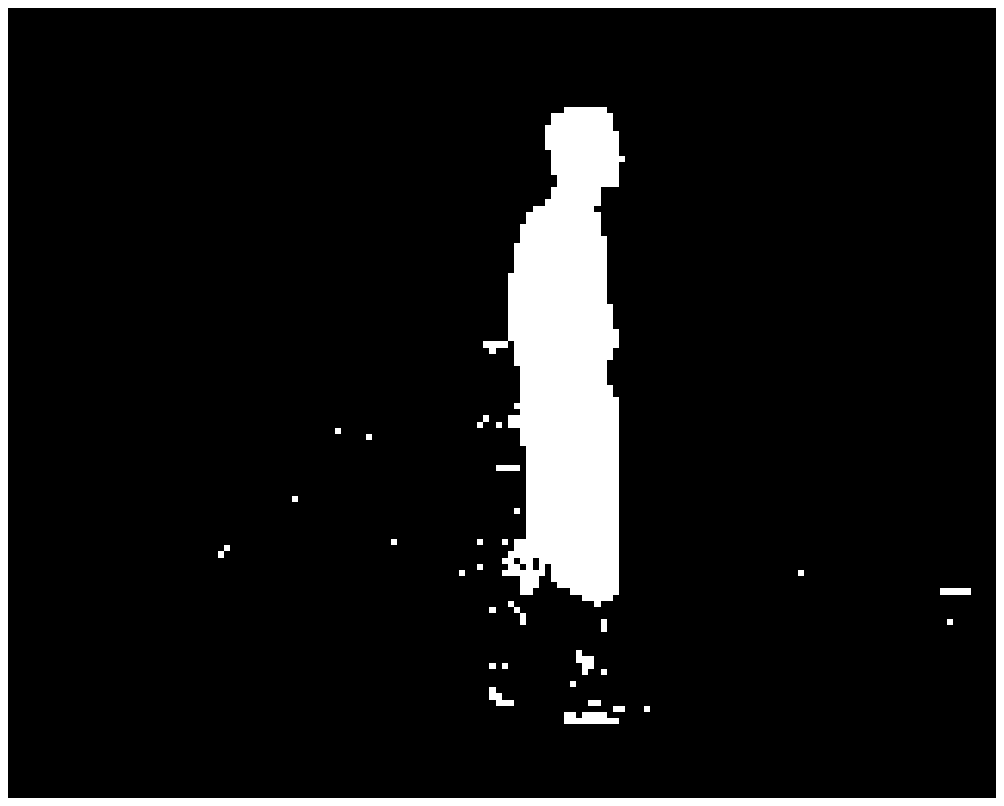}}
\end{minipage}
\begin{minipage}[t]{0.15\linewidth}
    \centerline{\includegraphics[width=\linewidth]{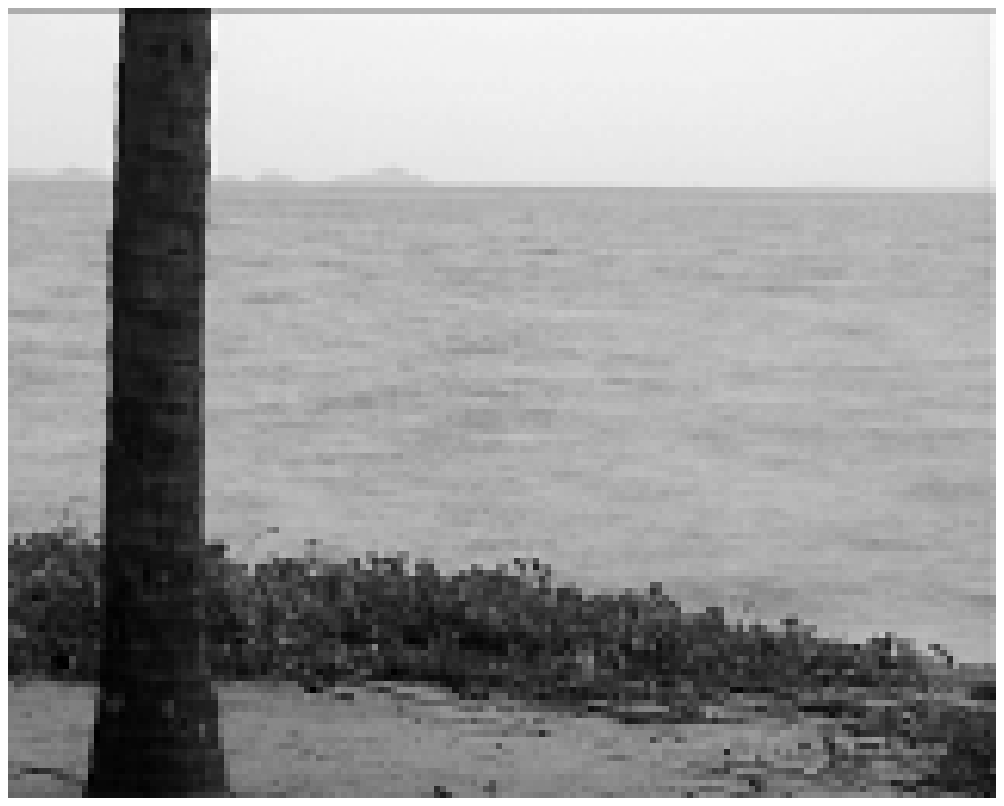}}
    \centerline{\includegraphics[width=\linewidth]{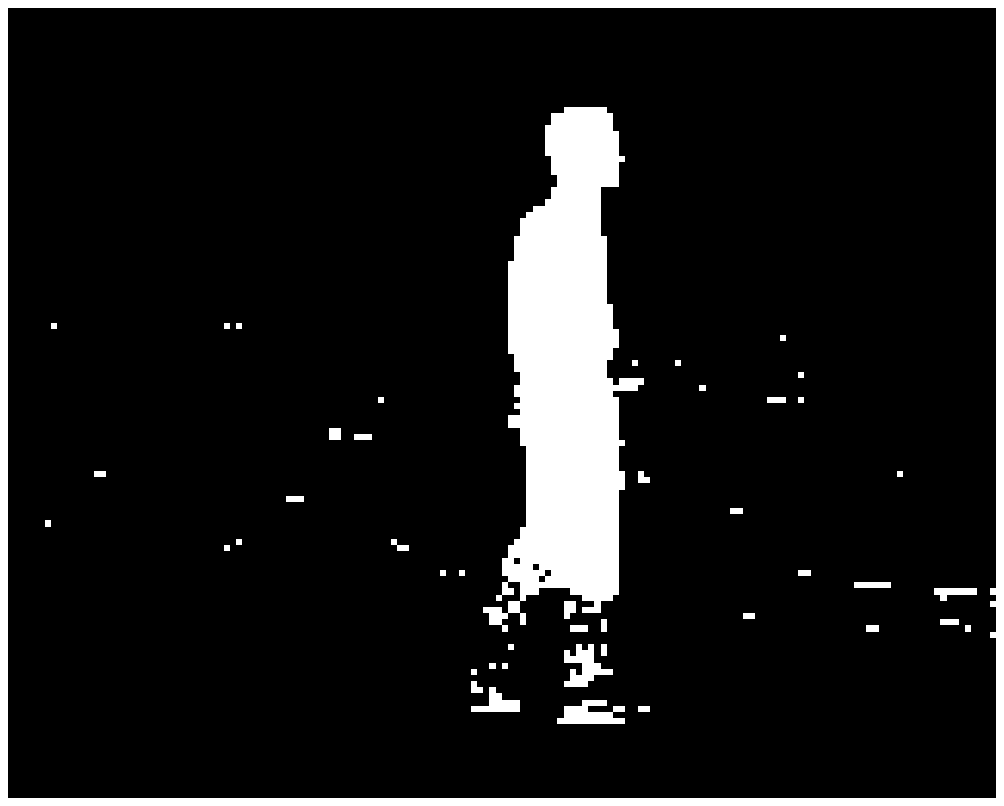}}
\end{minipage}
\begin{minipage}[t]{0.15\linewidth}
    \centerline{\includegraphics[width=\linewidth]{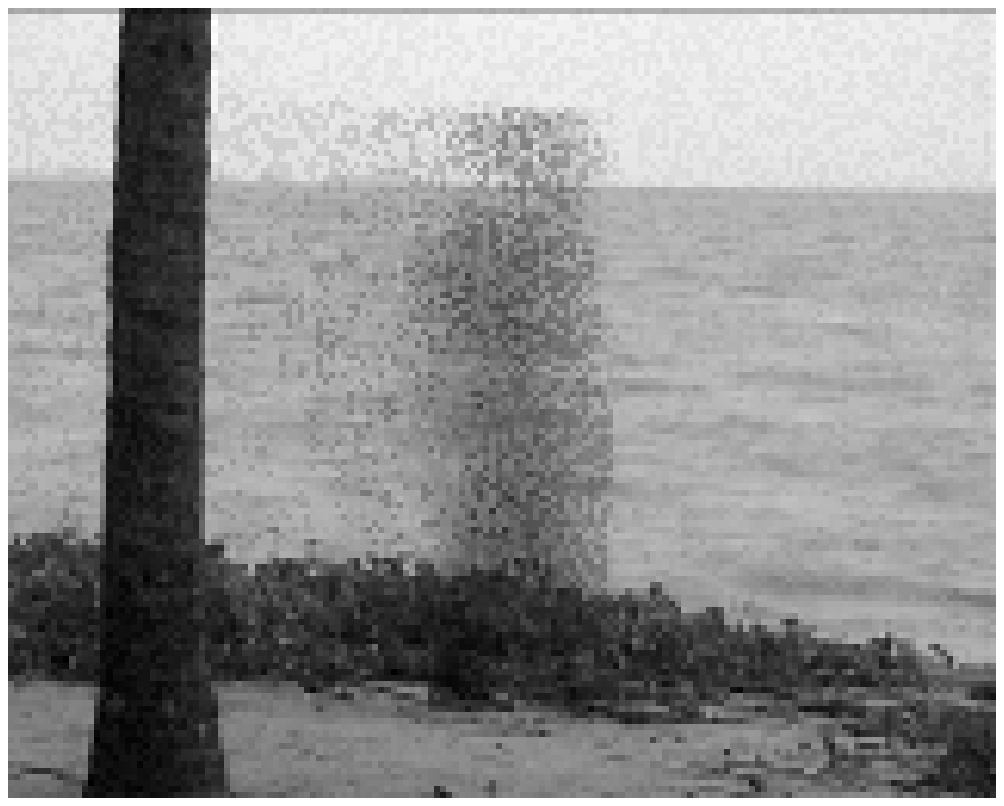}}
    \centerline{\includegraphics[width=\linewidth]{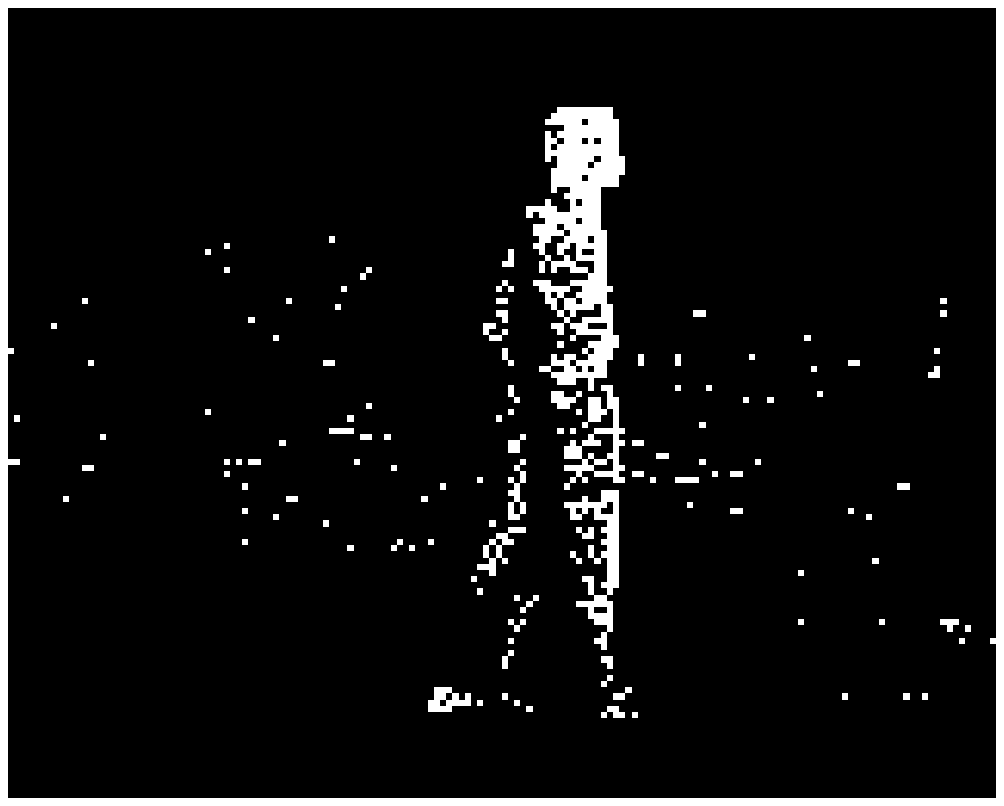}}
\end{minipage}
\centerline{(e)} 
\caption{Five sub-sequences of surveillance videos. Sequence information is given in Table \ref{Tab_data}. The last frame of each sequence and its manual segmentation are shown in Column 1. The corresponding results by four methods are presented from Column 2 to Column 5, respectively. The top panel is the estimated background and the bottom panel is the segmentation.}\label{Fig_BackSub}
\end{figure*}

In this part, we test DECOLOR on several real sequences selected from public datasets of background subtraction. Since we aim to evaluate the ability of algorithms in detecting moving objects at the start of videos, we focus on short clips composed of beginning frames of videos. All examples in Fig. \ref{Fig_BackSub} have only 24 or 48 frames corresponding to 1 or 2 seconds for a frame rate of 24 fps. We compare DECOLOR with three methods that are simple in implementation but effective in practice. The first one is PCP \cite{Candes09}, which is the state-of-the-art algorithm for RPCA. The second method is median filtration, a baseline method for unimodal background modeling. The median intensity value around each pixel is computed forming a background image. Then, each frame is subtracted by the background image and the difference is thresholded to generate a foreground mask. The advantage of using median rather than mean is that it is a more robust estimator to avoid blending pixel values, which is more proper for background estimation \cite{gutchess2001background}. The third method is mixture of Gaussians (MoG) \cite{stauffer1999adaptive}. It is popularly used for multimodal background modeling and has proven to be very competitive compared with other more sophisticated techniques for background subtraction \cite{piccardi2004background,parks2008evaluation}.

The sequences and results are presented in Fig. \ref{Fig_BackSub}. The first example shows an office with two people walking around. Although the objects are large and always presented in all frames, DECOLOR recovers the background and outputs a foreground mask accurately. Notice that the results are direct outputs of Algorithm \ref{Alg_overall} without any postprocessing. The results of PCP are relatively unsatisfactory. Ghosts of foreground remain in the recovered background. This is because the $\ell_1$-penalty used in PCP is not robust enough to remove the influence of contiguous occlusion. Such corruption of extracted background will result in false detections as shown in the segmentation result. Moreover, without the smoothness constraint, occasional light changes (\eg near the boundary of fluorescent lamps) or video noises give rise to small pieces of falsely detected regions. The results of median filtration depend on how long each pixel is taken by foreground. Thus, from the recovered background of median filtration we can find that the man near the door is clearly removed while the man turning at the corner leaves a ghost. Despite of scattered artifacts, MoG gives less false positives due to its multimodal modeling of background. However, blending of foreground intensity can be seen obviously in the recovered background, which results in more false negatives in the foreground mask, \eg the interior region of objects. Similar results can be found in next two examples.

The last two examples include dynamic background. Fig. \ref{Fig_BackSub}(d) presents a sequence clipped from a surveillance video of an airport, which is very challenging because the background involves a running escalator. Although the escalator is moving, it is recognized as a part of background by DECOLOR since its periodical motion gives repeated patterns. As we can see, the structure of the escalator is maintained in the background recovered by DECOLOR or PCP. This demonstrates the ability of low-rank representation to model dynamic background. Fig. \ref{Fig_BackSub}(e) gives another example with a water surface as background. Similarly, the low-rank modeling of background gives better results with less false detections on the water surface, and DECOLOR obtains a cleaner background compared against PCP.

We also give a quantitative evaluation for the segmentation results shown in Fig. \ref{Fig_BackSub}. The manual annotation is used as ground truth and the F-measure is calculated. As shown in Table \ref{Tab_bgsub}, DECOLOR outperforms other approaches on all sequences.

\begin{table}
\renewcommand{\arraystretch}{1.3}
\caption{Quantitative evaluation (F-measure) on the sequences shown in Fig. \ref{Fig_BackSub}.}
\label{Tab_bgsub}
\centering
\begin{tabular}{lcccc}
\toprule
Sequence & DECOLOR & PCP & Median & MoG \\
\hline
Fig. \ref{Fig_BackSub}(a) & 0.93 & 0.62 & 0.67 & 0.50 \\
Fig. \ref{Fig_BackSub}(b) & 0.82 & 0.66 & 0.71 & 0.35 \\
Fig. \ref{Fig_BackSub}(c) & 0.92 & 0.70 & 0.79 & 0.50 \\
Fig. \ref{Fig_BackSub}(d) & 0.82 & 0.49 & 0.51 & 0.36 \\
Fig. \ref{Fig_BackSub}(e) & 0.91 & 0.83 & 0.86 & 0.47 \\
\bottomrule
\end{tabular}
\end{table}

\subsubsection{Moving cameras}

\begin{figure*}
\centering
\begin{minipage}[t]{0.15\linewidth}
 \centerline{Image}
 \vspace{3mm}
\end{minipage}
\begin{minipage}[t]{0.15\linewidth}
 \centerline{Transformed}
\end{minipage}
\begin{minipage}[t]{0.15\linewidth}
 \centerline{Low-rank}
\end{minipage}
\begin{minipage}[t]{0.15\linewidth}
 \centerline{Segmentation}
\end{minipage}
\begin{minipage}[t]{0.15\linewidth}
 \centerline{Brox-Malik}
\end{minipage}
\begin{minipage}[t]{0.15\linewidth}
 \centerline{Truth}
\end{minipage}
\centering
 \includegraphics[width=0.15\linewidth]{figures/people1_D3.eps}
 \includegraphics[width=0.15\linewidth]{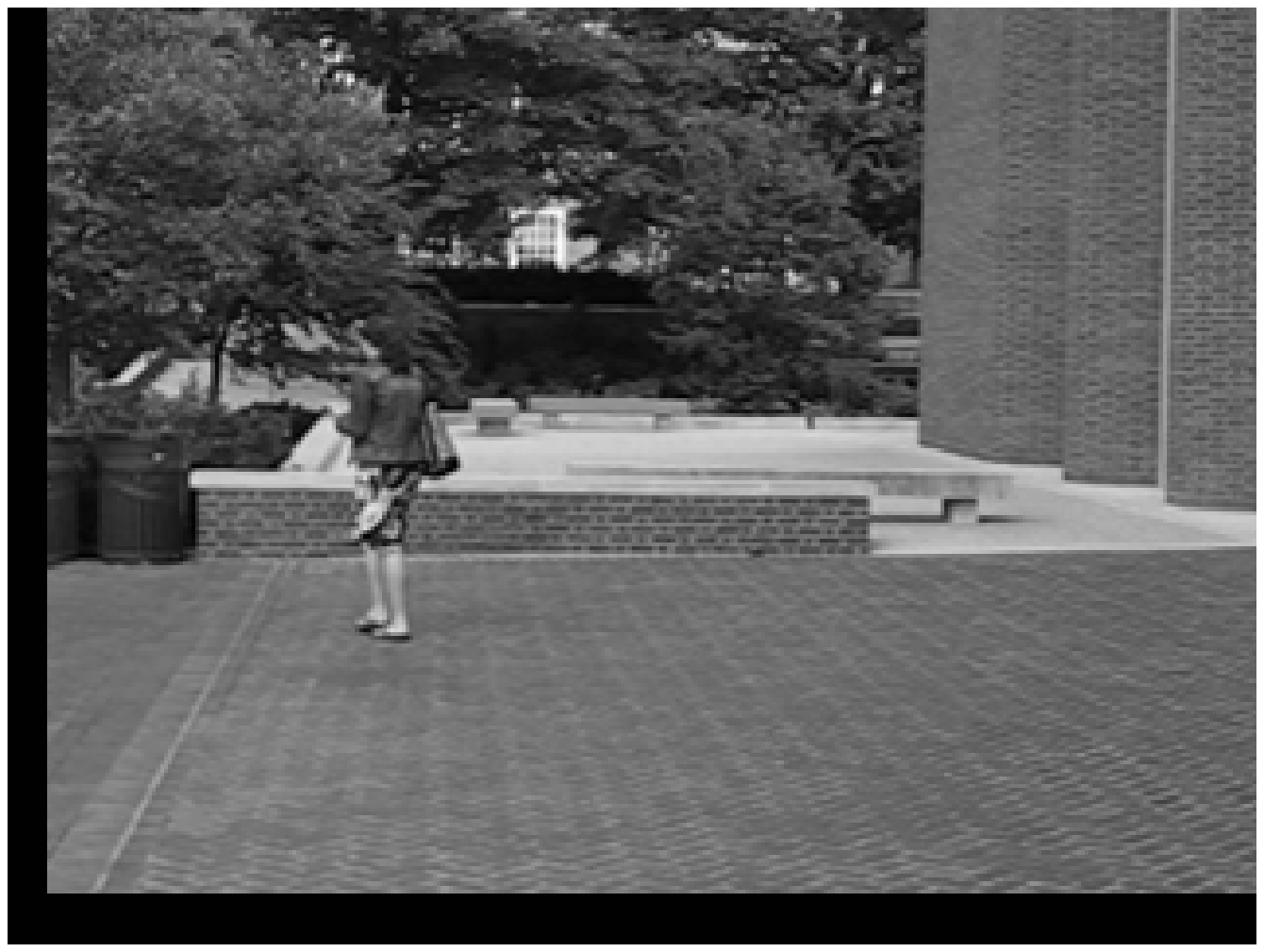}
 \includegraphics[width=0.15\linewidth]{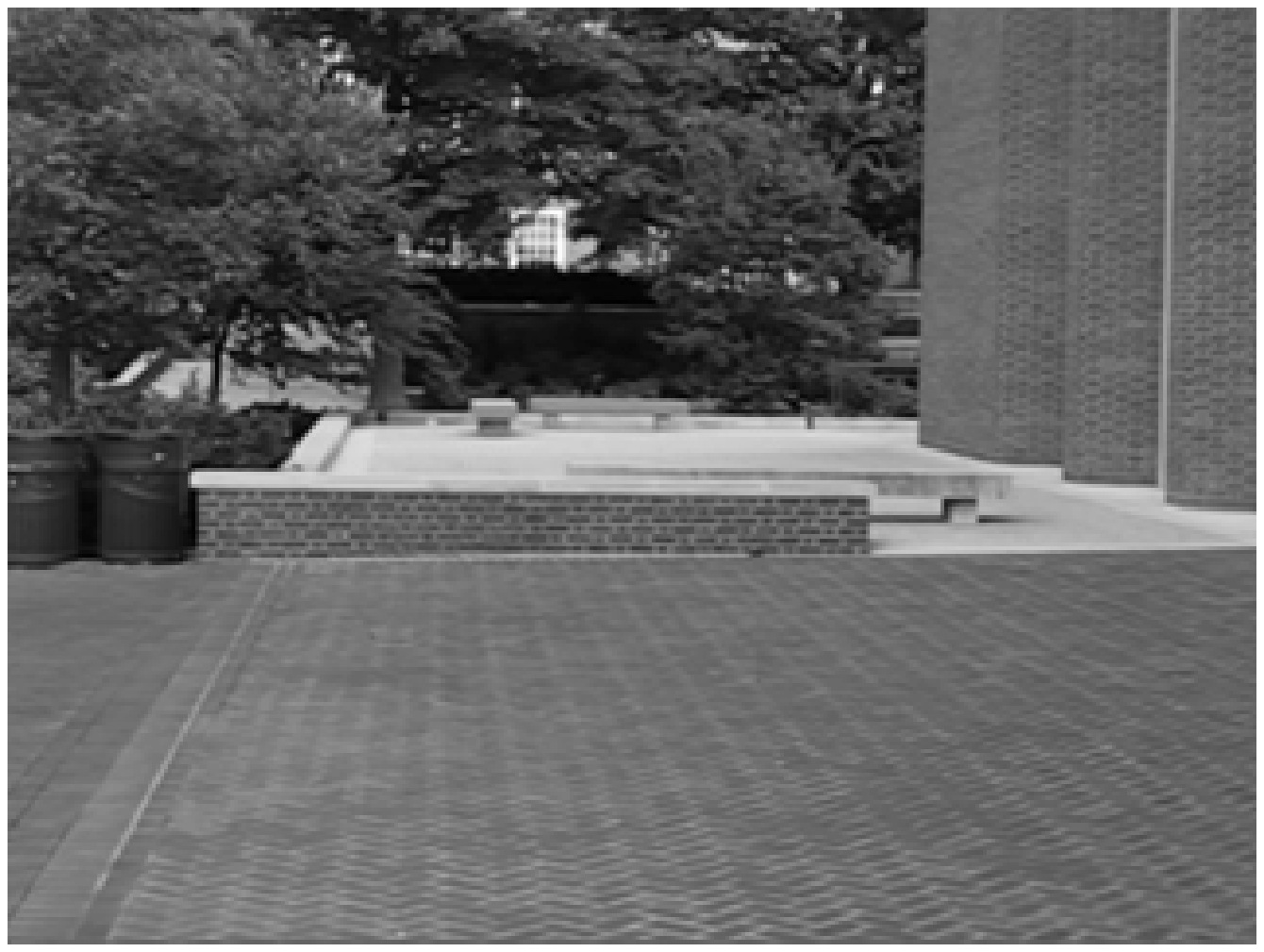}
 \includegraphics[width=0.15\linewidth]{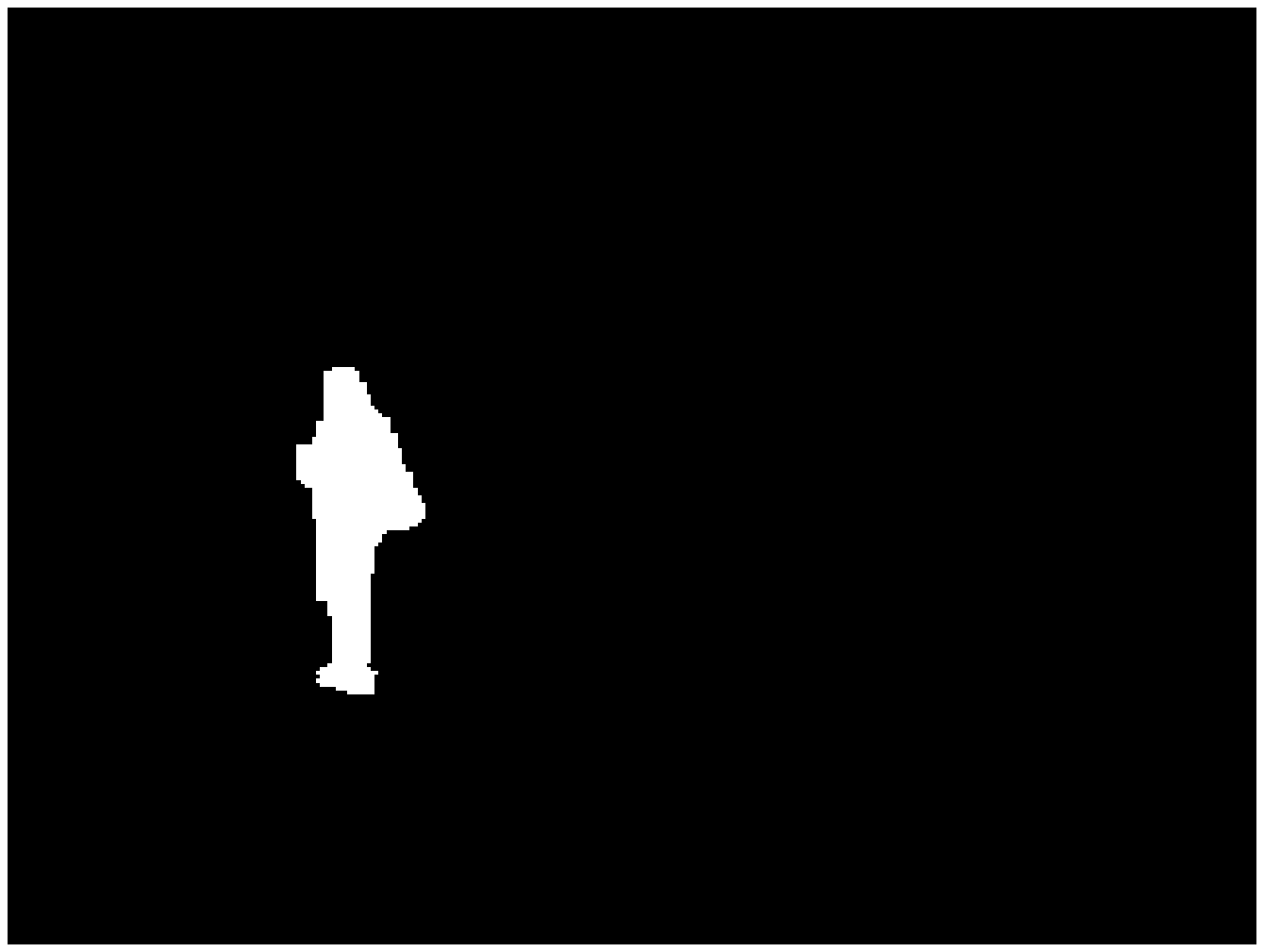}
 \includegraphics[width=0.15\linewidth]{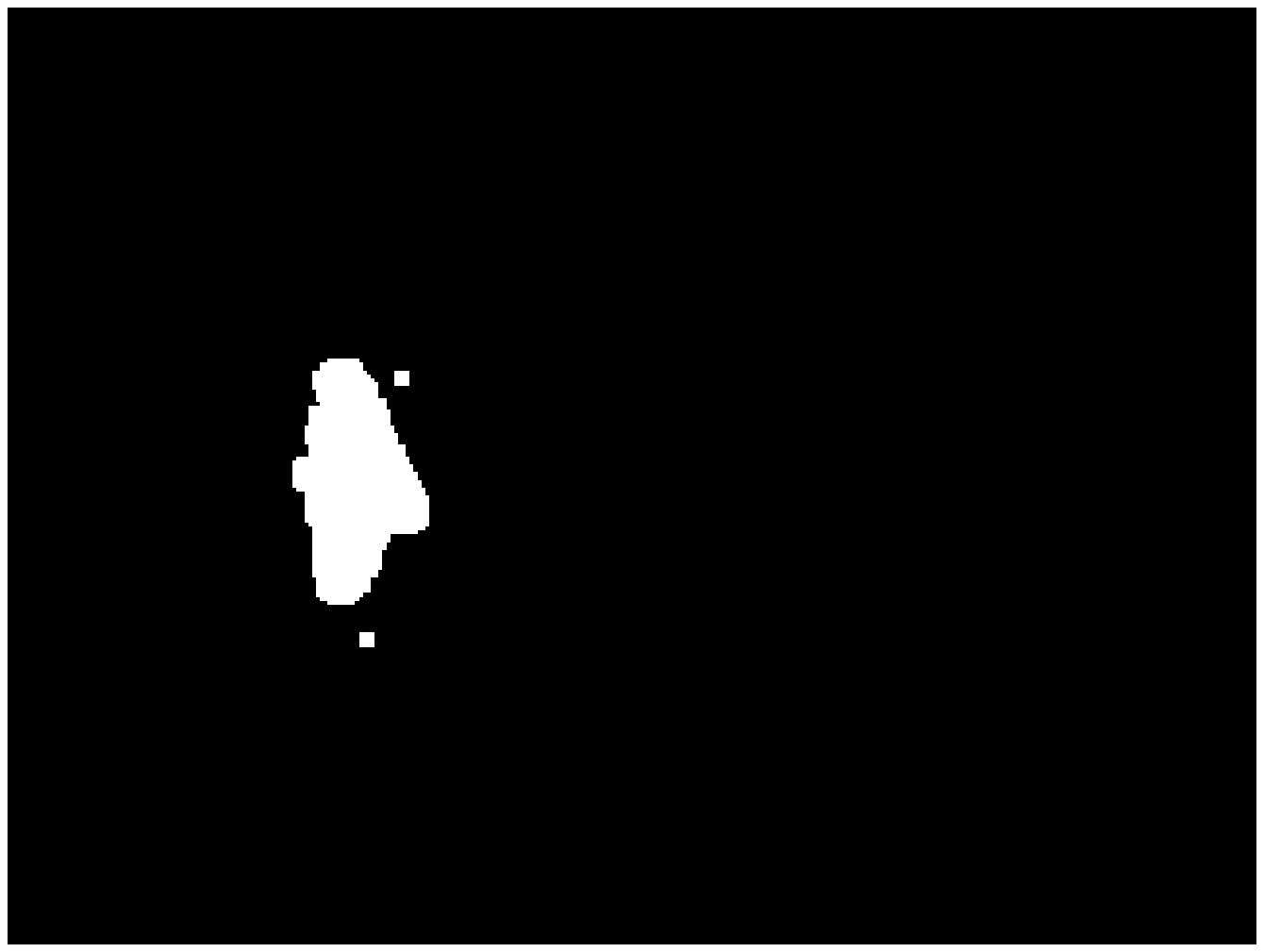}
 \includegraphics[width=0.15\linewidth]{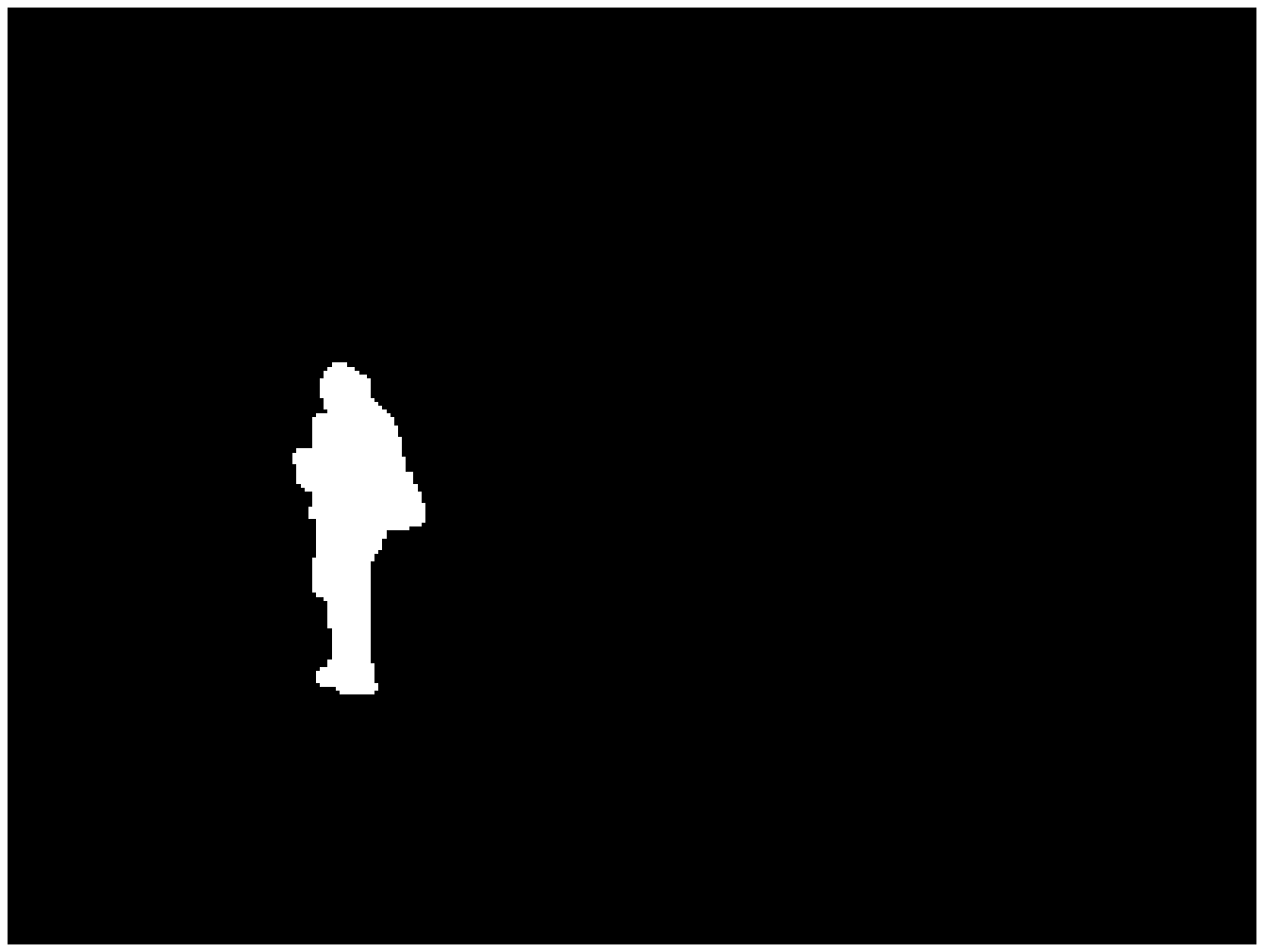}
\centerline{(a)} 
 \includegraphics[width=0.15\linewidth]{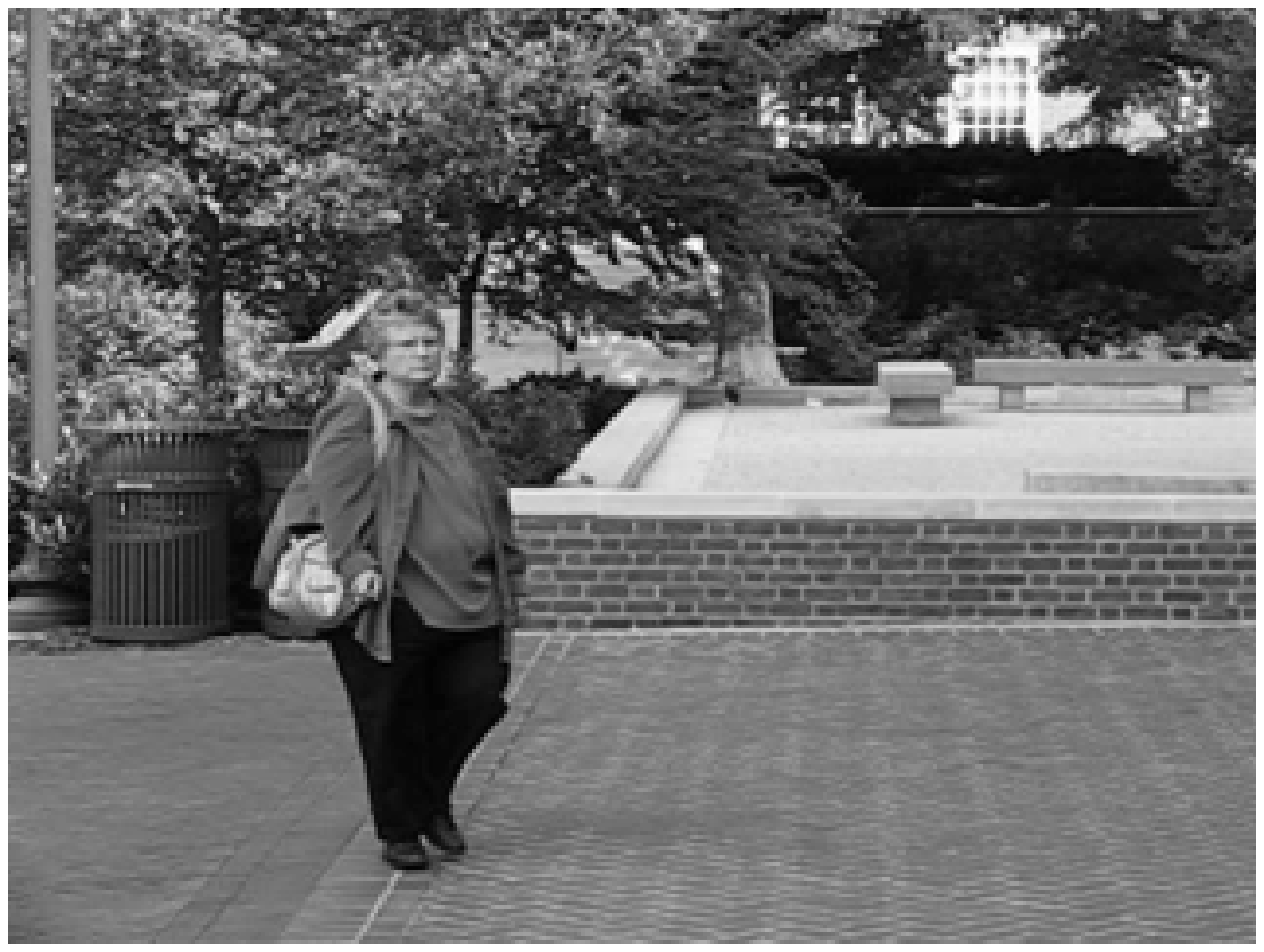}
 \includegraphics[width=0.15\linewidth]{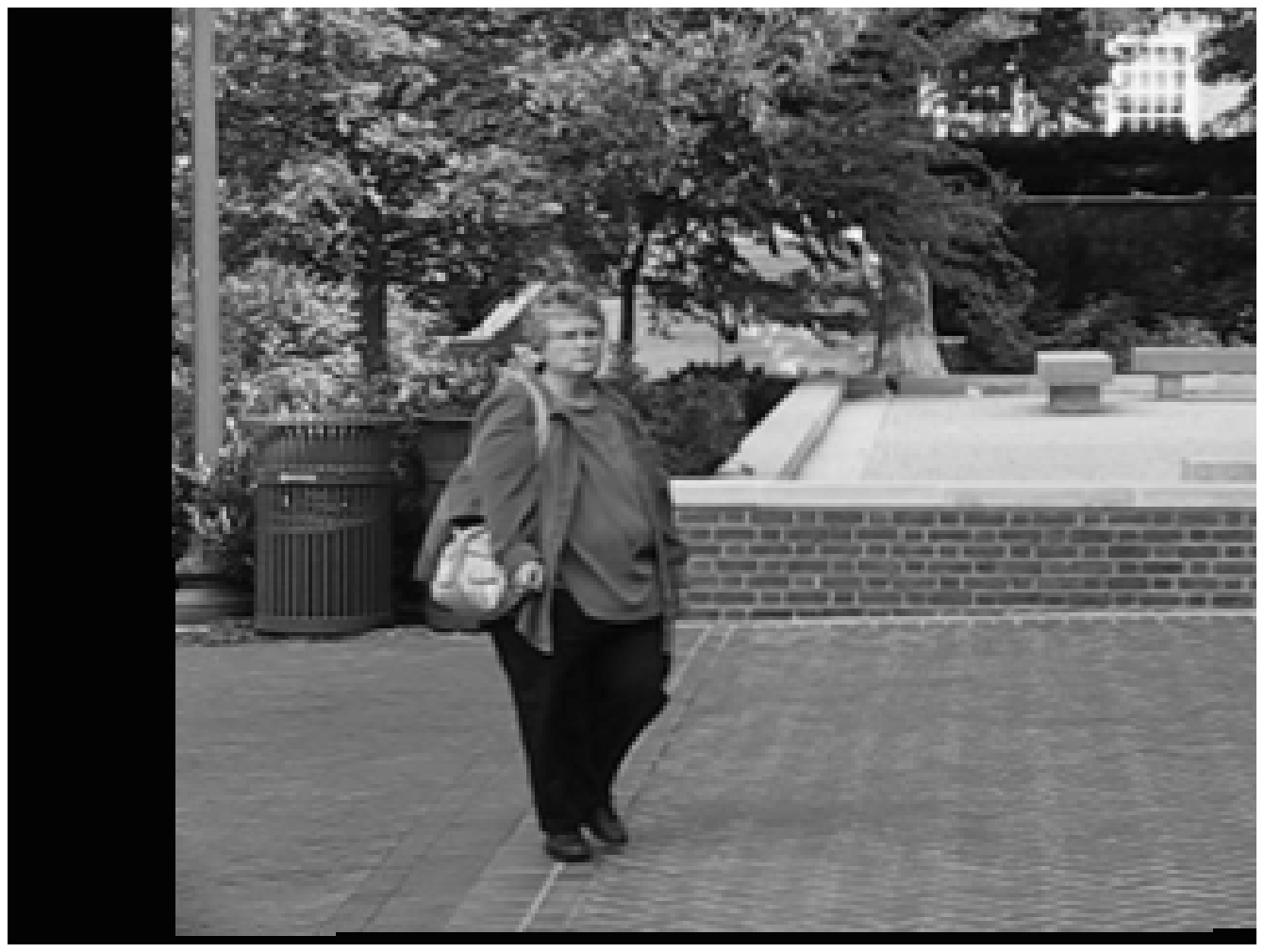}
 \includegraphics[width=0.15\linewidth]{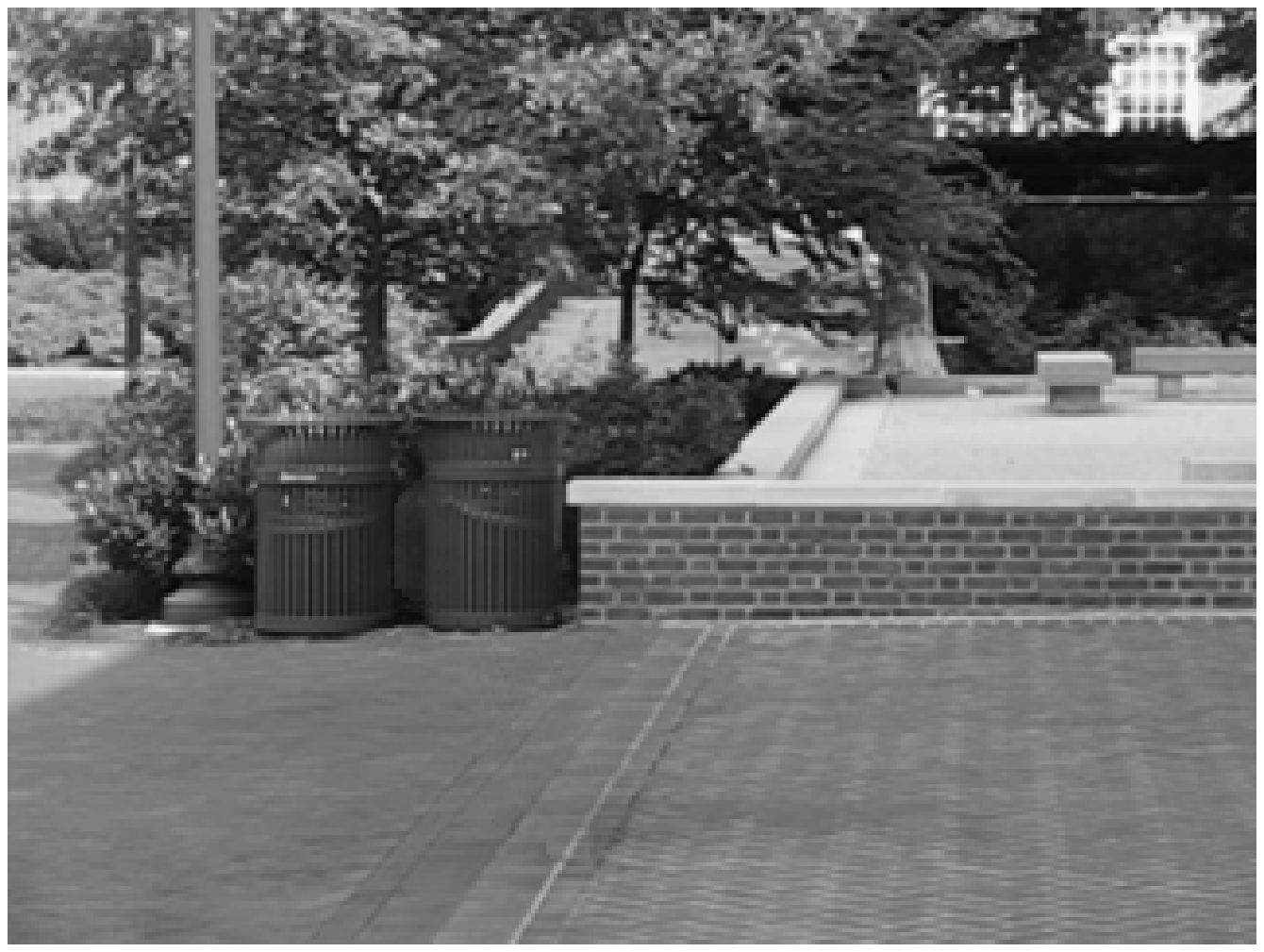}
 \includegraphics[width=0.15\linewidth]{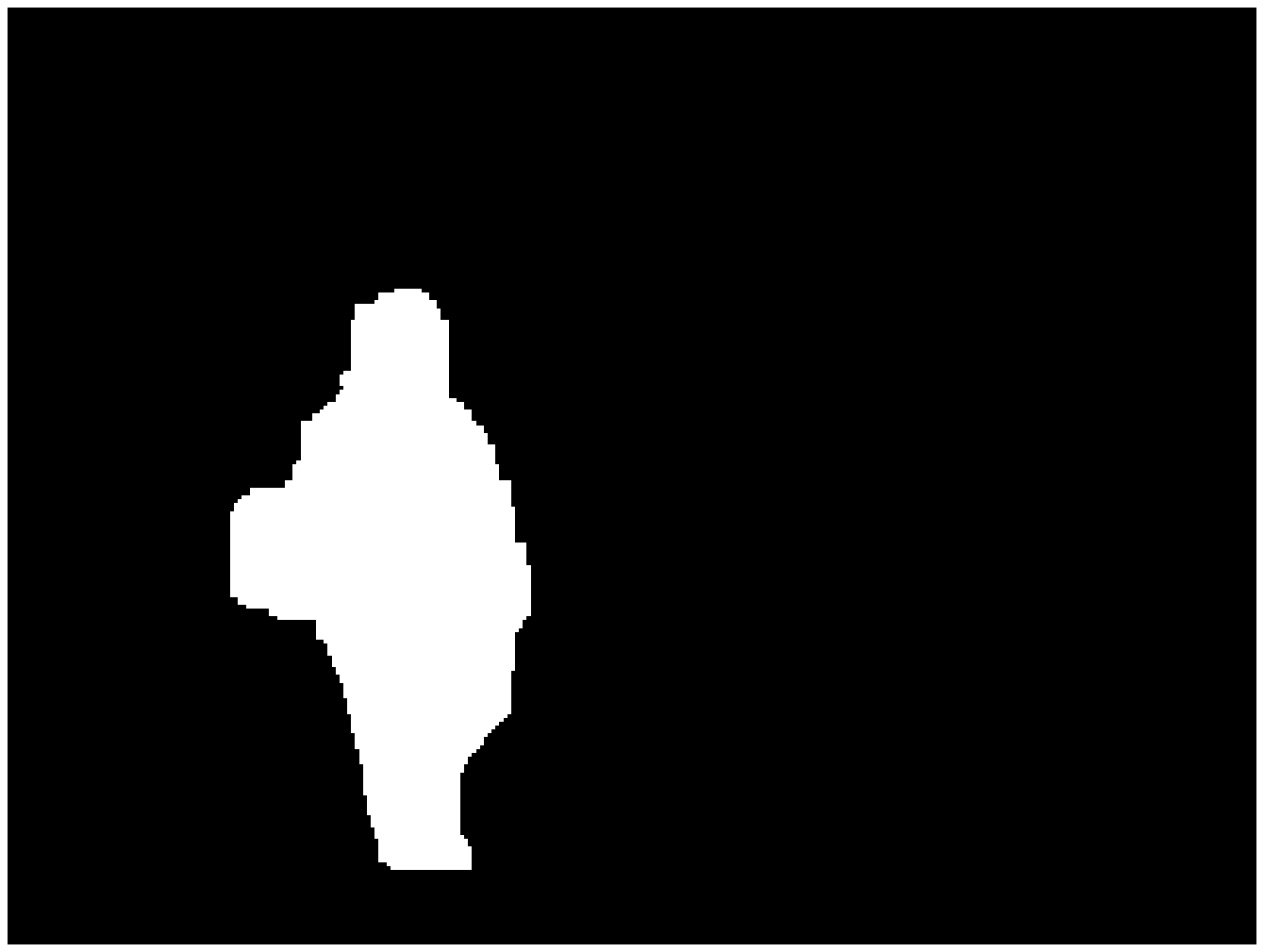}
 \includegraphics[width=0.15\linewidth]{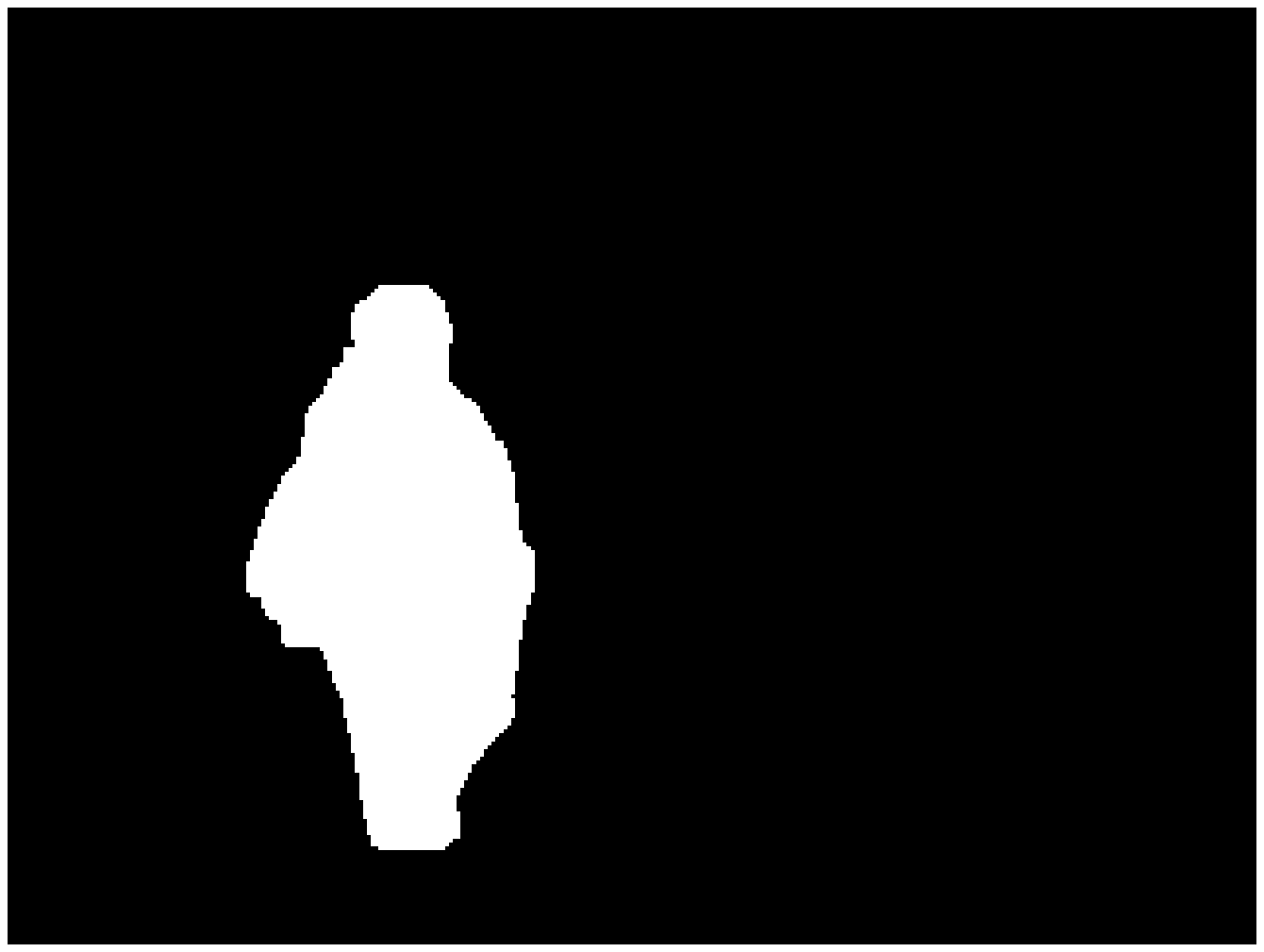}
 \includegraphics[width=0.15\linewidth]{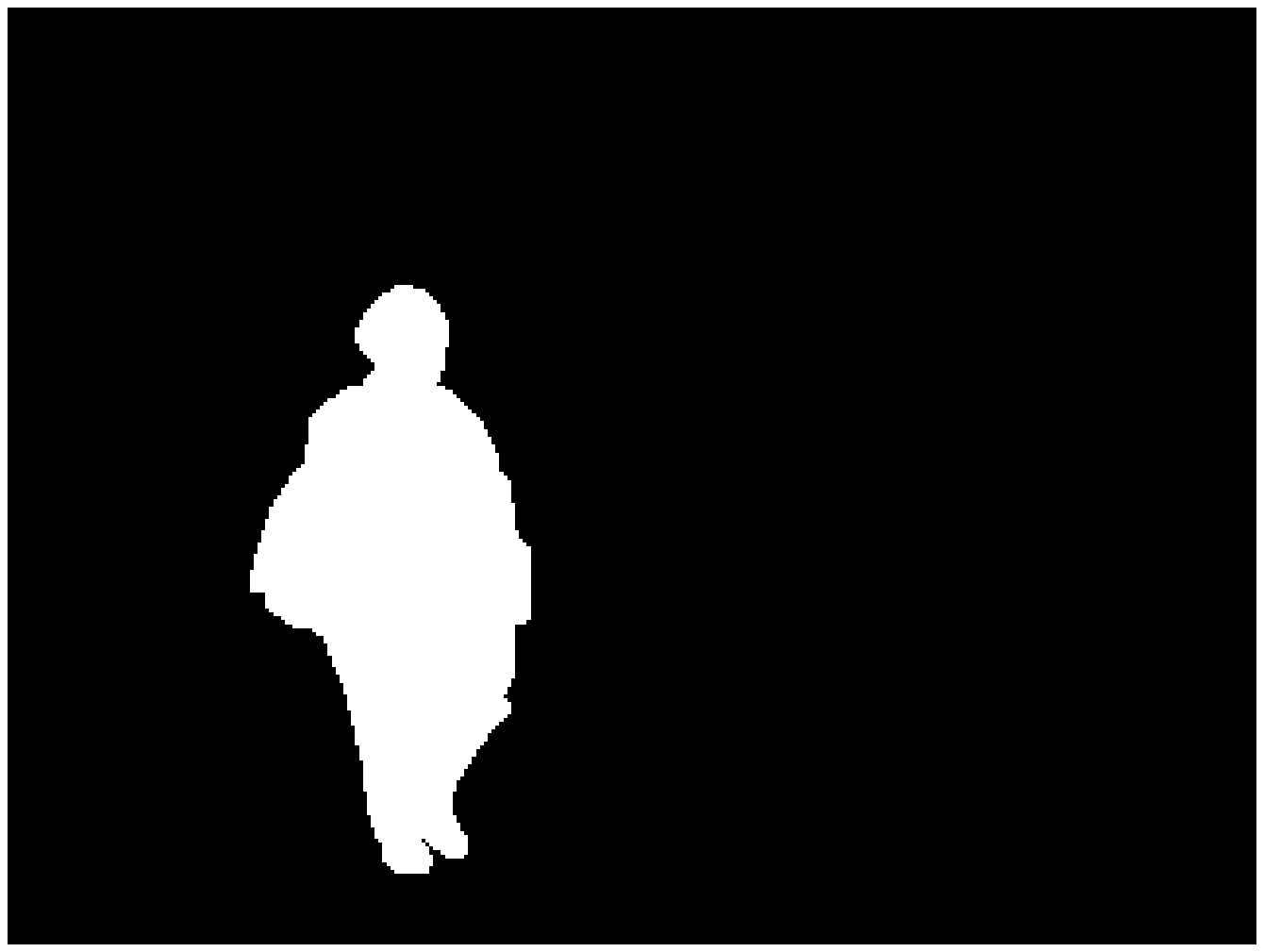}
\centerline{(b)} 
 \includegraphics[width=0.15\linewidth]{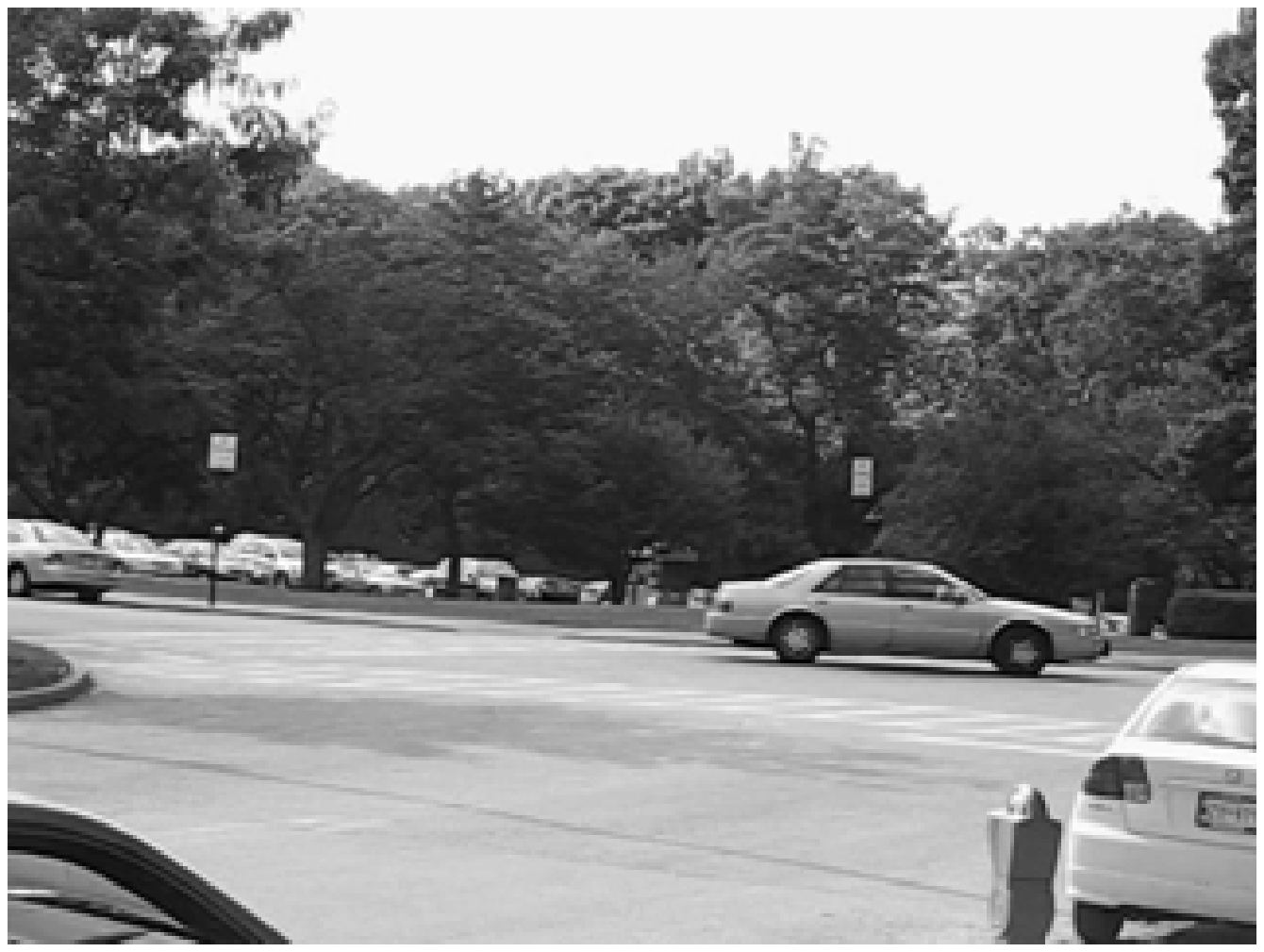}
 \includegraphics[width=0.15\linewidth]{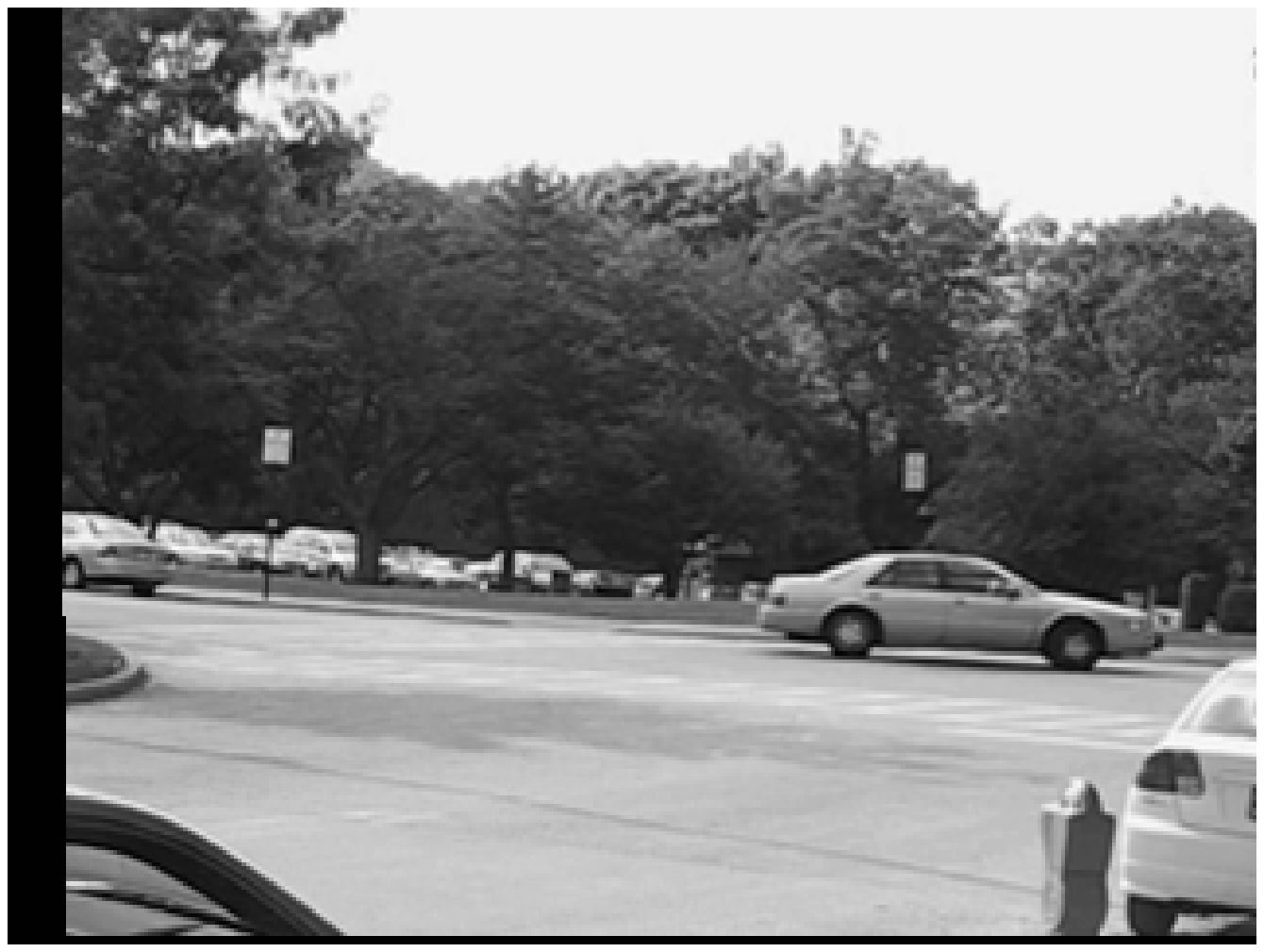}
 \includegraphics[width=0.15\linewidth]{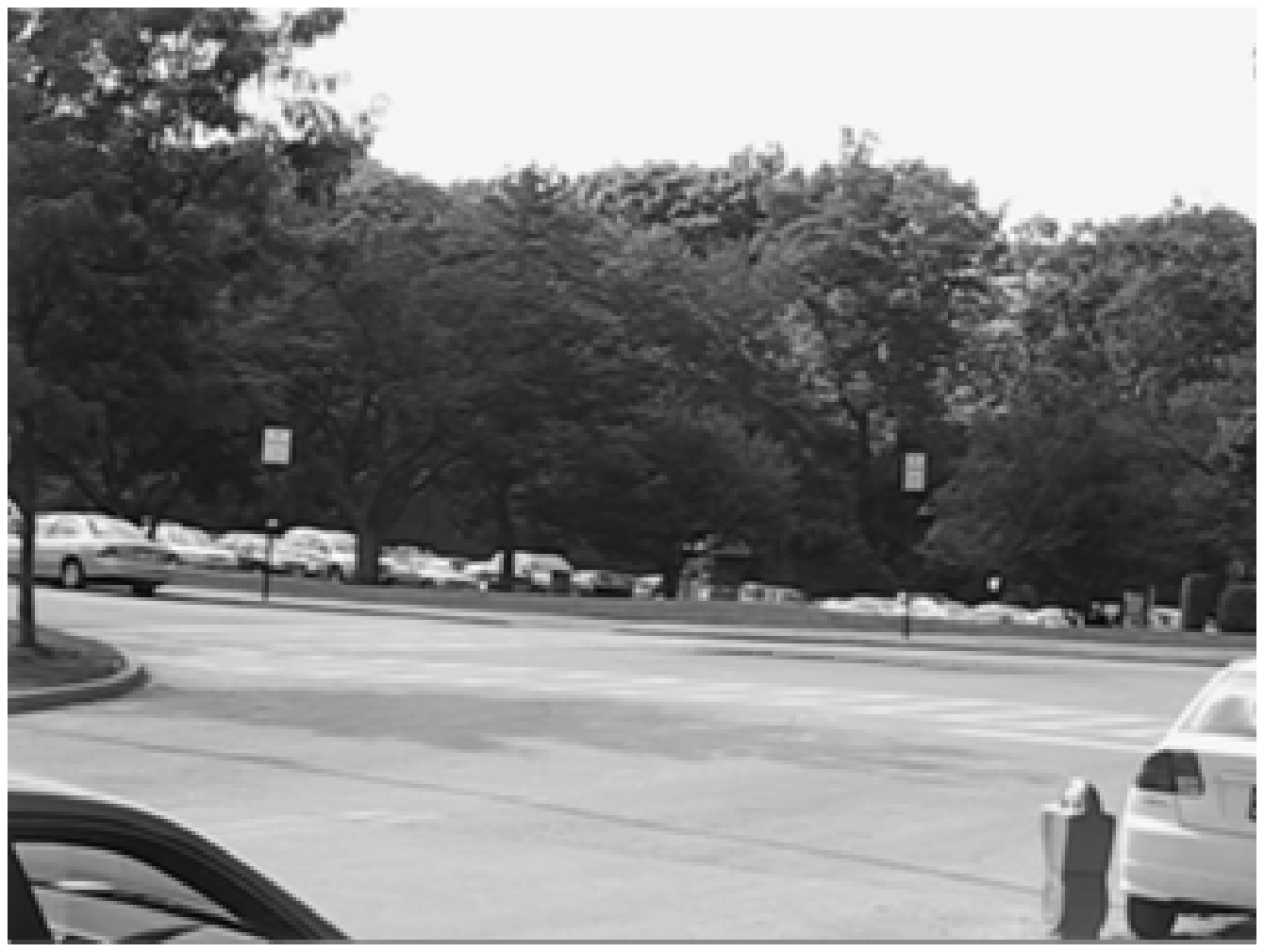}
 \includegraphics[width=0.15\linewidth]{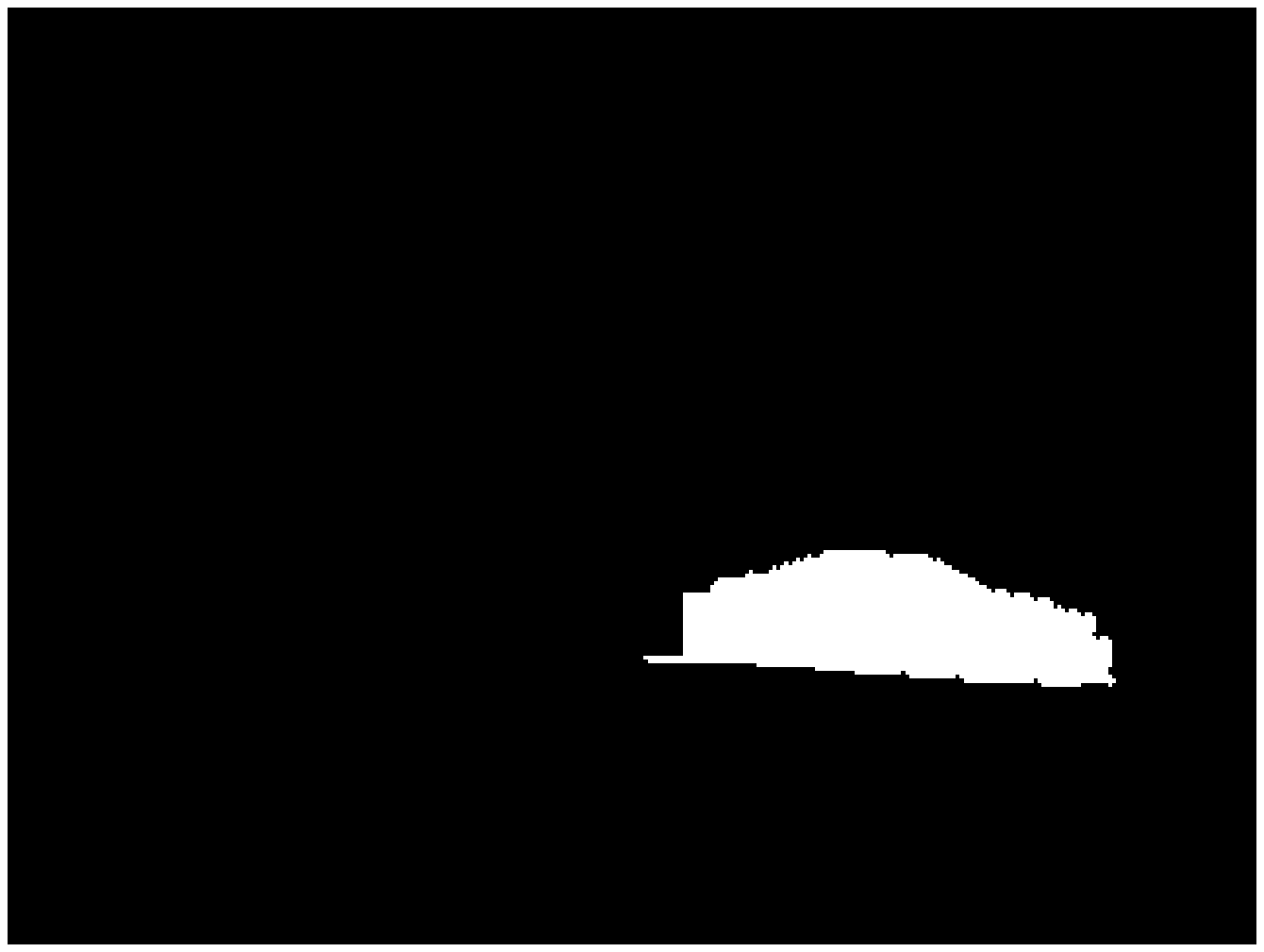}
 \includegraphics[width=0.15\linewidth]{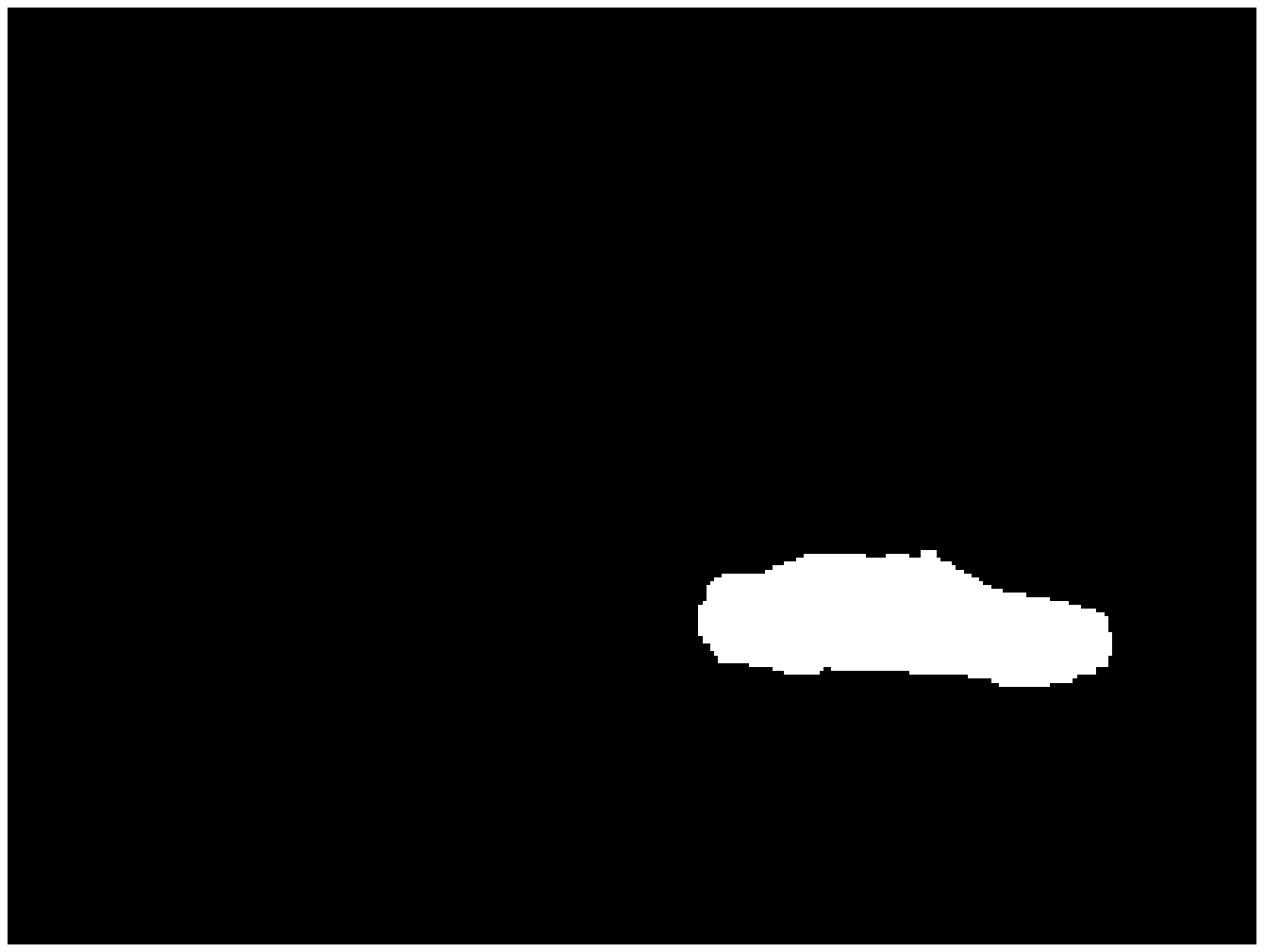}
 \includegraphics[width=0.15\linewidth]{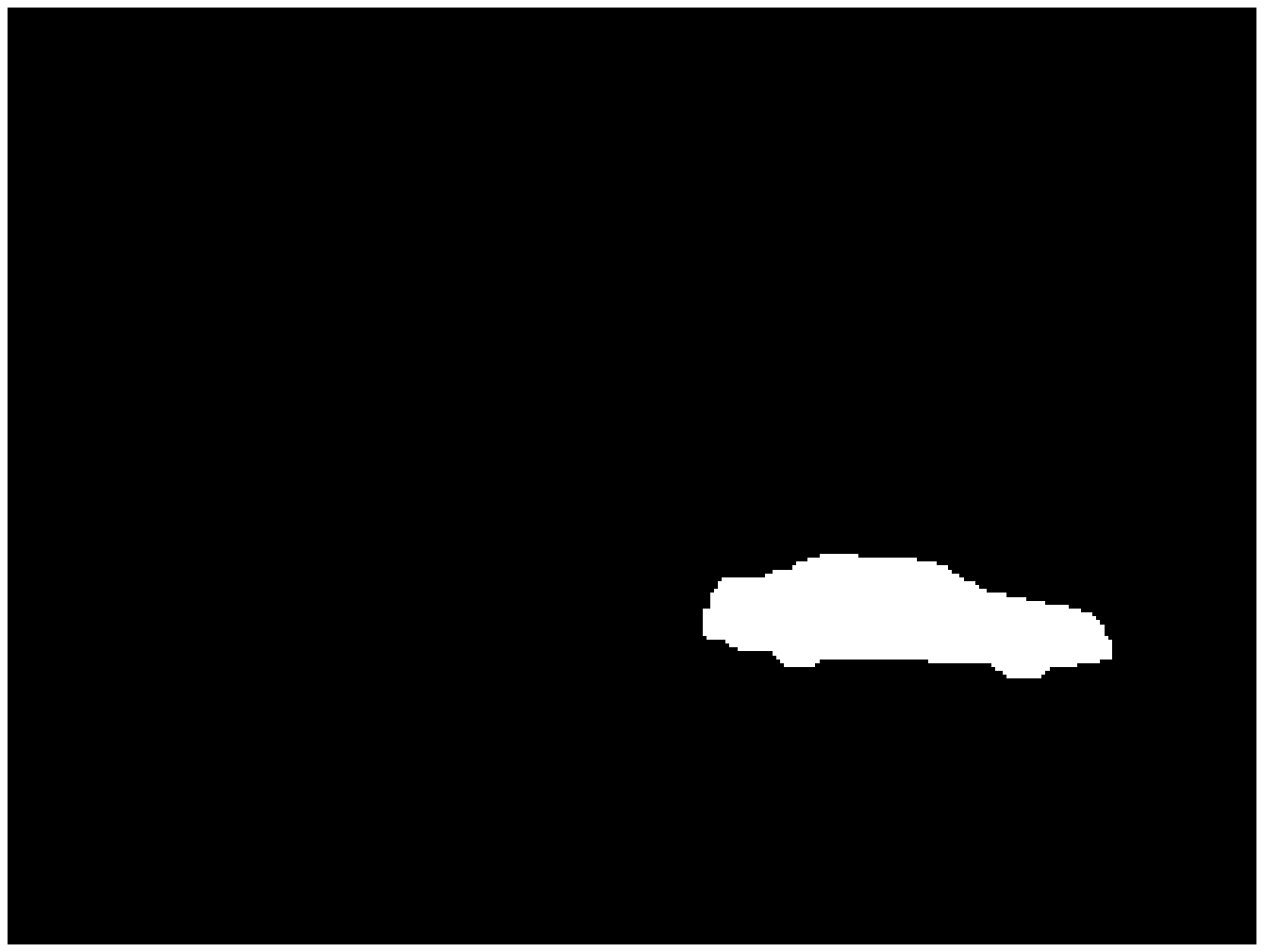}
\centerline{(c)} 
 \includegraphics[width=0.15\linewidth]{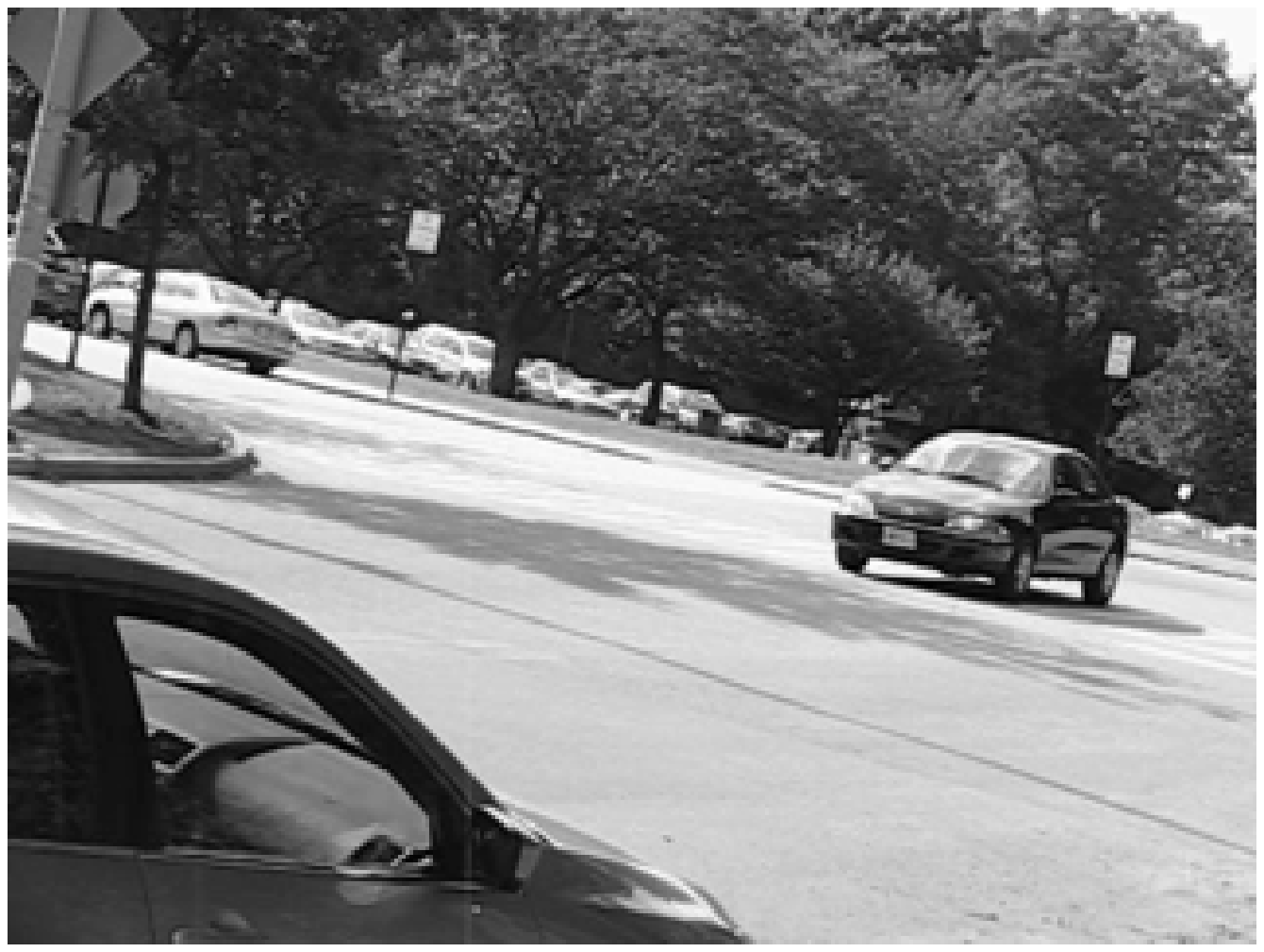}
 \includegraphics[width=0.15\linewidth]{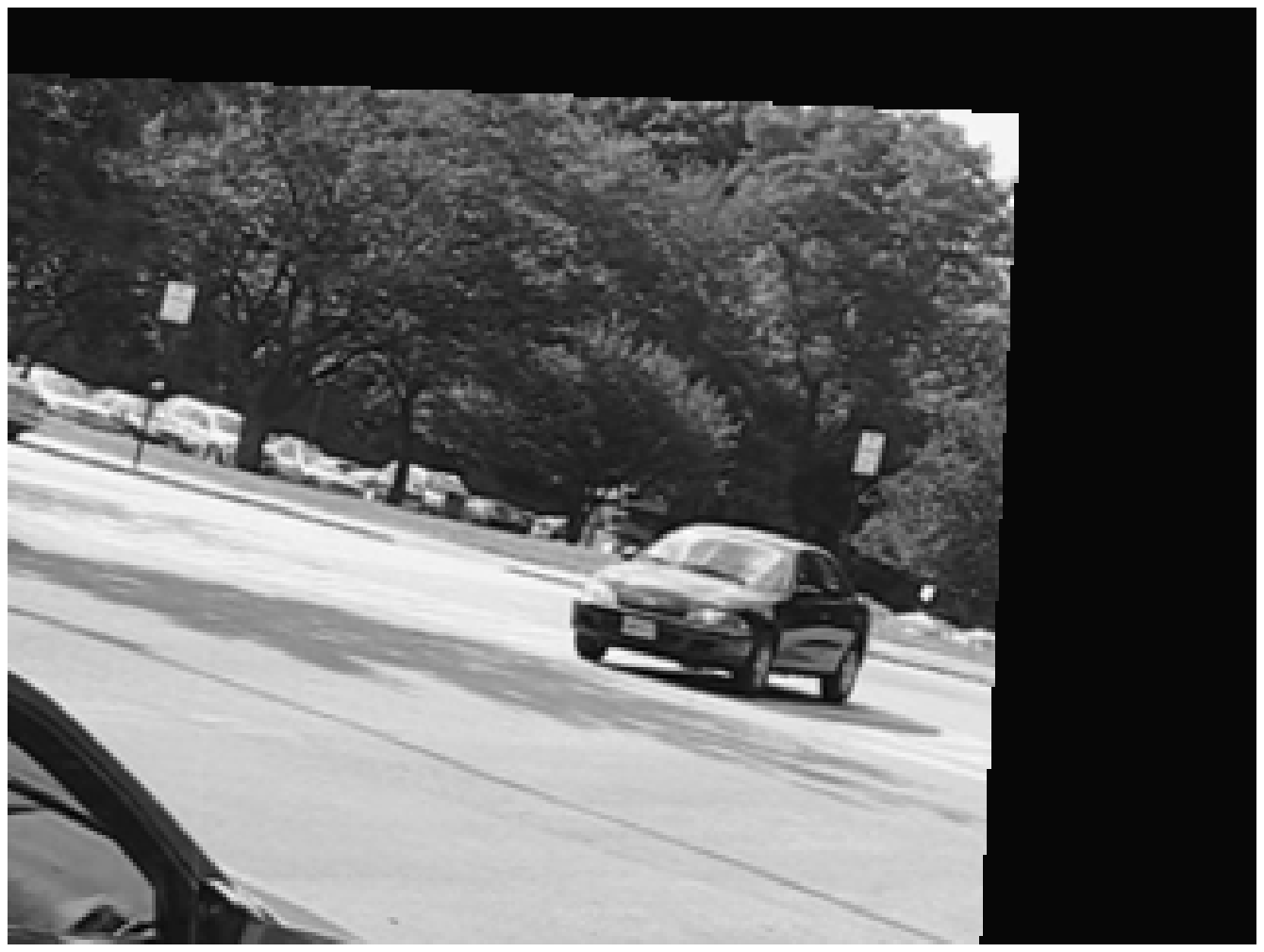}
 \includegraphics[width=0.15\linewidth]{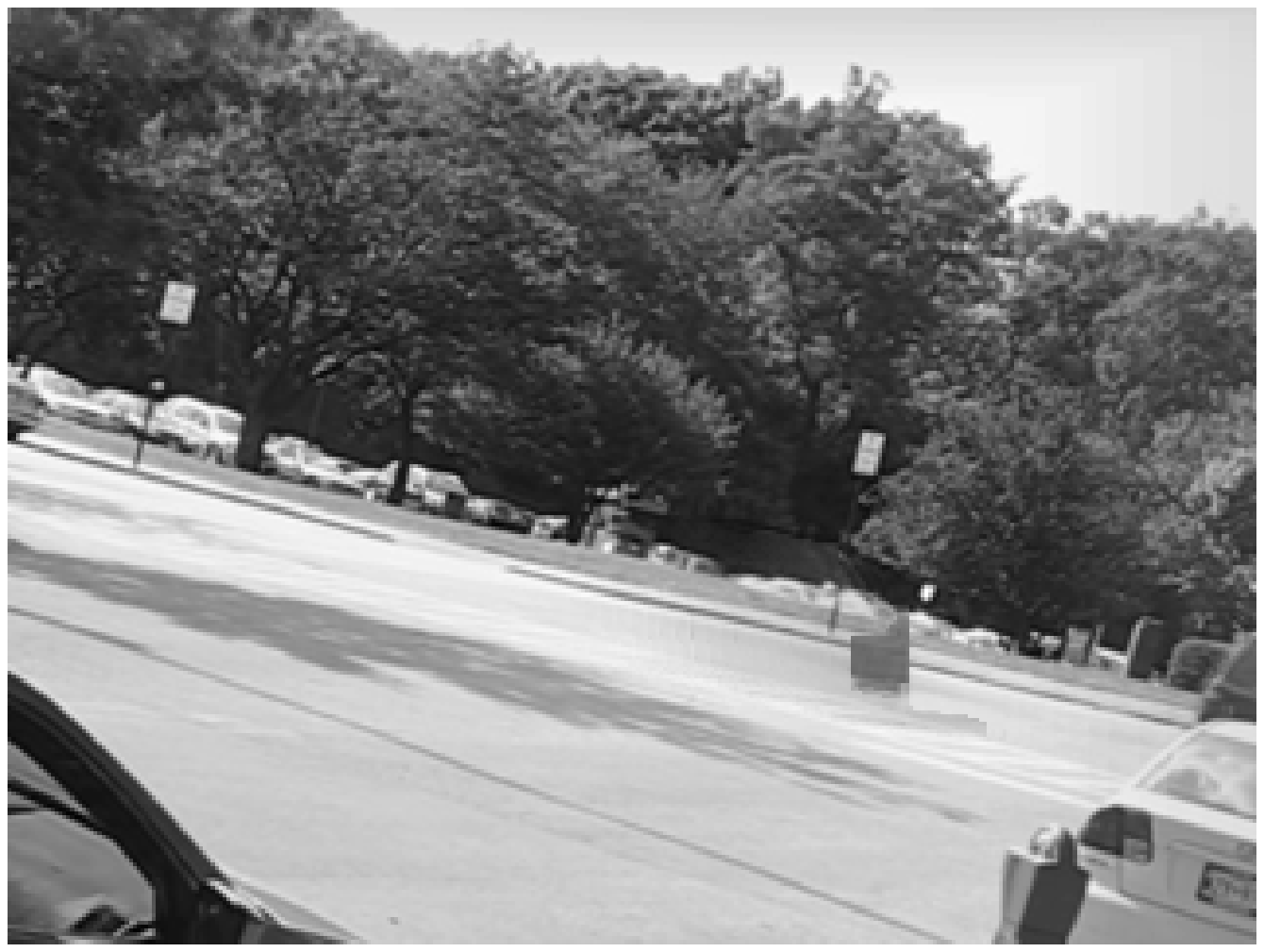}
 \includegraphics[width=0.15\linewidth]{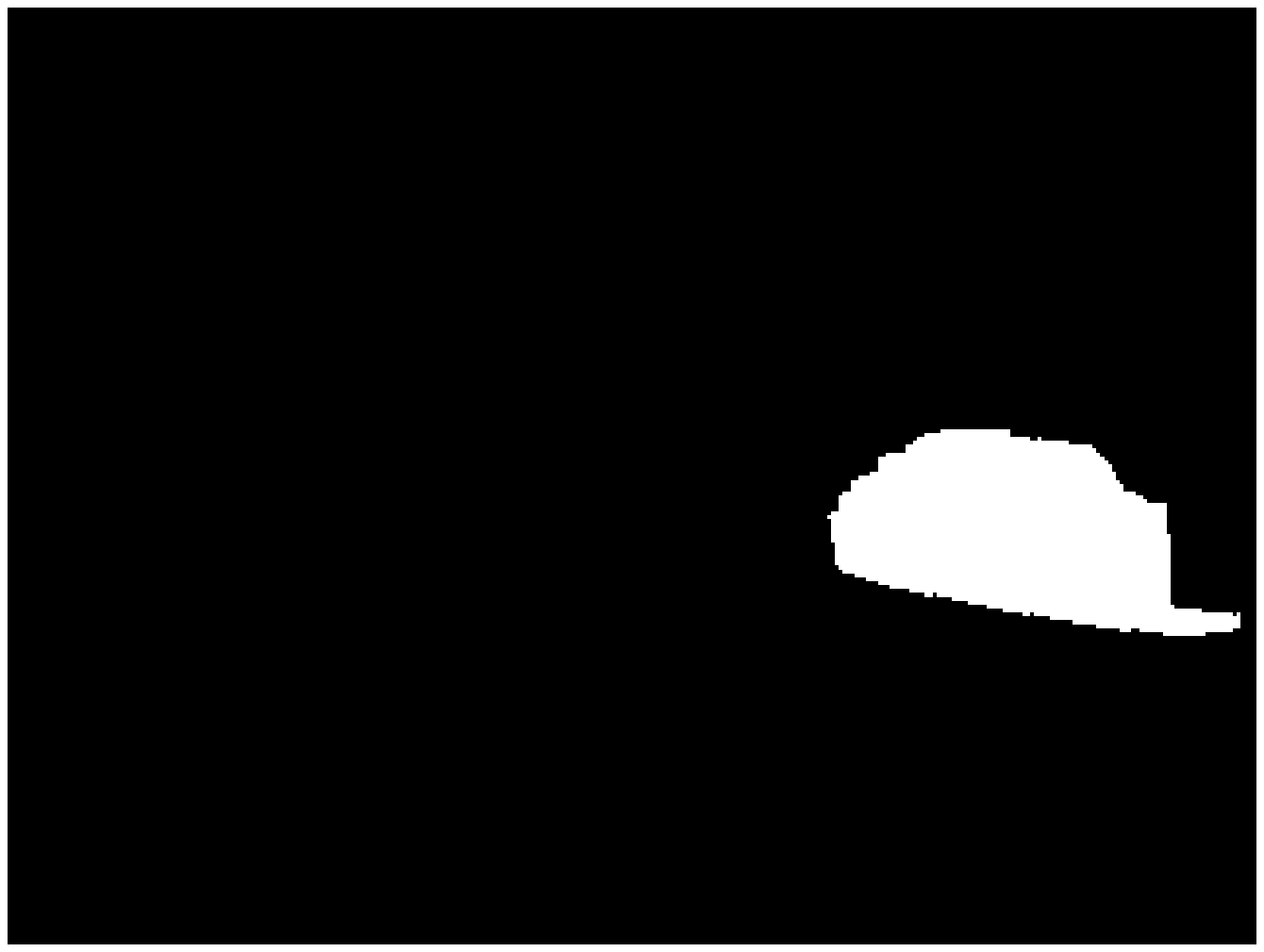}
 \includegraphics[width=0.15\linewidth]{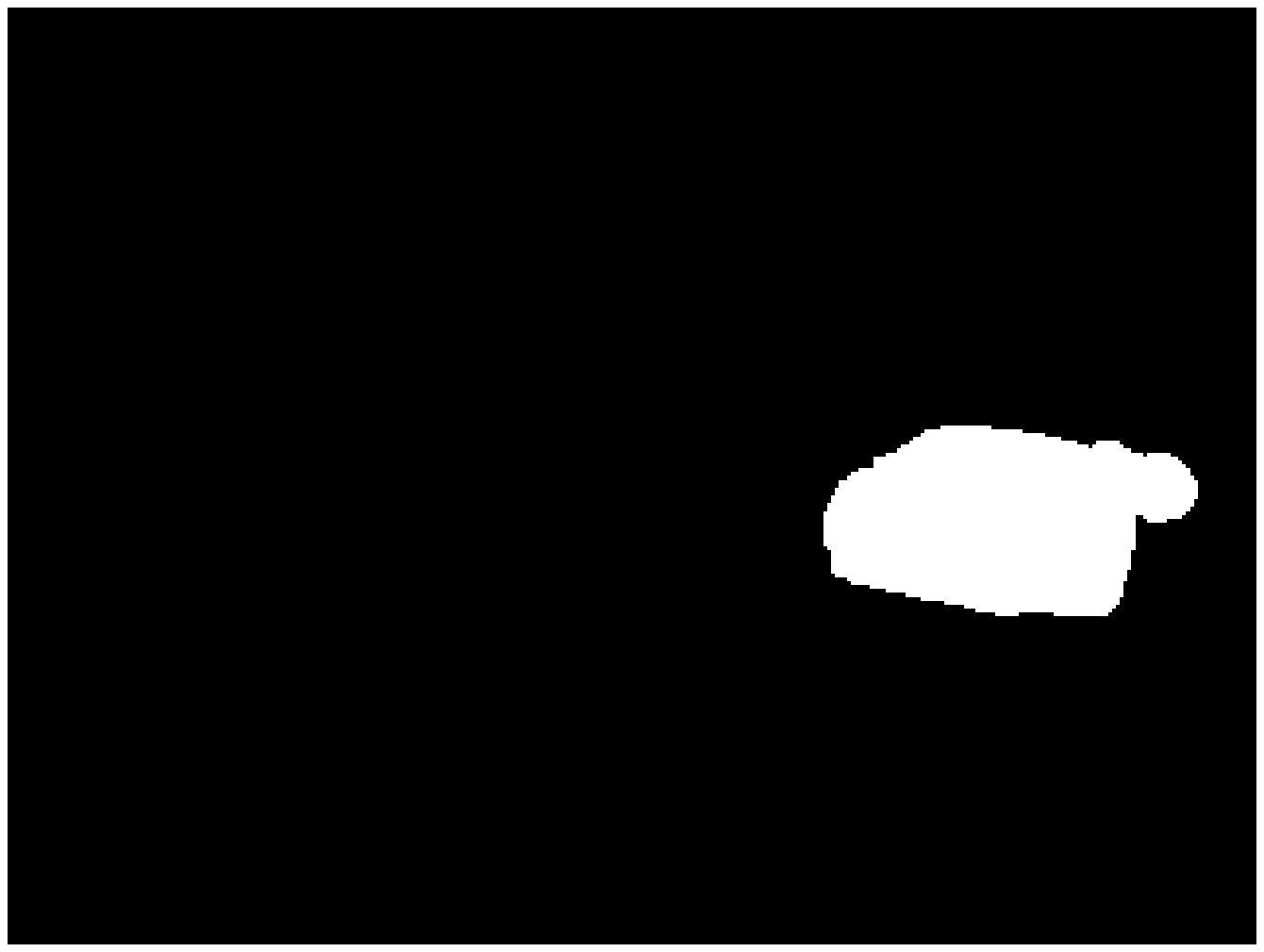}
 \includegraphics[width=0.15\linewidth]{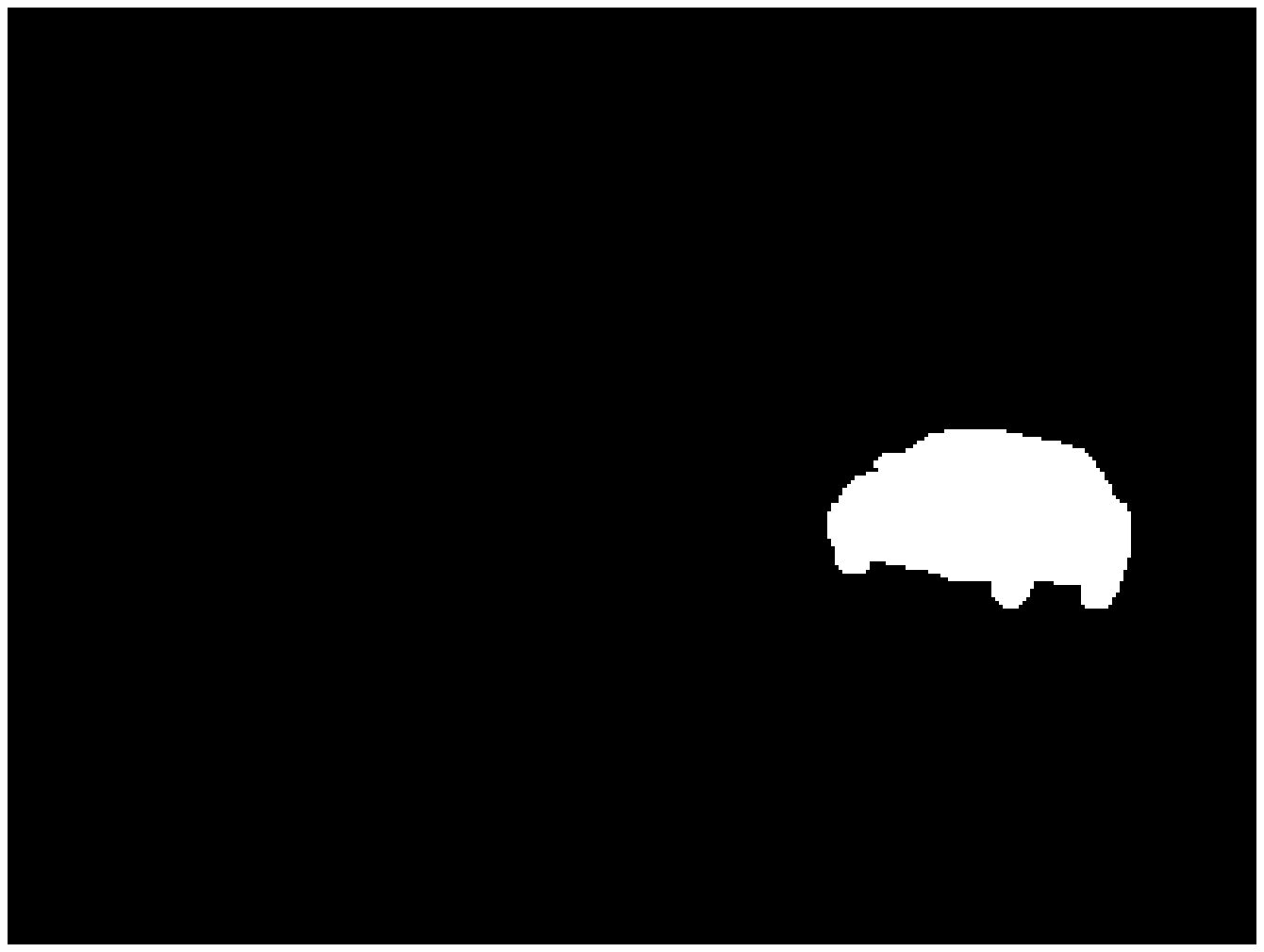}
\centerline{(d)} 
\caption{\small Four sequences captured by moving cameras. Sequence information is given in Table \ref{Tab_data}. Only the last frame of each sequence and the corresponding results are shown. From Column 2-4 present the results of DECOLOR, \ie the transformed image, the estimated background and the foreground mask. Column 5 shows the results given by the Brox and Malik's algorithm for motion segmentation \cite{brox2010object}. The last column shows the ground truth.}\label{Fig_MotionSeg}
\end{figure*}

Next, we demonstrate the potential of DECOLOR applied to motion segmentation problems using the Berkeley motion segmentation dataset\footnote{\url{http://lmb.informatik.uni-freiburg.de/resources/datasets/moseg.en.html}}. We use two \emph{people} sequences and twelve \emph{car} sequences, which are specialized for short-term analysis. Each sequence has several annotated frames as the ground truth for segmentation. Fig. \ref{Fig_MotionSeg} shows several examples and the results of DECOLOR. The transformed images $D\circ{\hat\tau}$ are shown in Column 2. Notice the extrapolated regions shown in black near the borders of these images. To minimize the influence of this numerical error, we constrain these pixels to be background when estimating $S$, but consider them as missing entries when estimating $B$. Fig. \ref{Fig_MotionSeg} demonstrates that DECOLOR can align the images, learn a background model and detect objects correctly.

For comparison, we also test the motion segmentation algorithm recently developed by Brox and Malik \cite{brox2010object}. The Brox-Malik algorithm analyzes the point trajectories along the sequence and segment them into clusters. To obtain pixel-level segmentation, the variational method \cite{ochs2011object} can be applied to turn the trajectory clusters into dense regions. This additional step makes use of the color and edge information in images \cite{ochs2011object}, while DECOLOR only uses the motion cue and directly generates the segmentation.

Quantitatively, we calculate the precision and recall of foreground detection, as shown in Table \ref{Tab_moseg}. In summary, for most sequences with moderate camera motion, the performance of DECOLOR is competitive. On the \emph{people} sequences, DECOLOR performs better. The feet of the lady are not detected by the Brox-Malik algorithm. The reason is that the Brox-Malik algorithm relies on correct motion tracking and clustering \cite{ochs2011object}, which is difficult when the object is small and moving nonrigidly. Instead, DECOLOR avoids the complicated motion analysis. However, DECOLOR works poorly on the cases where the background is a 3D scene with a large depth and the camera moves a lot, \eg the sequences named \emph{cars9} and \emph{cars10}. This is because the parametric motion model used in DECOLOR can only compensate for the planar background motion.

\begin{table}
\renewcommand{\arraystretch}{1.3}
\caption{Quantitative evaluation using the sequences from the Berkeley motion segmentation dataset \cite{brox2010object}. The overall result is the median value over all \emph{people} and \emph{car} sequences.}
\label{Tab_moseg}
\centering
\begin{tabular}{lcccc}
\toprule
&\multicolumn{2}{c}{DECOLOR}&\multicolumn{2}{c}{Brox-Malik \cite{brox2010object}} \\
\cline{2-5}
Sequence & Precision & Recall & Precision & Recall \\
\hline
Fig. \ref{Fig_MotionSeg}(a) & 93.6\% & 93.3\% & 89.0\% & 77.5\% \\
Fig. \ref{Fig_MotionSeg}(b) & 92.5\% & 96.5\% & 91.7\% & 89.2\% \\
Fig. \ref{Fig_MotionSeg}(c) & 83.7\% & 98.4\% & 82.4\% & 99.4\% \\
Fig. \ref{Fig_MotionSeg}(d) & 72.0\% & 98.0\% & 76.4\% & 99.8\% \\
Overall & 81.8\% & 90.8\% & 80.8\% & 99.2\% \\
\bottomrule
\end{tabular}
\end{table}

\subsubsection{Dynamic foreground}

\begin{figure}
\centering
\begin{minipage}{0.2\linewidth}
 \centerline{\includegraphics[width=1\linewidth]{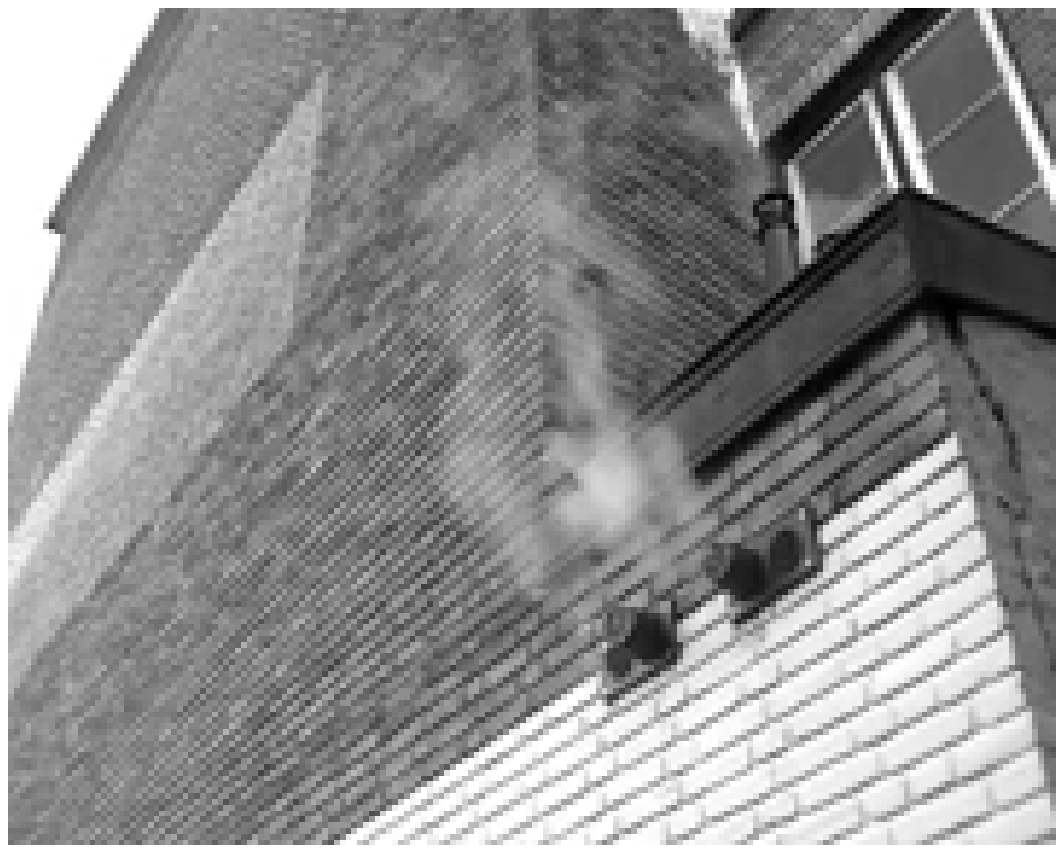}}
 \centerline{(a)}
\end{minipage}
\begin{minipage}{0.2\linewidth}
 \centerline{\includegraphics[width=1\linewidth]{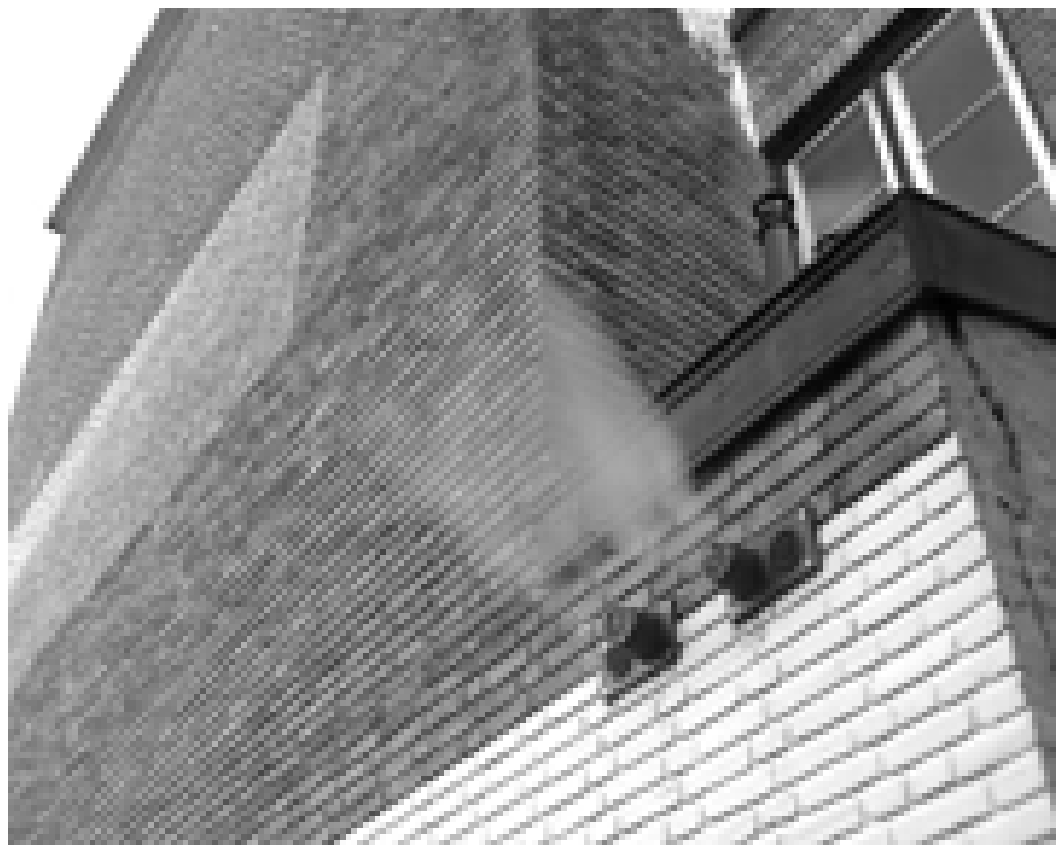}}
 \centerline{(b)}
\end{minipage}
\begin{minipage}{0.2\linewidth}
 \centerline{\includegraphics[width=1\linewidth]{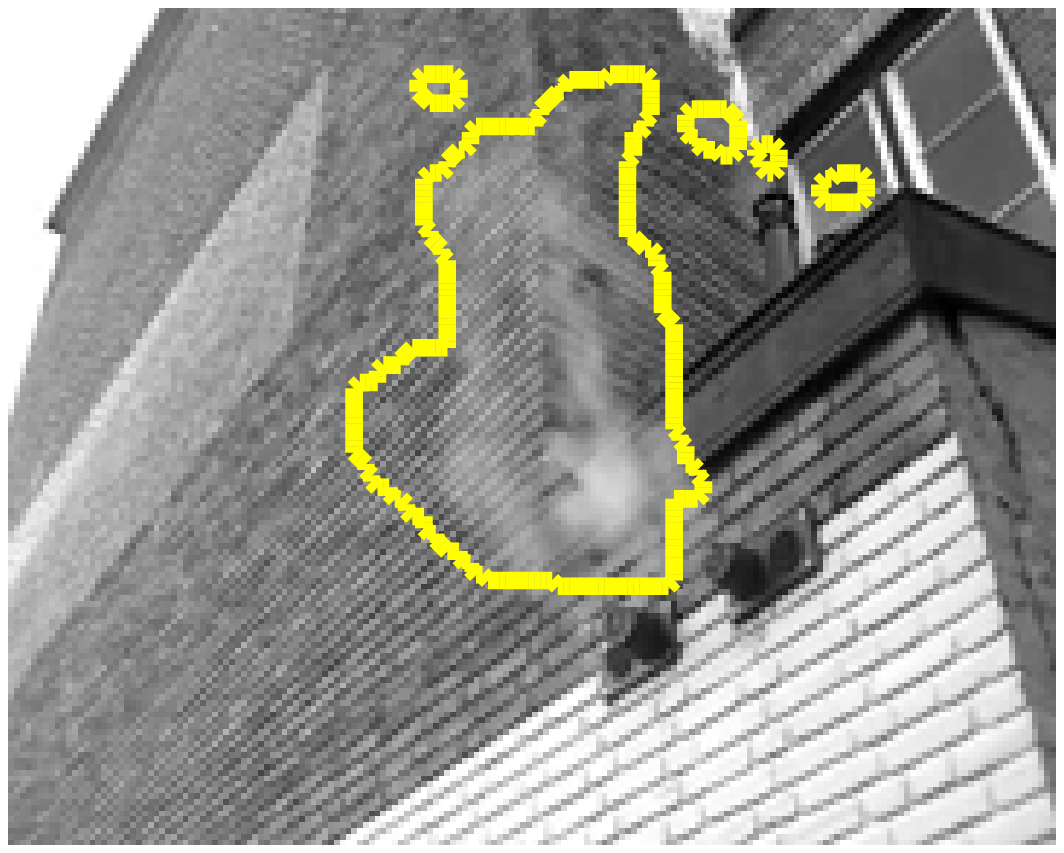}}
 \centerline{(c)}
\end{minipage}
\caption{An example of smoke detection. \textbf{(a)} Sample frame. \textbf{(b)} Estimated background. \textbf{(c)} Segmentation.
}\label{Fig_smoke}
\end{figure}

Dynamic texture segmentation has drawn some attentions in recent computer vision research \cite{fazekas2009dynamic,chan2009layered}. While we have shown that DECOLOR can model periodically varying textures like escalators or water surfaces as background, it is also able to detect fast changing textures, whose motion has little periodicity and can not be modeled as low-rank. Fig. \ref{Fig_smoke} shows such an example, where the smoke is detected as foreground. Here, the background behind smoke can not be recovered since it is always occluded.

\subsubsection{Computational cost}

Our algorithm is implemented in MATLAB. All experiments are run on a desktop PC with a 3.4 GHz Intel i7 CPU and 3 GB RAM. Since the graph cut is operated for each frame separately as discussed in Section \ref{section_graphcuts}, the dominant cost comes from the computation of SVD in each iteration. The cpu time of DECOLOR for sequences in Fig. \ref{Fig_BackSub} are  26.2, 13.3, 14.1, 11.4 and 14.4 seconds, while those of PCP are 26.8, 38.0, 15.7, 39.1, and 21.9 seconds, respectively. All results are obtained with a convergence precision of $10^{-4}$. The memory cost of DECOLOR and PCP are almost the same, since both of them need to compute SVD. The peak values of memory used in DECOLOR for sequences in Fig. \ref{Fig_BackSub}(a) and Fig. \ref{Fig_MotionSeg}(b) are around 65 MB and 210 MB, respectively.

\section{Discussion}\label{section_dis}

In this paper, we propose a novel framework named DECOLOR to segment moving objects from image sequences. It avoids complicated motion computation by formulating the problem as outlier detection and makes use of the low-rank modeling to deal with complex background.

We established the link between DECOLOR and PCP. Compared with PCP, DECOLOR uses the non-convex penalty and MRFs for outlier detection, which is more greedy to detect outlier regions that are relatively dense and contiguous. Despite of its satisfactory performance in our experiments, DECOLOR also has some disadvantages. Since DECOLOR minimizes a non-convex energy via alternating optimization, it converges to a local optimum with results depending on initialization of $\hat S$, while PCP always minimizes its energy globally. In all our experiments, we simply start from $\hat S = \mathbf{0}$. Also, we have tested other random initialization of $\hat S$ and it generally converges to a satisfactory result. This is because the SOFT-IMPUTE step will output similar results for each randomly generated $\hat S$ as long as $\hat S$ is not too dense.

As illustrated in Section \ref{section_Separable}, DECOLOR may misclassify unmoved objects or large textureless regions as background, since they are prone to entering the low-rank model. To address these problems, incorporating additional models such as object appearance or shape prior to improve the power of DECOLOR can be further explored in future.

Currently, DECOLOR works in a batch mode. Thus, it is not suitable for real-time object detection. In future, we plan to develop the online version of DECOLOR that can work incrementally, \eg the low-rank model extracted from beginning frames may be updated online when new frames arrive.

\bibliographystyle{IEEETran}
\bibliography{reference}

\begin{thebibliography}{10}
\providecommand{\url}[1]{#1}
\csname url@samestyle\endcsname
\providecommand{\newblock}{\relax}
\providecommand{\bibinfo}[2]{#2}
\providecommand{\BIBentrySTDinterwordspacing}{\spaceskip=0pt\relax}
\providecommand{\BIBentryALTinterwordstretchfactor}{4}
\providecommand{\BIBentryALTinterwordspacing}{\spaceskip=\fontdimen2\font plus
\BIBentryALTinterwordstretchfactor\fontdimen3\font minus
  \fontdimen4\font\relax}
\providecommand{\BIBforeignlanguage}[2]{{%
\expandafter\ifx\csname l@#1\endcsname\relax
\typeout{** WARNING: IEEEtran.bst: No hyphenation pattern has been}%
\typeout{** loaded for the language `#1'. Using the pattern for}%
\typeout{** the default language instead.}%
\else
\language=\csname l@#1\endcsname
\fi
#2}}
\providecommand{\BIBdecl}{\relax}
\BIBdecl

\bibitem{yilmaz2006object}
A.~Yilmaz, O.~Javed, and M.~Shah, ``{Object tracking: A survey},'' \emph{ACM
  computing surveys}, vol.~38, no.~4, pp. 1--45, 2006.

\bibitem{moeslund2006survey}
T.~Moeslund, A.~Hilton, and V.~Kruger, ``A survey of advances in vision-based
  human motion capture and analysis,'' \emph{Comput. Vis. Image Und.}, vol.
  104, no. 2-3, pp. 90--126, 2006.

\bibitem{papageorgiou1998general}
C.~Papageorgiou, M.~Oren, and T.~Poggio, ``A general framework for object
  detection,'' in \emph{Proc. of IEEE Int. Conf. Comput. Vis.}, 1998, p. 555.

\bibitem{viola2005detecting}
P.~Viola, M.~Jones, and D.~Snow, ``Detecting pedestrians using patterns of
  motion and appearance,'' \emph{Int. J. Comput. Vis.}, vol.~63, no.~2, pp.
  153--161, 2005.

\bibitem{grabner2006line}
H.~Grabner and H.~Bischof, ``On-line boosting and vision,'' in \emph{Proc. of
  IEEE Int. Conf. Compt. Vis. Pattern Recogn.}, 2006, pp. 260--267.

\bibitem{babenko2011robust}
B.~Babenko, M.-H. Yang, and S.~Belongie, ``Robust object tracking with online
  multiple instance learning,'' \emph{IEEE Trans. Pattern Anal. Mach. Intell.},
  vol.~33, no.~8, pp. 1619 --1632, 2011.

\bibitem{piccardi2004background}
M.~Piccardi, ``Background subtraction techniques: a review,'' in \emph{IEEE
  Int. Conf. on Systems, Man and Cybernetics}, 2004.

\bibitem{toyama1999wallflower}
K.~Toyama, J.~Krumm, B.~Brumitt, and B.~Meyers, ``{Wallflower: Principles and
  practice of background maintenance},'' in \emph{Proc. of IEEE Int. Conf.
  Comput. Vis.}, 1999.

\bibitem{vidal2004unified}
R.~Vidal and Y.~Ma, ``A unified algebraic approach to 2-d and 3-d motion
  segmentation,'' in \emph{Proc. of Eur. Conf. Comput. Vis.}, 2004.

\bibitem{Cremers05}
D.~Cremers and S.~Soatto, ``Motion competition: A variational approach to
  piecewise parametric motion segmentation,'' \emph{Int. J. Comput. Vis.},
  vol.~62, no.~3, pp. 249--265, 2005.

\bibitem{gutchess2001background}
D.~Gutchess, M.~Trajkovics, E.~Cohen-Solal, D.~Lyons, and A.~Jain, ``A
  background model initialization algorithm for video surveillance,'' in
  \emph{Proc. of IEEE Int. Conf. Comput. Vis.}, 2001.

\bibitem{nair2004unsupervised}
V.~Nair and J.~Clark, ``An unsupervised, online learning framework for moving
  object detection,'' in \emph{Proc. of IEEE Int. Conf. Compt. Vis. Pattern
  Recogn.}, vol.~2, 2004, pp. II--317.

\bibitem{Candes09}
E.~Candes, X.~Li, Y.~Ma, and J.~Wright, ``{Robust Principal Component
  Analysis?}'' \emph{Arxiv preprint arXiv:0912.3599}, 2009.

\bibitem{li2009markov}
S.~Li, \emph{Markov random field modeling in image analysis}.\hskip 1em plus
  0.5em minus 0.4em\relax Springer-Verlag New York Inc, 2009.

\bibitem{black1996robust}
M.~Black and P.~Anandan, ``The robust estimation of multiple motions:
  Parametric and piecewise-smooth flow fields,'' \emph{Comput. Vis. Image
  Und.}, vol.~63, no.~1, pp. 75--104, 1996.

\bibitem{Amiaz06}
T.~Amiaz and N.~Kiryati, ``Piecewise-smooth dense optical flow via level
  sets,'' \emph{Int. J. Comput. Vis.}, vol.~68, no.~2, pp. 111--124, 2006.

\bibitem{Brox06}
T.~Brox, A.~Bruhn, and J.~Weickert, ``Variational motion segmentation with
  level sets,'' in \emph{Proc. of Eur. Conf. Comput. Vis.}, 2006.

\bibitem{chan2009layered}
A.~Chan and N.~Vasconcelos, ``{Layered dynamic textures},'' \emph{IEEE Trans.
  Pattern Anal. Mach. Intell.}, vol.~31, no.~10, pp. 1862--1879, 2009.

\bibitem{cremers2003dynamic}
G.~Doretto, D.~Cremers, P.~Favaro, and S.~Soatto, ``Dynamic texture
  segmentation,'' in \emph{Proc. of IEEE Int. Conf. Comput. Vis.}, 2003.

\bibitem{fazekas2009dynamic}
S.~Fazekas, T.~Amiaz, D.~Chetverikov, and N.~Kiryati, ``{Dynamic texture
  detection based on motion analysis},'' \emph{Int. J. Comput. Vis.}, vol.~82,
  no.~1, pp. 48--63, 2009.

\bibitem{beauchemin1995computation}
S.~Beauchemin and J.~Barron, ``The computation of optical flow,'' \emph{ACM
  Computing Surveys}, vol.~27, no.~3, pp. 433--466, 1995.

\bibitem{tron2007benchmark}
R.~Tron and R.~Vidal, ``{A benchmark for the comparison of 3-D motion
  segmentation algorithms},'' in \emph{Proc. of IEEE Int. Conf. Compt. Vis.
  Pattern Recogn.}, 2007.

\bibitem{sheikh2009camera}
Y.~Sheikh, O.~Javed, and T.~Kanade, ``{Background subtraction for freely moving
  cameras},'' in \emph{Proc. of IEEE Int. Conf. Comput. Vis.}, 2009.

\bibitem{brox2010object}
T.~Brox and J.~Malik, ``Object segmentation by long term analysis of point
  trajectories,'' in \emph{Proc. of Eur. Conf. Comput. Vis.}, 2010.

\bibitem{Vidal10subspace}
R.~Vidal, ``Subspace clustering,'' \emph{IEEE Signal Processing Magzine},
  vol.~28, no.~2, pp. 52 --68, 2011.

\bibitem{ochs2011object}
P.~Ochs and T.~Brox, ``Object segmentation in video: a hierarchical variational
  approach for turning point trajectories into dense regions,'' in \emph{Proc.
  of Int. Conf. Comput. Vis.}, 2011.

\bibitem{Wren2002Pfinder}
C.~R. Wren, A.~Azarbayejani, T.~Darrell, and A.~P. Pentland, ``Pfinder:
  Real-time tracking of the human body,'' \emph{IEEE Trans. Pattern Anal. Mach.
  Intell.}, vol.~19, no.~7, pp. 780--785, 2002.

\bibitem{stauffer1999adaptive}
C.~Stauffer and W.~Grimson, ``{Adaptive Background Mixture Models for Real-Time
  Tracking},'' in \emph{Proc. of IEEE Int. Conf. Compt. Vis. Pattern Recogn.},
  1999.

\bibitem{Elgammal2000nonparametric}
A.~M. Elgammal, D.~Harwood, and L.~S. Davis, ``Non-parametric model for
  background subtraction,'' in \emph{Proc. of Eur. Conf. Comput. Vis.}, 2000.

\bibitem{mittal2004motion}
A.~Mittal and N.~Paragios, ``Motion-based background subtraction using adaptive
  kernel density estimation,'' in \emph{Proc. of IEEE Int. Conf. Compt. Vis.
  Pattern Recogn.}, 2004.

\bibitem{matsuyama2000background}
T.~Matsuyama, T.~Ohya, and H.~Habe, ``Background subtraction for non-stationary
  scenes,'' in \emph{Proc. of Asian Conf. Comput. Vis.}, 2000.

\bibitem{kim2005real}
K.~Kim, T.~Chalidabhongse, D.~Harwood, and L.~Davis, ``Real-time
  foreground-background segmentation using codebook model,'' \emph{Real-time
  Imaging}, vol.~11, no.~3, pp. 172--185, 2005.

\bibitem{friedman1997image}
N.~Friedman and S.~Russell, ``Image segmentation in video sequences: A
  probabilistic approach,'' in \emph{Uncertainty in artificial intelligence},
  1997.

\bibitem{rittscher2000probabilistic}
J.~Rittscher, J.~Kato, S.~Joga, and A.~Blake, ``A probabilistic background
  model for tracking,'' in \emph{Proc. of Eur. Conf. Comput. Vis.}, 2000.

\bibitem{monnet2003background}
A.~Monnet, A.~Mittal, N.~Paragios, and V.~Ramesh, ``{Background modeling and
  subtraction of dynamic scenes},'' in \emph{Proc. of IEEE Int. Conf. Comput.
  Vis.}, 2003.

\bibitem{zhong2003segmenting}
J.~Zhong and S.~Sclaroff, ``Segmenting foreground objects from a dynamic
  textured background via a robust kalman filter,'' in \emph{Proc. of IEEE Int.
  Conf. Comput. Vis.}, 2003.

\bibitem{wright2010sparse}
J.~Wright, Y.~Ma, J.~Mairal, G.~Sapiro, T.~Huang, and S.~Yan, ``Sparse
  representation for computer vision and pattern recognition,'' \emph{Proc. of
  the IEEE}, vol.~98, no.~6, pp. 1031--1044, 2010.

\bibitem{Oliver2000bayesian}
N.~Oliver, B.~Rosario, and A.~Pentland, ``{A Bayesian computer vision system
  for modeling human interactions},'' \emph{IEEE Trans. Pattern Anal. Mach.
  Intell.}, vol.~22, no.~8, pp. 831--843, 2000.

\bibitem{cevher2008sparse}
V.~Cevher, M.~Duarte, C.~Hegde, and R.~Baraniuk, ``Sparse signal recovery using
  markov random fields,'' in \emph{NIPS}, 2008.

\bibitem{huang2009learning}
J.~Huang, X.~Huang, and D.~Metaxas, ``Learning with dynamic group sparsity,''
  in \emph{Proc. of IEEE Int. Conf. Comput. Vis.}, 2009.

\bibitem{mairal2010network}
J.~Mairal, R.~Jenatton, G.~Obozinski, and F.~Bach, ``Network flow algorithms
  for structured sparsity,'' in \emph{NIPS}, 2010.

\bibitem{wang2006novel}
H.~Wang and D.~Suter, ``A novel robust statistical method for background
  initialization and visual surveillance,'' in \emph{Proc. of Asian Conf.
  Comput. Vis.}, 2006.

\bibitem{geman1984stochastic}
S.~Geman and D.~Geman, ``Stochastic relaxation, gibbs distributions, and the
  bayesian restoration of images,'' \emph{IEEE Trans. Pattern Anal. Mach.
  Intell.}, vol.~6, pp. 721--741, 1984.

\bibitem{recht2010guaranteed}
B.~Recht, M.~Fazel, and P.~Parrilo, ``{Guaranteed Minimum-Rank Solutions of
  Linear Matrix Equations via Nuclear Norm Minimization},'' \emph{SIAM Review},
  vol.~52, no.~3, pp. 471--501, 2010.

\bibitem{Mazumder2010spectral}
R.~Mazumder, T.~Hastie, and R.~Tibshirani, ``{Spectral Regularization
  Algorithms for Learning Large Incomplete Matrices},'' \emph{J. Mach. Learn.
  Res}, vol.~11, pp. 2287--2322, 2010.

\bibitem{cai2010singular}
J.~Cai, E.~Cand{\`e}s, and Z.~Shen, ``A singular value thresholding algorithm
  for matrix completion,'' \emph{SIAM Journal on Optimization}, vol.~20, p.
  1956, 2010.

\bibitem{Boykov01}
Y.~Boykov, O.~Veksler, and R.~Zabih, ``Fast approximate energy minimization via
  graph cuts,'' \emph{IEEE Trans. Pattern Anal. Mach. Intell.}, vol.~23,
  no.~11, pp. 1222--1239, 2001.

\bibitem{kolmogorov2004energy}
V.~Kolmogorov and R.~Zabih, ``{What Energy Functions Can Be Minimizedvia Graph
  Cuts?}'' \emph{IEEE Trans. Pattern Anal. Mach. Intell.}, vol.~26, no.~2, pp.
  147--159, 2004.

\bibitem{de2003framework}
F.~De~La~Torre and M.~Black, ``A framework for robust subspace learning,''
  \emph{Int. J. Comput. Vis.}, vol.~54, no.~1, pp. 117--142, 2003.

\bibitem{ke2005robust}
Q.~Ke and T.~Kanade, ``Robust l1 norm factorization in the presence of outliers
  and missing data by alternative convex programming,'' in \emph{Proc. of IEEE
  Int. Conf. Compt. Vis. Pattern Recogn.}, 2005.

\bibitem{zhou2010stable}
Z.~Zhou, X.~Li, J.~Wright, E.~Candes, and Y.~Ma, ``{Stable principal component
  pursuit},'' in \emph{Int. Symp. on Inf. Theory}, 2010.

\bibitem{She10}
Y.~She and A.~B. Owen, ``Outlier detection using nonconvex penalized
  regression,'' \emph{Arxiv preprint arXiv:1006.2592}, 2010.

\bibitem{zhao2006model}
P.~Zhao and B.~Yu, ``On model selection consistency of lasso,'' \emph{The J.
  Mach. Learn. Res}, vol.~7, pp. 2541--2563, 2006.

\bibitem{mazumder2011sparsenet}
R.~Mazumder, J.~Friedman, and T.~Hastie, ``Sparsenet: Coordinate descent with
  non-convex penalties,'' \emph{(To Appear) J. Am. Stat. Assoc.}, 2011.

\bibitem{donoho2006compressed}
D.~Donoho, ``Compressed sensing,'' \emph{IEEE Trans. Inf. Theory}, vol.~52,
  no.~4, pp. 1289--1306, 2006.

\bibitem{Zhou10}
Z.~Zhou, A.~Wagner, H.~Mobahi, J.~Wright, and Y.~Ma, ``Face recognition with
  contiguous occlusion using markov random fields,'' in \emph{Proc. of IEEE
  Int. Conf. Comput. Vis.}, 2010.

\bibitem{Peng10}
Y.~Peng, A.~Ganesh, J.~Wright, W.~Xu, and Y.~Ma, ``{RASL: Robust alignment by
  sparse and low-rank decomposition for linearly correlated images},'' in
  \emph{Proc. of IEEE Int. Conf. Compt. Vis. Pattern Recogn.}, 2010.

\bibitem{yuan2006model}
M.~Yuan and Y.~Lin, ``Model selection and estimation in regression with grouped
  variables,'' \emph{J. Roy. Stat. Soc. B Met.}, vol.~68, no.~1, pp. 49--67,
  2006.

\bibitem{zhao2009composite}
P.~Zhao, G.~Rocha, and B.~Yu, ``The composite absolute penalties family for
  grouped and hierarchical variable selection,'' \emph{The Annals of
  Statistics}, vol.~37, no.~6A, pp. 3468--3497, 2009.

\bibitem{szeliski2010computer}
R.~Szeliski, \emph{{Computer Vision: Algorithms and Applications}}.\hskip 1em
  plus 0.5em minus 0.4em\relax Springer, 2010.

\bibitem{odobez1995robust}
J.~Odobez and P.~Bouthemy, ``Robust multiresolution estimation of parametric
  motion models,'' \emph{J. Visual Commun. Image repres.}, vol.~6, no.~4, pp.
  348--365, 1995.

\bibitem{davis2006relationship}
J.~Davis and M.~Goadrich, ``The relationship between precision-recall and roc
  curves,'' in \emph{ICML}, 2006.

\bibitem{candes2008enhancing}
E.~Candes, M.~Wakin, and S.~Boyd, ``Enhancing sparsity by reweighted ? 1
  minimization,'' \emph{J. Fourier Anal. Appl.}, vol.~14, no.~5, pp. 877--905,
  2008.

\bibitem{li2004statistical}
L.~Li, W.~Huang, I.~Gu, and Q.~Tian, ``{Statistical modeling of complex
  backgrounds for foreground object detection},'' \emph{IEEE Trans. Image
  Processing}, vol.~13, no.~11, pp. 1459--1472, 2004.

\bibitem{parks2008evaluation}
D.~Parks and S.~Fels, ``Evaluation of background subtraction algorithms with
  post-processing,'' in \emph{IEEE Int. Conf. on Advanced Video and Signal
  Based Surveillance}, 2008, pp. 192--199.

\end{thebibliography}

\end{document}